\documentclass[article]{elsarticle}

\usepackage{amsmath,amssymb,amsthm,enumerate,amsfonts,setspace}
\usepackage{geometry}
\usepackage {multirow}
\usepackage{tabularx}
\usepackage{makecell}
\usepackage{graphicx, subfigure}
\usepackage[ruled]{algorithm2e}
\usepackage{float} 

\usepackage{xcolor}
\usepackage[normalem]{ulem} 
\newcommand\hl{\bgroup\markoverwith
	{\textcolor[rgb]{0.8,0.8,0.8}{\rule[-.5ex]{1pt}{2.5ex}}}\ULon}
\usepackage[font={bf},textfont=md,labelfont={footnotesize},labelformat={default}]{caption}
\usepackage{caption}
\usepackage[colorlinks,bookmarksopen,bookmarksnumbered,citecolor=blue, linkcolor=blue, urlcolor=blue]{hyperref}

\theoremstyle{definition}
\newtheorem{definition}{Definition}[section]
\newtheorem{remark}{Remark}[section]

\theoremstyle{plain}
\newtheorem{theorem}{Theorem}[section]

\begin{document}

\begin{frontmatter}

\title{\Large A novel multiobjective evolutionary algorithm based on decomposition and multi-reference points strategy}



\author[mymainaddress]{Wang Chen}
\author[mysecondaryaddress]{Jian Chen}
\author[mymainaddress]{Weitian Wu}
\author[mythirdaryaddress]{Xinmin Yang}
\author[myfourtharyaddress]{Hui Li}

\address[mymainaddress]{College of Mathematics, Sichuan University, Chengdu 610065, China}
\address[mysecondaryaddress]{College of Sciences, Shanghai University, Shanghai 200444, China}
\address[mythirdaryaddress]{School of Mathematical Sciences, Chongqing Normal University, Chongqing 401331, China}
\address[myfourtharyaddress]{School of Mathematics and Statistics, Xi'an Jiaotong University, Xi'an 710049, China}

\begin{abstract}
Many real-world optimization problems such as engineering design can be eventually modeled as the corresponding multiobjective optimization problems (MOPs) which must be solved to obtain approximate Pareto optimal fronts. Multiobjective evolutionary algorithm based on decomposition (MOEA/D) has been regarded as a significantly promising approach for solving MOPs. Recent studies have shown that MOEA/D with uniform weight vectors is well-suited to MOPs with regular Pareto optimal fronts, but its performance in terms of diversity usually deteriorates when solving MOPs with irregular Pareto optimal fronts. In this way, the solution set obtained by the algorithm can not provide more reasonable choices for decision makers. In order to efficiently overcome this drawback, we propose an improved MOEA/D algorithm by virtue of the well-known Pascoletti-Serafini scalarization method and a new strategy of multi-reference points. Specifically, this strategy consists of the setting and adaptation of reference points generated by the techniques of equidistant partition and projection. For performance assessment, the proposed algorithm is compared with existing four state-of-the-art multiobjective evolutionary algorithms on  benchmark test problems with various types of Pareto optimal fronts. According to the experimental results, the proposed algorithm exhibits better diversity performance than that of the other compared algorithms. Finally, our algorithm is applied to two real-world MOPs in engineering optimization successfully.
\end{abstract}


\end{frontmatter}


\section{Introduction}
The problems of simultaneously optimizing multiple conflicting objectives often arise in engineering, finance, transportation and many other fields; see \citep{A2000, RB2013, Z2015m, CC2019, TI2020}. These problems are called \emph{multiobjective optimization problems} (MOPs). It is not possible mathematically to define a single optimal solution for a given MOP but we have a set of trade-offs, that is, a set of so-called Pareto optimal solutions in the decision space, which constitute the Pareto optimal set. The image of Pareto optimal set in objective space is known as the Pareto optimal front (POF). Finding the entire POF is very time-consuming since the POF of most MOPs is frequently composed of exponential or even an infinite number of solutions. Moreover, the decision makers may not be interested in having an unduly large number of solutions. Thus, a commonly-used technique in practice is to find a representative approximation of the true POF.

Over the past two decades, we have witnessed a large variety of methods for solving MOPs; see \citep{MMR2008, FDS2009, QR2013, WPG2015, QR2018, EW2020, E2021}, the survey papers \citep{JM2002m, RW2005, FD2014, TSS2016, MYL2020, RI2020a} and the books \citep{M1999, C2007, E2008, PZZ2017}. Among the various methods mentioned in the above literature, multiobjective evolutionary algorithms (MOEAs) have attracted tremendous attention by many researchers. A reasonable interpretation is that the population-based heuristic search mechanism makes a MOEA find a suitable approximation of the entire POF in a single run. Three goals of a MOEA summarized in \cite{TSS2016} are - 1)  to find a set of solutions as close as possible to the POF (known as \emph{convergence}), 2) to find a well distributed set of solutions (known as \emph{diversity}), 3) to cover the entire POF (known as \emph{coverage}). In order to achieve these goals, several existing MOEAs can be broadly classified into three groups: Pareto dominance-based approaches (e.g., nondominated sorting genetic algorithm II (NSGA-II) \citep{DAPM2002}), indicator-based approaches (e.g., hypervolume estimation algorithm (HypE) \citep{BZ2011}) and decomposition-based approaches (e.g., cellular multiobjective genetic algorithm (cMOGA) \citep{MIG2001}, MOEA/D \citep{ZL2007}).

In the above-mentioned three groups, decomposition-based approaches have become increasingly popular recently. MOEA/D, proposed by  \cite{ZL2007}, is the most well-known and effective method in decomposition-based MOEAs. The philosophy behind MOEA/D is that it decomposes a target MOP into a series of scalar optimization subproblems by means of a set of uniform weight vectors generated by the lattice method proposed by \cite{DD1998} and a scalarization method (or, decomposition method) such as Tchebycheff, and then solves these subproblems simultaneously by using an evolutionary algorithm and evolving a population of solutions. Although MOEA/D with even weight vectors is well-suited to MOPs with regular POFs (i.e., simplex-like, e.g., a triangle and a sphere), many recent studies \citep{Q2014, LDZS2019, MYL2020} have suggested that its performance is often bottlenecked by MOPs with irregular POFs (e.g., disconnected, degenerate, inverted, highly nonlinear and badly scaled). From the variants (see the survey papers \citep{TSS2016, MYL2020, WSLM2020}) gestated by MOEA/D in the past a dozen years, it can be seen that the predefined uniform weight vectors and the scalarization approach in MOEA/D limit the diversity of population to a great extent. Therefore, the adjustment of weight vectors and the improvement of scalarization approach become two crucial ingredients in the variants of MOEA/D.

\begin{itemize}\setlength{\itemsep}{-0.03in}
	\item \emph{Weight Vectors}: The weight vectors determine the search directions and, to a certain extent, the distribution of the final solution set \citep{MYL2020}. In MOEA/D \citep{ZL2007}, the weight vectors are predefined and cannot be changed during the search process. It is exciting that various interesting attempts \citep{Q2014, GLT2012, JCZ2011, CJOS2016,  LSZ2019, DWT2020} have been made to adjust the weight vectors adaptively during the evolution process. For instance, $pa\lambda$-MOEA/D \citep{JCZ2011} uses a method called Pareto-adaptive weight vectors to automatically adjust the weight vectors via the geometrical features of the estimated POF. $pa\lambda$-MOEA/D is suitable for MOPs whose POFs have a symmetric shape, but deteriorates on MOPs with more complex POFs. DMOEA/D \citep{GLT2012} adopts the technique of equidistant interpolation to adjust weight vectors after several generations according to the projection of the current nondominated solutions.  MOEA/D-AWA \citep{Q2014} dynamically adjusts the weight vectors at the later stage of evolution. To be specific, MOEA/D-AWA periodically deletes weight vectors in crowed areas and adds ones in sparse areas.  RVEA* \citep{CJOS2016} employs the preset weight vectors in the initialization and then uses random weight vectors to replace the invalid weight vectors associated with no solution in the evolutionary process. For the MOPs with different objective scales, \cite{LSZ2019} gave a new strategy to adjust the weight vectors. This strategy modifies the component of each weight vector by multiplying a factor, which corresponds to the range of associated objective values of solutions in current population.
	\item \emph{Scalarization Approaches}: The scalarization method defines an improvement region or a contour line for every subproblem, which can significantly affect the search ability of the evolving population. In MOEA/D, the authors presented three kinds of scalarization methods, namely, weighted sum (WS), Tchebycheff (TCH) and penalty-based boundary intersection (PBI). However, the solutions obtained by these scalarization approaches with uniform weight vectors are not always uniformly distributed along POF and the performance of PBI is suffered from the penalty parameter. Moreover, the choice of scalarization method plays critical role in the performance of MOEA/D on a particular problem and it is not an easy task to choose an appropriate scalarization approach for different MOPs \citep{TSS2016}. To alleviate these drawbacks, many different scalarizing functions have been proposed in the literature; see the survey papers \citep{TSS2016, WSLM2020}. For example,  \cite{ISTN2010} used the augmented weighted Tchebycheff within the framework of MOEA/D so as to cope with the problem of selecting a suitable scalarization method for a particular problem. \cite{JY2015} proposed the reverse Tchebycheff approach, which can deal with the problems with highly nonlinear and convex POFs. \cite{WZZ2016} investigated the search ability of $L_{p}$ scalarization method with different contour lines determined by the $p$ values and then introduced a Pareto adaptive scalarizing approximation to approach the optimal $p$ value at different search stages. \cite{MZT2018} proposed the Tchebycheff scalarization approach with $l_{p}$-norm constraint on direction vectors in which the experimental results show that this method is capable of obtaining high quality solutions.
\end{itemize}

These methods in the aforementioned literature improve the diversity of the final solution set obtained by corresponding algorithms to a certain extent. Unfortunately, the ability of scalarization function and the adjustment of weight vectors are still limited. In particular, for the MOP whose POF has the shape of highly nonlinearity and convexity, e.g., the hatch cover design problem \citep{TI2020} and GLT3 \citep{GLT2012}, these methods do not seem to be very effective. More importantly, we observe that various scalarization approaches introduced in MOEA/D and its variants all take into account the information on the weight vectors and the ideal point (see the scalarization approaches summarized in Table \ref{scalarization} of Subsection \ref{sec2.2}). From the geometric point of view, we attribute these methods to a category of ``\emph{single-point and multi-directions}", as shown in Fig. \ref{graphical-interpretation} in Subsection \ref{sec2.2}. Herein, \emph{single-point} stands for the ideal point or the nadir point and \emph{multi-directions} denote a set of preset uniformly distributed weight vectors. The left part of Fig. \ref{tch_ps} in Subsection \ref{motivation} reveals that the diversity of the final solution set is vulnerable to this geometric phenomenon.

In order to essentially change this geometric phenomenon, an intuitive idea called ``\emph{multi-points and single-direction}" is proposed (see the right part of Fig. \ref{tch_ps} in Subsection \ref{motivation}). As a result, the aforesaid scalarization method may no longer be applicable. Now, the question is whether there is a scalarization method without considering the weight vectors and the ideal point, and then matches this idea? Here, it is shown that the answer to this question is positive. The Pascoletti-Serafini (PS) scalarization method with additional constraints proposed by \cite{ PS1984} does not rely on the weight vector and the ideal point. It has two parameters, i.e., reference point and direction, which by varying them in $R^{m}$ ($m$-dimensional Euclidean space), all Pareto optimal solutions can be obtained for a given MOP. An advantage of the PS scalarization method is that it is very general in the sense that many other well-known and commonly-used scalarization methods such as the WS method, the $\epsilon$-constraint method, the generalized WS method and the TCH method can be seen as a special case of it (see Section 2.5 in \cite{E2008}). For researches on theories and applications of the PS method and its variants, we refer the interested readers to the literature \citep{E2008, E2009a, BBA2014, KKK2014, AGK2018, DE2019t, TY2021} for more details. Note that \cite{E2008} indicated that the optimal solution obtained by the PS scalarization method is the intersection point between the POF and the negative ordering cone along the line generated by reference point and direction. Therefore, in order to obtain a set of uniformly distributed solutions which can well approximate the true POF, the setting of the reference point and the direction in the PS scalarization approach is crucial.

In this paper, we propose a new multiobjective evolutionary algorithm based on decomposition and adaptive multi-reference points strategy, termed as MOEA/D-AMR, which performs well in diversity. The main contributions of this paper can be concluded as follows:

\begin{itemize}\setlength{\itemsep}{-0.03in}
	\item Using a standard trick from mathematical programming, we equivalently transform the PS scalarization problem into a minimax optimization problem when each component of the direction is restricted to positive. Based on the proposed idea (i.e., ``\emph{multi-points and single-direction}"), a given MOP is decomposed into a series of the transformed minimax optimization subproblems.
	\item A strategy of setting multi-reference points is introduced. More specifically, the selection range of reference points is limited to a convex hull formed by the projection points of the vertices of the hypercube $[0,1]^{m}$ on a hyperplane and then the generation of reference points is realized by using two techniques including equidistant partition and projection.
	\item A multi-reference points adjustment strategy based on the obtained solutions in the later stage of evolution is proposed. This strategy can identify the promising reference points, delete unpromising ones and generate some new ones.
	\item We verify the diversity performance of the proposed algorithm by comparing it with four representative MOEAs on a series of benchmark multiobjective test problems with regular and irregular POFs. The proposed algorithm is used to solve two real-world MOPs in engineering optimization including the hatch cover design and the rocket injector design. The experimental results illustrate the effectiveness of our algorithm.
\end{itemize}

The rest of this paper is organized as follows. Section 2 gives some fundamental definitions related to multiobjective optimization and recalls several scalarization approaches. Section 3 discusses the motivation of the proposed algorithm and illustrates the details of its implemention. Algorithmic comparions and analysis on test problems are presented in Section 4, followed by applications on real-world MOPs in Section 5. Lastly, Section 6 concludes this paper and identifies some future plans.

\section{Related works}
We start with the descriptions of some basic concepts in multiobjective optimization. Then, we recall some scalarization methods used in MOEA/D framework. Finally, a brief review of the Pascoletti-Serafini scalarization method is presented.

\subsection{Basic concepts}

Throughtout this paper, for $n,m\in\mathbb{N}$, where $\mathbb{N}$ denotes the set of natural numbers, we use the symbols

\begin{equation*}
	\langle m\rangle=\{1,\ldots,m\},\quad n_{m}=\underbrace{(n,\ldots,n)}\limits_{m}.
\end{equation*}
\noindent Let $R_{+}^{m}$ be the nonnegative orthant of $m$-dimensional Euclidean space $R^{m}$, i.e,

$$R_{+}^{m}=\{y=(y_{1},\ldots,y_{m})\in R^{m}:y_{i}\geq0,i\in\langle m\rangle\}.$$

We consider the following multiobjective optimization problem:

\begin{equation}\label{mop}
	\begin{aligned}
		&\text{min}\quad F(x)=(f_{1}(x),...,f_{m}(x))^{T}\\
		&\text{s.t.}\quad\; x=(x_{1},\ldots,x_{n})^{T}\in \Omega,
	\end{aligned}
\end{equation}
\noindent where $x$ is a decision variable vector, $\Omega=\prod_{i=1}^{n}[l_{i},u_{i}]\subseteq R^{n}$ is the decision (search) space, $l_{i}$ and $u_{i}$ are the lower and upper bounds of the $i$-th decision variable $x_{i}$, respectively. $F:\Omega\rightarrow R^{m}$ consists of $m$ real-valued objective functions $f_{i}$, $i\in\langle m\rangle$ and $R^{m}$ is also called the objective space. Since these objectives conflict with one another, it is not possible to find a feasible point that minimizes all objective functions at the same time. Therefore, it is necessary to give a concept of optimality which described in \cite{M1999}.

\begin{definition}
	A point $\hat{x}\in\Omega$ is called a
	\begin{enumerate}[(i)]\setlength{\itemsep}{-0.03in}
		\item Pareto optimal solution of (\ref{mop}) if there is no $x\in\Omega$ such that $f_{i}(x)\leq f_{i}(\hat{x})$ for all $i\in\langle m\rangle$ and $f_{j}(x)< f_{j}(\hat{x})$ for at least one index $j\in\langle m\rangle$.
		\item weakly Pareto optimal of (\ref{mop}) if there is no $x\in\Omega$ such that $f_{i}(x)< f_{i}(\hat{x})$ for all $i\in\langle m\rangle$.
	\end{enumerate}
\end{definition}

\begin{remark}
	It is obvious that if $\hat{x}\in\Omega$ is a Pareto optimal solution of (\ref{mop}), then $\hat{x}$ is a weakly Pareto optimal solution.
\end{remark}

\begin{definition}
	If $\hat{x}$ is a (weakly) Pareto optimal solution of (\ref{mop}), then $F(\hat{x})$ is called a (weakly) Pareto optimal vector.
\end{definition}

\begin{definition}
	The set of all Pareto optimal solutions is called the Pareto optimal set (POS). The set of all Pareto optimal vectors, ${\rm POF}=\{F(x)\in R^{m}:x\in{\rm POS}\}$, is called the Pareto
	optimal front (POF).
\end{definition}

\begin{definition}\label{ideal_point_def}
	A point $z^{*}=(z_{1}^{*},\ldots,z_{m}^{*})$ is called an ideal point if $z_{i}^{*}=\min_{x\in\Omega}f_{i}(x)$ for each $i\in\langle m\rangle$.
\end{definition}

\begin{remark}
	Note that we have in Definition \ref{ideal_point_def} that $f_{i}(x)-z_{i}^{*}\geq0$ for all $x\in\Omega$, $i\in\langle m\rangle$.
\end{remark}

\begin{definition}
	A point $z^{{\rm nad}}=(z_{1}^{{\rm nad}},\ldots,z_{m}^{{\rm nad}})$ is called a nadir point if $z_{i}^{{\rm nad}}=\max_{x\in{\rm POS}}f_{i}(x)$ for each $i\in\langle m\rangle$.
\end{definition}

\subsection{Scalarization methods in MOEA/D framework}\label{sec2.2}
Let $w=(w_{1},\ldots,w_{m})$ be a weight vector, i.e., $\sum_{i=1}^{m}w_{i}=1$ and $w_{i}> 0$ for all $i\in\langle m\rangle$. The formulas of three traditional scalarization approaches (i.e., WS \citep{M1999}, TCH  \citep{M1999} and PBI \citep{ZL2007}) and other six scalarization methods (i.e., the augmented Tchebycheff (a-TCH) \citep{ISTN2010}, the modified Tchebycheff (m-TCH) \citep{LZK2014}, the reverse Tchebycheff (r-TCH) \citep{JY2015}, $L_{p}$ scalarization \citep{WZZ2016}, the multiplicative scalarizing function (MSF) \citep{JYWL2017} and the Tchebycheff with $l_{p}$-norm constraint ($p$-TCH) \citep{MZT2018}) used in MOEA/D framework are summarized in Table \ref{scalarization}. More scalarization approaches within the framework of MOEA/D can be found in the survey papers \citep{TSS2016, WSLM2020}.

\begin{table}[H]\footnotesize
	\centering
	\caption{\footnotesize The scalarization methods used in MOEA/D framework.}
	\begin{tabular}{ll}
		\hline
		Scalarization & Formulas \\
		\hline
		WS & $\min_{x\in\Omega}\quad g^{{\rm ws}}(x|w)=\sum_{i=1}^{m}w_{i}f_{i}(x)$\\
		TCH &$\min_{x\in\Omega}\quad g^{{\rm tch}}(x|w, z^{*})=\max_{1\leq i\leq m}\{w_{i}(f_{i}(x)-z_{i}^{*})\}$\\
		PBI &$\min_{x\in\Omega}\quad g^{{\rm pbi}}(x|w, z^{*})=d_{1}+\theta d_{2}$, $d_{1}=\frac{\|(F(x)-z^{*})^{T}w\|}{\|w\|}$, $d_{2}=\|F(x)-(z^{*}+d_{1}w)\|$, $\theta>0$\\
		a-TCH  &  $ \min_{x\in\Omega}\quad g^{{\rm atch}}(x|w, z^{*})=\max_{1\leq i\leq m}\{w_{i}|z_{i}^{*}-f_{i}(x)|\}+\rho\sum_{j=1}^{m}|f_{j}(x)-z_{j}^{*}|$, $\rho>0$ \\
		m-TCH & $ \min_{x\in\Omega}\quad g^{{\rm mtch}}(x|w, z^{*})=\max_{1\leq i\leq m}\left\{\frac{f_{i}(x)-z_{i}^{*}}{w_{i}}\right\}$\\
		r-TCH & $ \max_{x\in\Omega}\quad g^{{\rm rtch}}(x|w, z^{{\rm nad}})=\min_{1\leq i\leq m}\{w_{i}(z_{i}^{{\rm nad}}-f_{i}(x))\}$\\
		$L_{p}$  & $\min_{x\in\Omega}\quad  g^{{\rm wd}}(x|w, z^{*})=(\sum_{i=1}^{m}(\frac{1}{w_{i}})^{p}(f_{i}(x)-z_{i}^{*})^{p})^{\frac{1}{p}}$, $p\geq1$\\
		MSF  & $\min_{x\in\Omega}\quad g^{{\rm msf}}(x|w, z_{i}^{*})=\frac{\left[\max_{1\leq i\leq m}\left\{\frac{1}{w_{i}}(f_{i}(x)-z_{i}^{*})\right\}\right]^{1+\beta}}{\left[\min_{1\leq i\leq m}\left\{\frac{1}{w_{i}}(f_{i}(x)-z_{i}^{*})\right\}\right]^{\beta}}$\\
		$p$-TCH & $\min_{x\in\Omega}\quad g^{{\rm ptch}}(x|\lambda, z^{*})=\max_{1\leq i\leq m}\left\{\frac{f_{i}(x)-z_{i}^{*}}{\lambda_{i}}\right\}$,  $\lambda=(\lambda_{1},\ldots,\lambda_{m})$ with $\|\lambda\|_{p}=1$ and $\lambda_{i}>0,i\in\langle m\rangle$ \\
		\hline
	\end{tabular}
	\label{scalarization}
\end{table}

We select six representative scalarization methods from Table \ref{scalarization} to give some illustrations.

\begin{figure}[H]
	\centering
	\subfigure[\footnotesize WS]{
		\begin{minipage}[t]{0.255\linewidth}
			\centering
			\includegraphics[width=4cm,height=4cm]{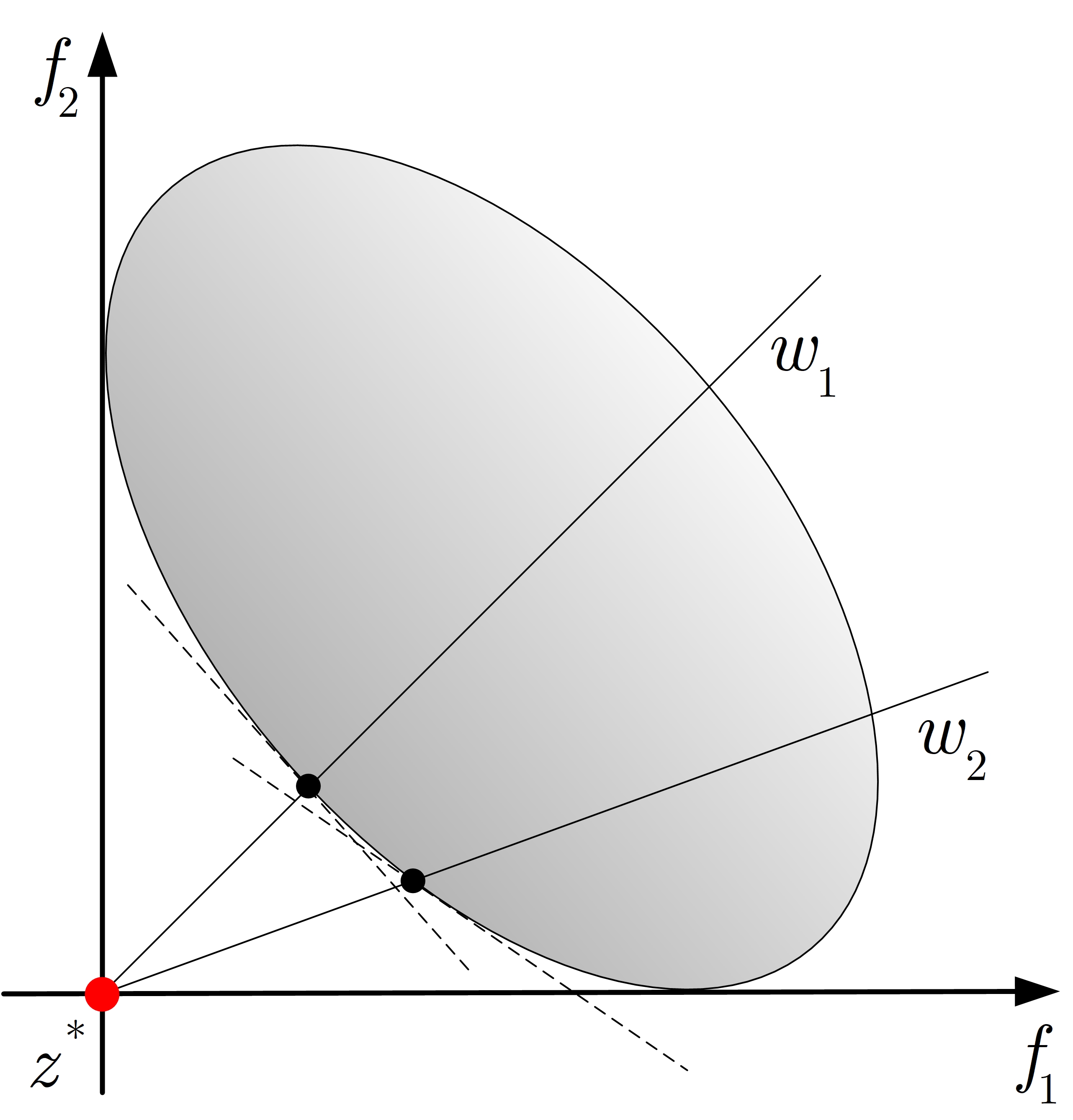}
		\end{minipage}%
	}%
	\subfigure[\footnotesize TCH]{
		\begin{minipage}[t]{0.255\linewidth}
			\centering
			\includegraphics[width=4cm,height=4cm]{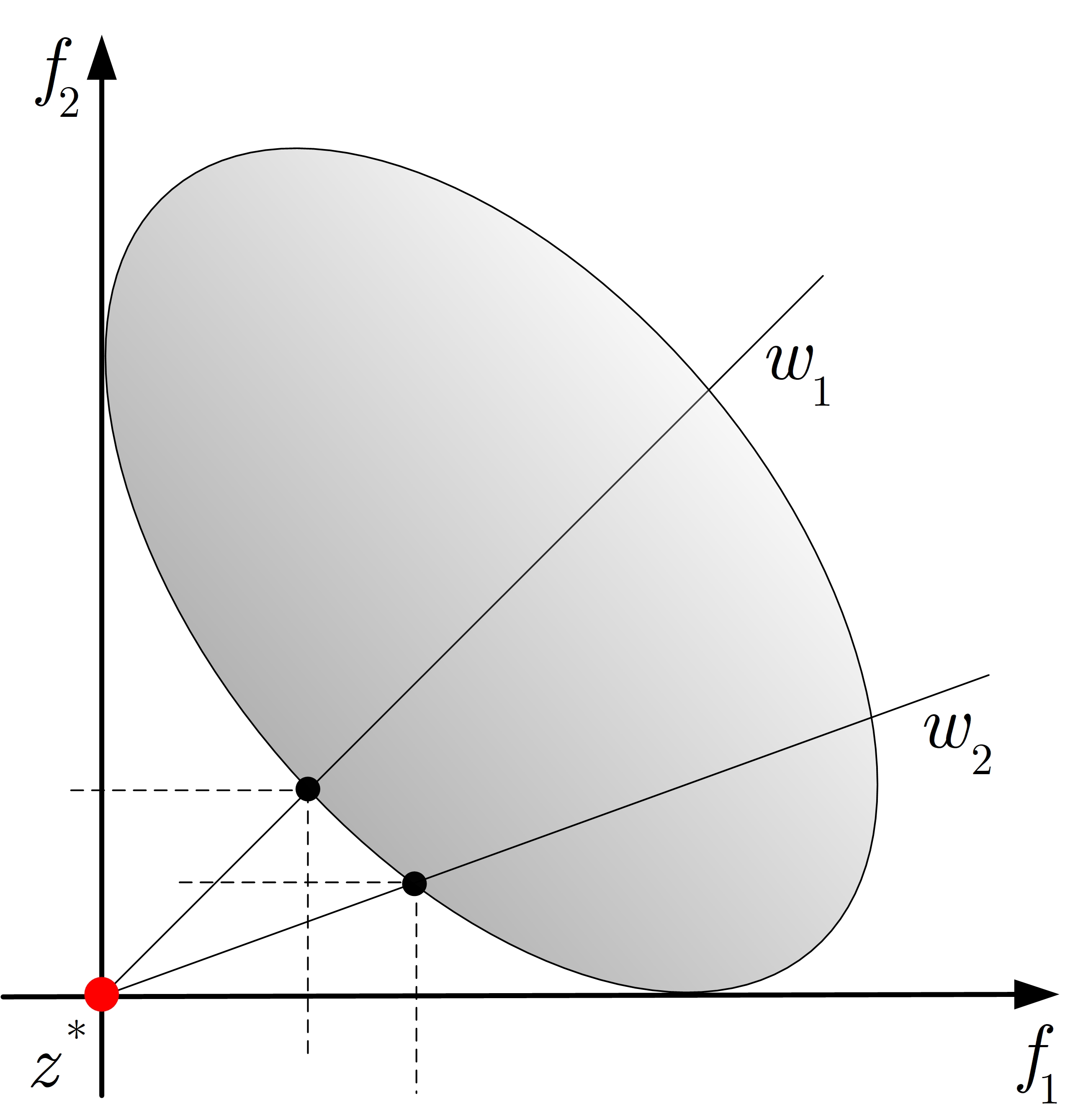}
		\end{minipage}%
	}%
	\subfigure[\footnotesize PBI]{
		\begin{minipage}[t]{0.255\linewidth}
			\centering
			\includegraphics[width=4cm,height=4cm]{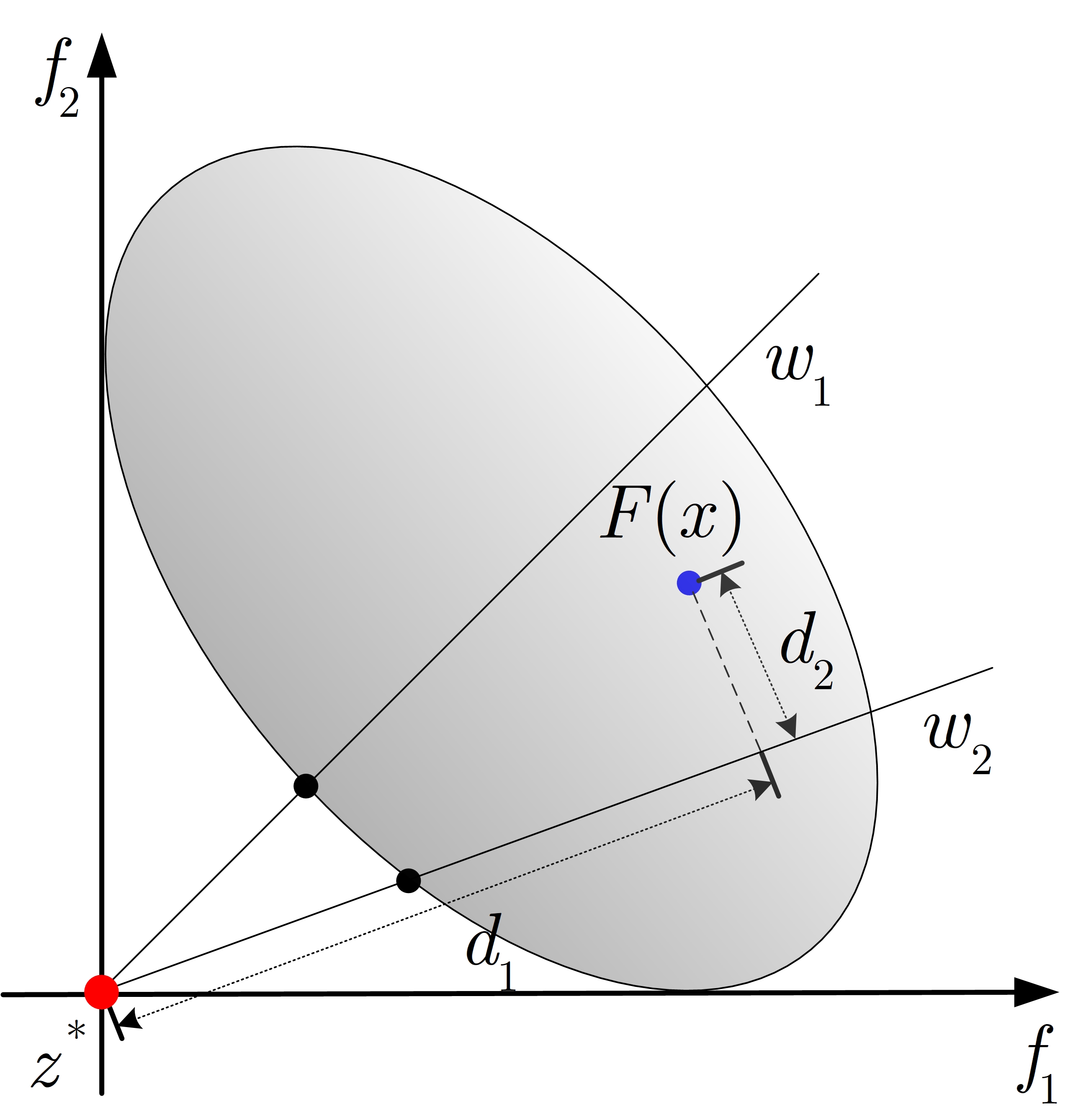}
		\end{minipage}
	}   \\
	\subfigure[\footnotesize r-TCH]{
		\begin{minipage}[t]{0.255\linewidth}
			\centering
			\includegraphics[width=4cm,height=4cm]{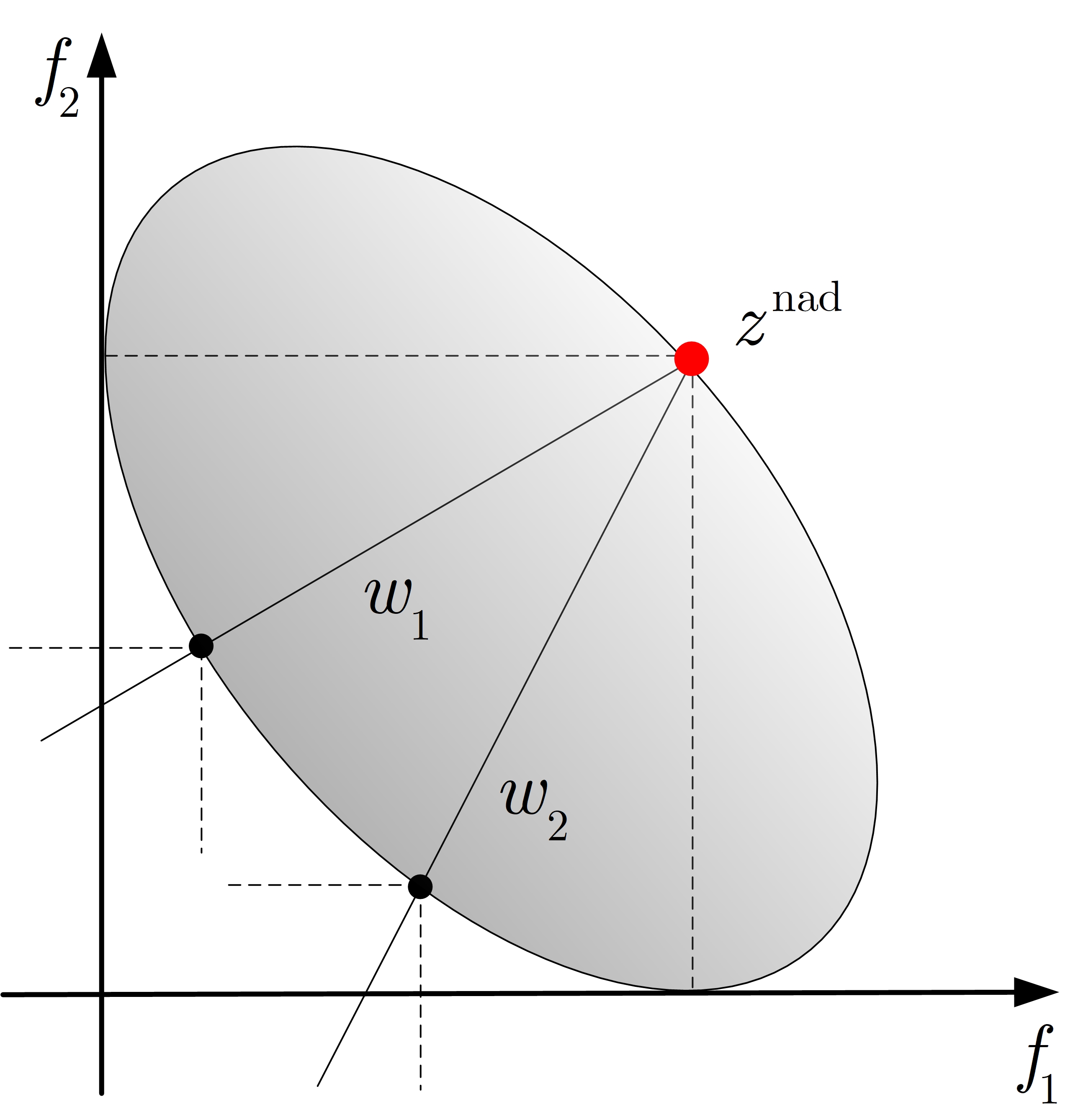}
		\end{minipage}%
	}%
	\subfigure[\footnotesize $L_{p}$]{
		\begin{minipage}[t]{0.255\linewidth}
			\centering
			\includegraphics[width=4cm,height=4cm]{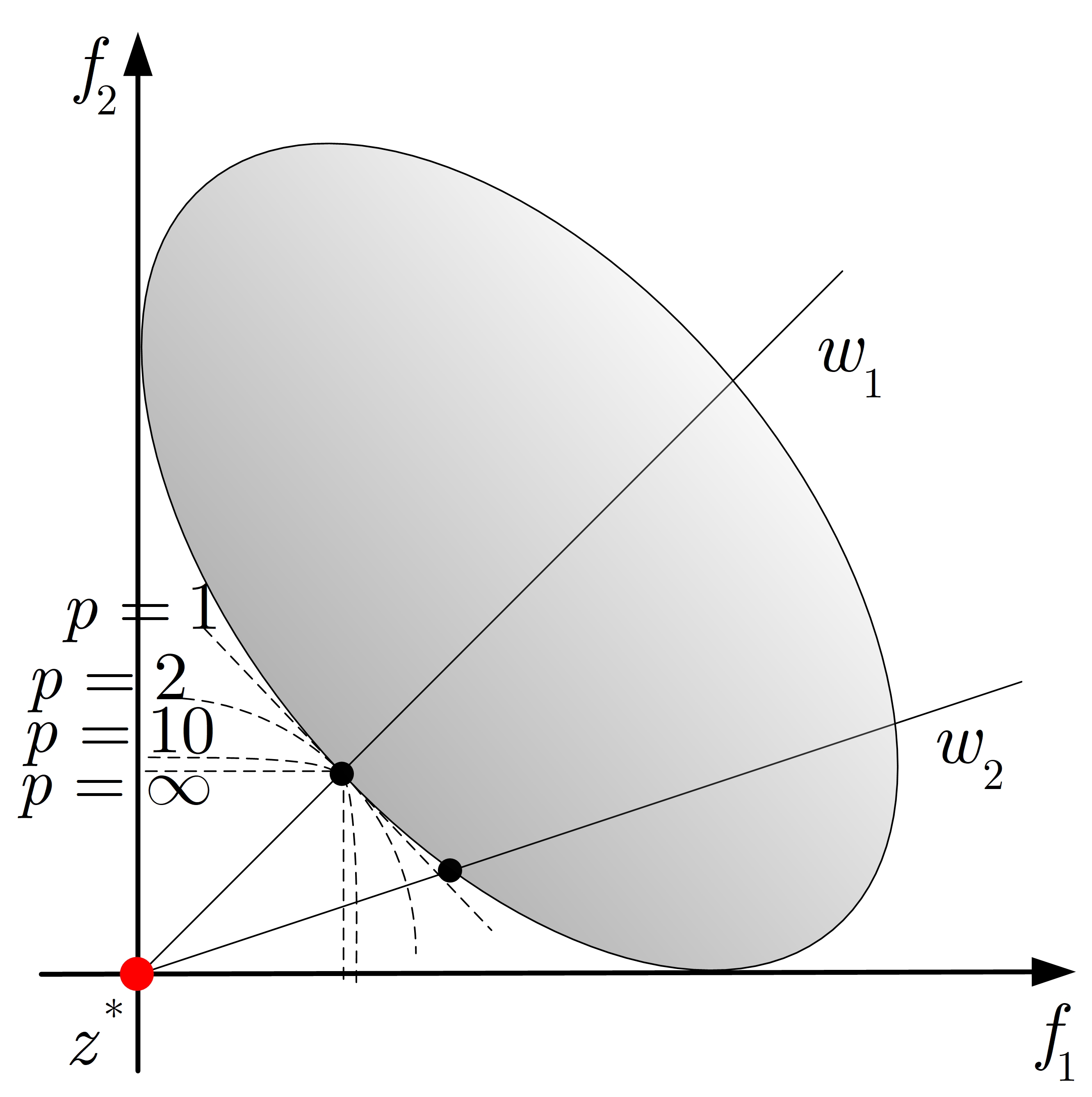}
		\end{minipage}%
	}%
	\subfigure[\footnotesize MSF]{
		\begin{minipage}[t]{0.255\linewidth}
			\centering
			\includegraphics[width=4cm,height=4cm]{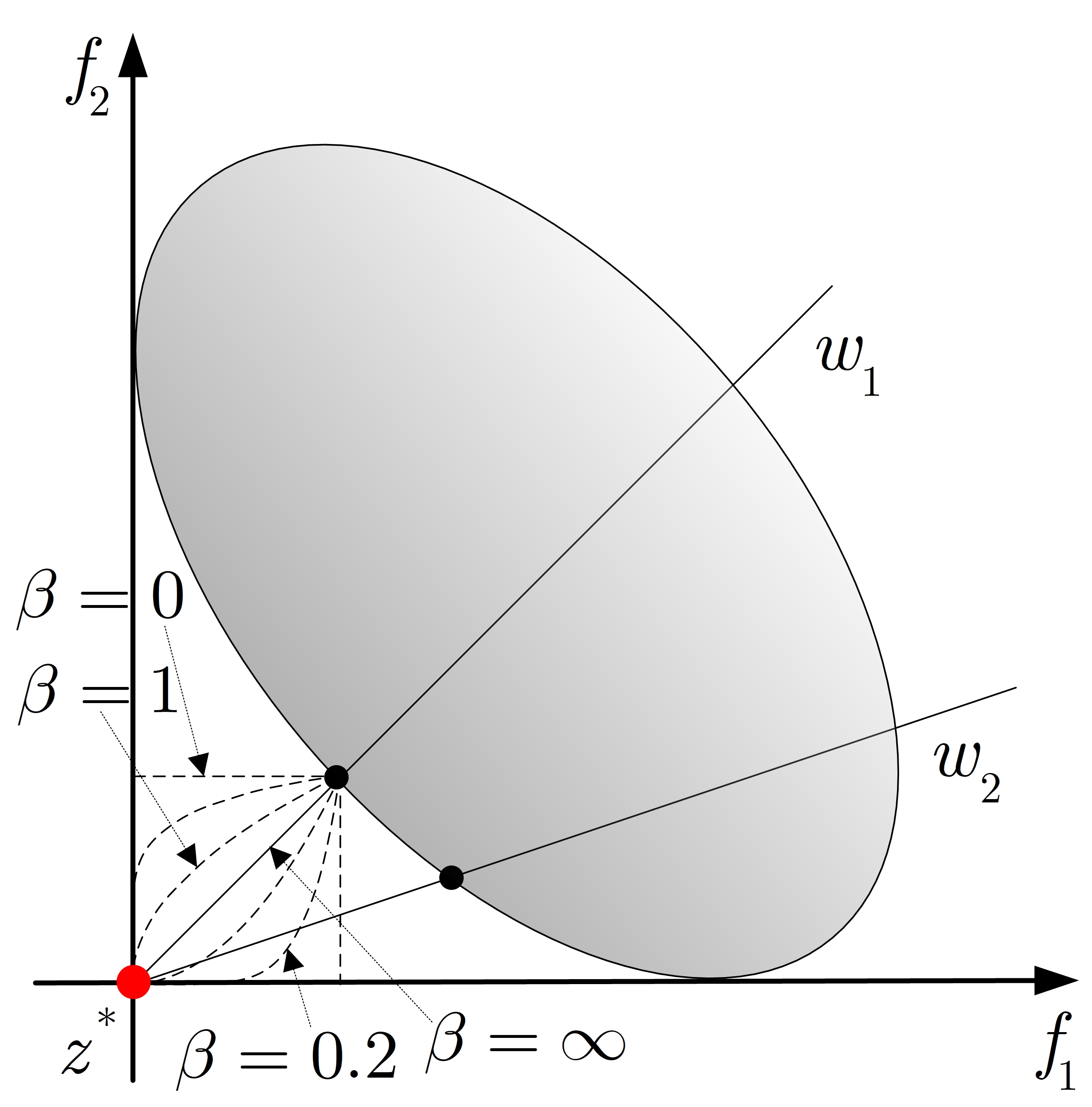}
		\end{minipage}
	}
	\centering
	\caption{\footnotesize Graphical interpretations of six representative scalarization approaches used in MOEA/D framework. Herein, we only consider two weight vectors $w^{1}$ and $w^{2}$. The black solid points denote the optimal solutions of the corresponding scalarizing subproblems.}
	\label{graphical-interpretation}
\end{figure}

\begin{itemize}\setlength{\itemsep}{-0.03in}
	\item \emph{WS method}: The WS method is also called the linear scalarization. It is probably the most commonly used scalarization technique for MOPs in traditional mathematical programming. It associates every objective function with a weighting coefficient and optimizes real-valued function of weighted sum of the objectives. A major difficulty of this approach is that if the POF is not convex, then there does not exist any weight vector such that the solution lies in the nonconvex part.
	\item \emph{TCH method}: This approach uses preference information received from a decision maker to find the Pareto optimal solution. The preference information consists of a weight vector and an ideal point. It is used as an scalarization method in MOEA/D for continuous MOPs, because it can deal with both convex POFs and concave POFs. It is noteworthy that the final solution set obtained by MOEA/D with the TCH approach is not well-distributed along the regular POFs.
	\item \emph{PBI method}: This is a direction-based decomposition approach, which uses two distance to reflect the assessment of convergence measured by $d_{1}$ and diversity measured by $d_{2}$ of population in MOEA/D, respectively (see Fig. \ref{graphical-interpretation}(c)). The penalty parameter $\theta$ in PBI is used to control the balance between convergence and diversity. A drawback of this method is that the penalty parameter is to be properly tuned.
	\item \emph{r-TCH method}: It is an inverted version of the TCH method, which is determined by a weight vector and a nadir point (see Fig. \ref{graphical-interpretation}(d)). Compared with the TCH approach, the r-TCH approach is superior in diversity when solving a MOP whose POF has a shape of highly nonlinearity and convexity. However, its performance deteriorates on MOPs with highly nonlinear and concave POFs.
	\item \emph{$L_{p}$ method}: Contour curves of this method with different $p$ values are shown in Fig. \ref{graphical-interpretation}(e). In this approach, there is a trade-off dependent on the $p$ value between the search ability of the method and its robustness on POF geometries \citep{WZZ2016}. As the value of $p$ increases, the search ability of the associated $L_{p}$ scalarization approach decreases. Therefore, a strategy called Pareto adaptive scalarizing approximation is introduced to determine the optimal $p$ value. The WS and TCH can be derived by setting $p=1$ and $p=\infty$, respectively \citep{WZZ2016}.
	\item \emph{MSF method}: A main feature of the MSF approach is the shape or positioning of its contour lines, which play a key role in search ability. As compared with the $L_{p}$ method, the opening angle of contour lines of MSF approach is less than $\pi/2$ and it is controlled by the parameter $\beta$. The geometric figure of this approach is depicted in Fig. \ref{graphical-interpretation}(f). With the increasing of $\beta$, the geometric figures become closer to the used weight vector. Clearly, when $\beta=+\infty$, the geometric figure overlaps with the weight vector and when $\beta=0$, MSF degenerates to m-TCH.
\end{itemize}

When $m=2$, the graphical interpretations of the six scalarization approaches and the positions of optimal solutions for subproblems are plotted in Fig. \ref{graphical-interpretation}.

\subsection{The Pascoletti-Serafini scalarization method}\label{ps_scalarization_method}
This subsection focuses on the scalarization method introduced by \cite{PS1984}. The scalar problem of the Pascoletti-Serafini (PS) scalarization method with respect to the ordering cone $R_{+}^{m}$ is defined as follows:

\begin{equation}\label{ps}
	\begin{aligned}
		\min&\quad t\\
		\text{s.t.}&\quad a+tr-F(x)\in R^{m}_{+}, \\
		&\quad x\in \Omega, t\in R,
	\end{aligned}
\end{equation}
\noindent where $a=(a_{1},\ldots,a_{m})$ and $r=(r_{1},\ldots,r_{m})$ are parameters selected from $R^{m}$.

\begin{remark}
	\begin{enumerate}[(i)]\setlength{\itemsep}{-0.03in}
		\item The paremeters $a$ and $r$ in (\ref{ps}) are respectively described as a reference point and a direction in \cite{E2008}.
		\item Compared with the scalarization methods presented in Table \ref{scalarization}, it is obvious that the PS method does not take into account the information on the weight vector and the ideal point or the nadir point.
	\end{enumerate}
\end{remark}

The geometric interpretation of PS method presented in \cite{E2008} or \cite{KKK2014} is that the ordering cone $-R^{m}_{+}$ is moved in the direction $r$ (or $-r$) on the line $a+tr$ starting in the point $a$ until the set $(a+tr-R^{m}_{+})\cap F(\Omega)$ is reduced to the empty set, where $a+tr-R_{+}^{m}=\{a+tr-y:y\in R^{m}_{+}\}$. The smallest value $\hat{t}$ for which $(a+\hat{t}r-R^{m}_{+})\cap F(\Omega)\neq\emptyset$ is the optimal value of (\ref{ps}). If the point pair $(\hat{t},\hat{x})$ is an optimal solution of (\ref{ps}), then $F(\hat{x})$ with $F(\hat{x})\in(a+\hat{t}r-R_{+}^{m})\cap F(\Omega)$ is an at least weakly Pareto optimal vector and by a variation of the parameters $(a,r)\in R^{m}\times R^{m}$, all Pareto optimal vectors can be obtained. To have an intuitive glimpse of (\ref{ps}), we use the following Fig. \ref{ps_Graphical} to give the graphical illustration of (\ref{ps}) in the case of $m=2$.

\begin{figure}[H]
	\centering
	\includegraphics[width=10cm,height=4cm]{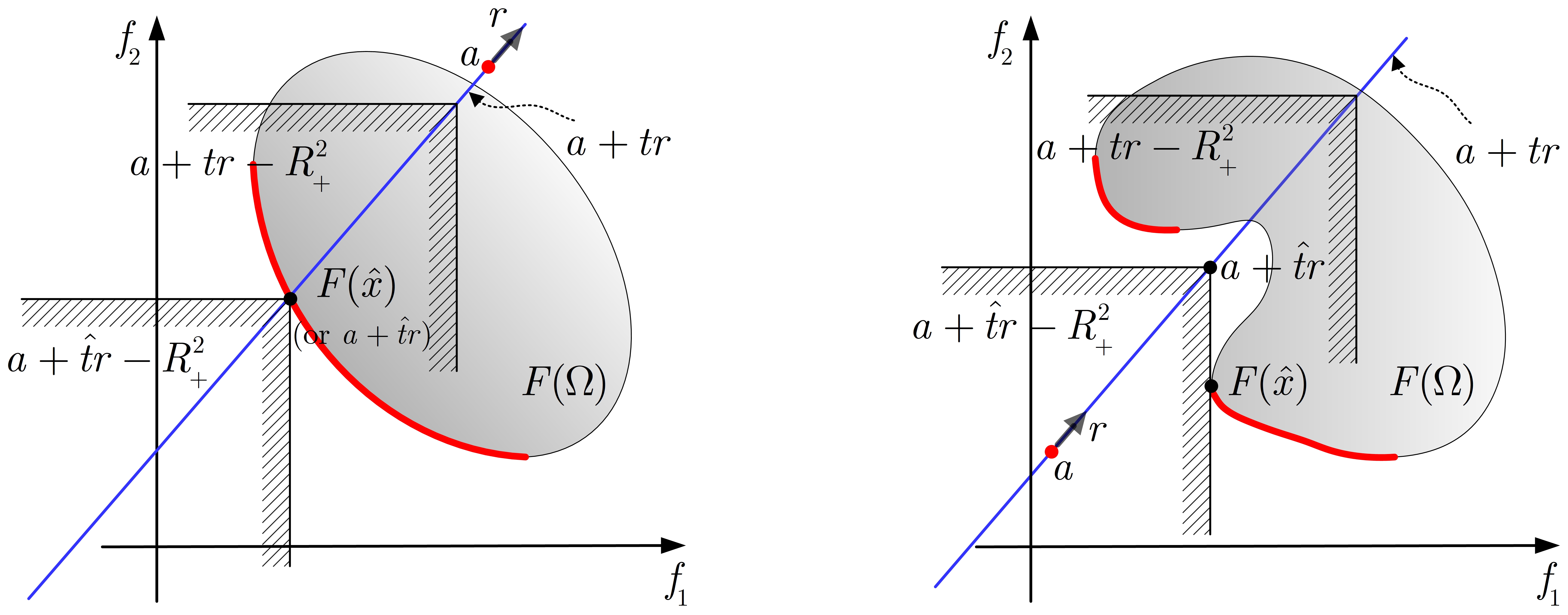}
	\caption{\footnotesize Graphical illustration of (\ref{ps}). The red curves stand for the POFs. The left part has a continuous POF while the right part has a discontinuous POF.}
	\label{ps_Graphical}
\end{figure}

Some interesting properties of this scalarization approach can be found in \cite{E2008}. Here we concentrate on the following major result (see Theorem 2.1(b) and (c) in \cite{E2008}), which gives the relation of solutions between (\ref{mop}) and (\ref{ps}).

\begin{theorem}\label{thm1}
	\begin{enumerate}[{\rm(i)}]
		\setlength{\itemsep}{-0.03in}
		\item If $\hat{x}$ is a Pareto optimal solution of (\ref{mop}), then $(\hat{x}, 0)$ is an optimal solution of  (\ref{ps}) for the parameter $a=F(\hat{x})$ and for arbitrary $r\in R_{+}^{m}\backslash\{0_{m}\}$.
		\item If $(\hat{x}, \hat{t})$ is an optimal solution of  (\ref{ps}), then $\hat{x}$ is a weakly Pareto solution of (\ref{mop}).
	\end{enumerate}
\end{theorem}

\begin{remark}\label{rem2.4}
	As reported in \cite{E2008}, a direct result of Theorem \ref{thm1} is that we can find all Pareto optimal solutions of (\ref{mop}) for a constant parameter $r\in R_{+}^{m}\backslash\{0_{m}\}$ by varying the parameter $a\in R^{m}$ only.
\end{remark}

\section{The proposed algorithm}
In this section, the motivation on use of the PS method and an adaptive multi-reference points adjustment strategy for controlling the diversity is first discussed. Then we give specific answers to the questions proposed in motivation. Finally, the detailed implementation of the proposed algorithm is presented.

\subsection{Motivation}\label{motivation}
\begin{figure}[htbp]
	\centering
	\includegraphics[width=10cm,height=4cm]{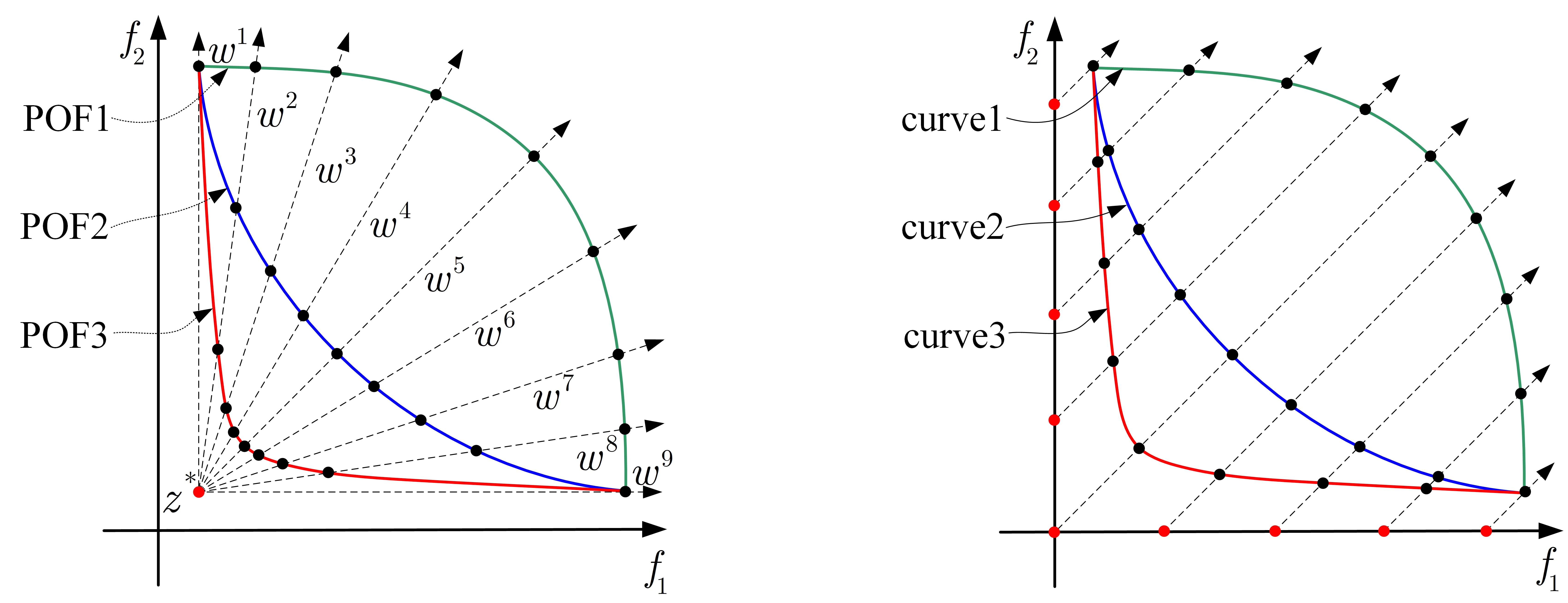}
	\caption{\footnotesize Intersection points between a series of lines and some curves in two dimensional space.}
	\label{tch_ps}
\end{figure}

It follows from these scalarization approaches given in Table \ref{scalarization} and the geometric interpretations shown in Fig. \ref{graphical-interpretation} that there is an interesting phenomenon, which we call ``\emph{single-point and multi-directions}". This is attributed to the ideal point (or the nadir point) and the uniform weight vectors. Such a phenomenon results in the fact that the optimal solutions of all subproblems defined by uniform weight vectors in MOEA/D can form a good approximation for the regular POF. However, it usually leads to the unsatisfactory performance on solving MOPs with irregular POFs. Let us now take an example to illustrate this case. In the left part of Fig. \ref{tch_ps}, there are nine uniform weight vectors $w^{1},\ldots,w^{9}$ and an ideal point $z^{*}$ depicted by red solid point. The green, blue and red curves denote POF1, POF2 and POF3, respectively, and the black solid points stand for the optimal solutions for subproblems. It is clear to see that the solutions are roughly uniformly distributed along POF1 and POF2. However, when POF shape is highly nonlinear and convex like POF3 depicted in the left part of Fig. \ref{tch_ps}, uniform weighting strategy can not produce a set of evenly distributed solutions.

It can be said that the intersection points between a set of equidistant parallel lines and any curves are almost evenly distributed along the curves. Based on this fact, a fundamental idea comes up naturally: ``\emph{multi-points and single-direction}". As shown in the right part of Fig. \ref{tch_ps}, there are nine equidistant parallel lines generated by nine red solid reference points and a direction. The black solid points are the intersection points between the nine equidistant parallel lines and three curves. Obviously, these intersection points are uniformly distributed along curve1--curve3.

\begin{figure}[htbp]
	\centering
	\includegraphics[width=4cm,height=4cm]{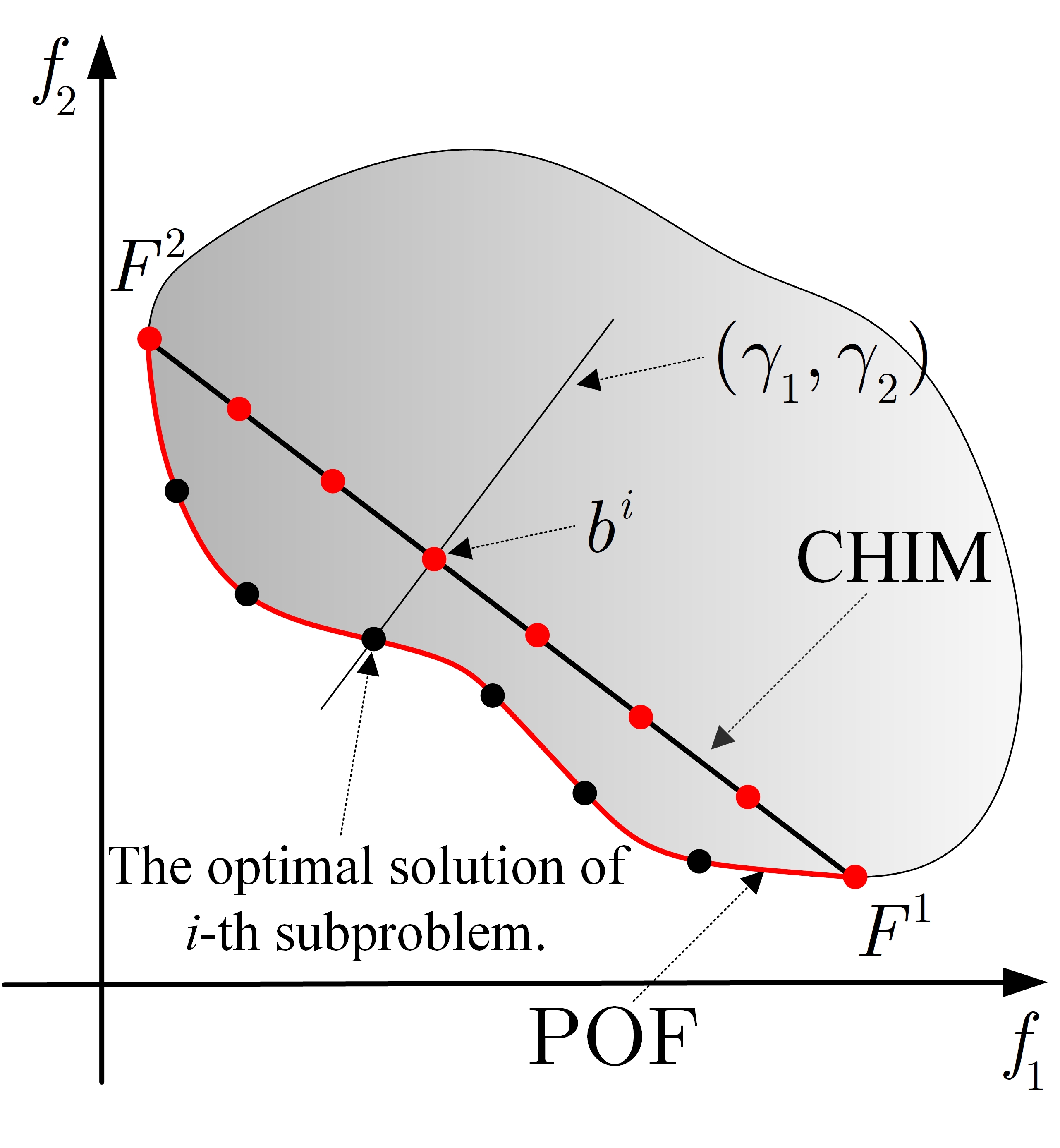}
	\caption{\footnotesize Illustration of NBI-style Tchebycheff method.}
	\label{nbi_tch}
\end{figure}

Traces of this idea of studying the solutions involved in MOPs go back to the work of \cite{DD1998}. They proposed a direction-based decomposition approach, which is called normal boundary intersection (NBI). This approach can obtain the intersection points between the lower boundary part of the image set of the decision space and a series of straight lines defined by a normal vector and a group of evenly distributed points in the convex hull of individual minima (CHIM). Despite the intersection points generated by NBI are uniform, the method has limitations recognized by \cite{DD1998}. In particular, the NBI method may fail to cover the entire POF for problems with more than three objectives. In addition, this method can not guarantee the Pareto optimality of these intersection points. An advantage of the NBI method is that it is relatively insensitive to the scales of objective functions. \cite{ZLD2010} pointed out that NBI can not be easily used within MOEA/D framework because it has additional constraints. A meaningful attempt in their work is that they absorbed the strengthens of NBI and TCH, and introduced the NBI-style Tchebycheff method for biobjective optimization problem. We next briefly describe this method. Consider $m=2$ in (\ref{mop}) and let $F^{1}=(F_{1}^{1},F_{2}^{1})$ and $F^{2}=(F_{1}^{2},F_{2}^{2})$ be two extreme points of the POF of (\ref{mop}) in the objective space. The reference points $b^{i}$, $i\in\langle N\rangle$, are evenly distributed along the line segment linking $F^{1}$ and $F^{2}$, i.e., $b^{i}=\xi_{i}F^{1}+(1-\xi_{i})F^{2}$, where $\xi_{i}=\frac{N-i}{N-1}$ for $i\in\langle N\rangle$. Therefore, the $i$-th NBI-style Tchebycheff scalarizing subproblem is

\begin{equation}\label{nbi_tch_sub}
	\min\limits_{x\in\Omega}\quad g^{{\rm nbi\text{-}tch}}(x|b^{i},\gamma)=\max\{\gamma_{1}(f_{1}(x)-b_{1}^{i}),\gamma_{2}(f_{2}(x)-b_{2}^{i})\},
\end{equation}
\noindent where $\gamma_{1}=|F^{2}_{2}-F^{1}_{2}|$ and $\gamma_{2}=|F^{2}_{1}-F^{1}_{1}|$. The graphical illustation of the NBI-style Tchebycheff method is shown in Fig. \ref{nbi_tch}. As we have seen, the optimal solutions to the above subproblems can be uniformly distributed along the POF. However, as pointed out by \cite{LDZS2019}, the main weakness of the NBI-style Tchebycheff method is that it can not be extended to handle the MOPs with more than two objectives.

According to the descriptions in Subsection \ref{ps_scalarization_method}, it is worth mentioning that the setting of reference point and direction in the PS scalarization method is quite flexible. With the help of the flexibility of reference point and direction, \cite{EW2020} designed a new approximation algorithm to compute a box-coverage of the POF. We observe from their work that the reference points (the lower bounds of some boxes) and the directions (the differences between the lower bounds and the upper bounds of these boxes) are dynamically updated. Based on the flexibility of parameters in the PS scalarization method, the idea (i.e., ``\emph{multi-points and single-direction}") and the decomposition strategy \citep{ZL2007}, the primary motivation of this paper is to decompose a target MOP into a number of PS scalar optimization subproblems which have different reference points and same direction, and optimize them simultaneously in a collaborative manner. For example, in the right part of Fig. \ref{tch_ps}, there are nine decomposed PS scalar optimization subproblems defined the reference points depicted by the red solid points and the same direction depicted by the dotted line. From the graphical illustration in Subsection \ref{ps_scalarization_method}, the black solid points are the optimal solutions of all subproblems and they are uniformly distributed. Although this way can approximate continuous POF very well, it may also face the same dilemma as MOEA/D \citep{ZL2007}, that is, the performance degrades on MOPs with discontinuous POF or even degenerate POF. Therefore, in our proposed algorithm, the reference points related to the subproblems can be properly adjusted to obtain better performance in diversity. Given the above-mentioned descriptions, three inevitable questions are:
\begin{enumerate}[(1)]\setlength{\itemsep}{-0.03in}
	\item How can we embed the PS scalarization method into the MOEA/D framework?
	\item How to set the multi-reference points and the direction?
	\item How to adaptively adjust the multi-reference points?
\end{enumerate}

The following three subsections present the solutions for these questions.

\subsection{The transformation of PS scalarization method}

It is obvious that (\ref{ps}) cannot be easily used within the framework of decomposition-based multiobjective evolutionary algorithms since it introduces extra constraints. However, when $r_{i}>0$, $i\in\langle m\rangle$, (\ref{ps}) is equivalent to the following optimization problem:

\begin{equation}\label{ps1}
	\begin{aligned}
		\min&\quad t\\
		\text{s.t.}&\quad t\geq \frac{f_{i}(x)-a_{i}}{r_{i}},i\in\langle m\rangle,\\
		&\quad x\in \Omega, t\in R.
	\end{aligned}
\end{equation}
\noindent Note that, using a standard trick from mathematical programming, (\ref{ps1}) can be rewritten as the following minimax optimization problem:

\begin{equation}\label{ps2}
	\min\limits_{x\in\Omega}\quad g^{{\rm ps}}(x|a,r)=\max\limits_{1\leq i\leq m}\left\{\frac{1}{r_{i}}(f_{i}(x)-a_{i})\right\}.
\end{equation}
\noindent An advantage of (\ref{ps2}) is that it can be used as a scalarizing function in MOEA/D framework. Obviously, Theorem \ref{thm1} also holds for (\ref{ps2}).

\begin{remark}
	Notice that, when $m=2$ in (\ref{ps2}), if $a=b^{i}$ and $r=\gamma$, then (\ref{ps2}) reduces to the $i$-th NBI-style Tchebycheff subproblem (\ref{nbi_tch_sub}). In addition, some scalarization approaches used in MOEA/D framework are the special cases of (\ref{ps2}) by selecting suitable values for parameters (see Table \ref{special_case}).
	\begin{table}[htbp]\footnotesize
		\centering
		\caption{\footnotesize The special cases of (\ref{ps2}).}
		\begin{tabular}{ll}
			\hline
			Parameters in (\ref{ps2}) & Special cases \\
			\hline
			$a_{i}=z_{i}^{*}$, $r_{i}=\frac{1}{w_{i}}$, $i\in\langle m\rangle$ & TCH \\
			$a_{i}=z_{i}^{*}$, $r_{i}=w_{i}$, $i\in\langle m\rangle$ & m-TCH \\
			$a_{i}=z_{i}^{*}$, $r_{i}=\lambda_{i}$, $\|\lambda\|_{p}=1$ and $\lambda_{i}>0,i\in\langle m\rangle$ & $p$-TCH \\
			$a_{i}=z_{i}^{*}$, $r_{i}=(\sum_{i=1}^{m} \frac{1}{w_{i}}) / \frac{1}{w_{i}} $, $i\in\langle m\rangle$ & \cite{Q2014} \\
			\hline
		\end{tabular}
		\label{special_case}
	\end{table}
\end{remark}

\subsection{The setting of multi-reference points and direction}\label{sec3.3}
Our aim is to obtain a good approximation of the true POF for a given MOP by solving (\ref{ps2}) for some parameters. From Remark \ref{rem2.4}, all Pareto optimal solutions of (\ref{mop}) can be obtained for a fixed $r\in R_{+}^{m}\backslash\{0_{m}\}$ by varying $a\in R^{m}$. However, it seems unnecessary to choose the parameters $a$ in the whole space $R^{m}$. We observe that, Theorem 2.11 in \cite{E2008} points out that the parameter $a$ only need to be varied in a hyperplane. We now present the result as follows:
\begin{theorem}\label{thm2}
	Let $\hat{x}\in\Omega$ be a Pareto optimal solution of (\ref{mop}) and define a hyperplane $$\mathcal{H}=\{y\in R^{m}:\varsigma^{\top}y=\kappa\}$$
	with $\varsigma\in R^{m}\backslash\{0_{m}\}$ and $\kappa\in R$. Let $r\in R_{+}^{m}$ with $\varsigma^{\top}r\neq0$ be arbitrarily given. Then there exist a parameter $a\in\mathcal{H}$ and some $\hat{t}\in R$ such that $(\hat{x},\hat{t})$ is an optimal solution of  (\ref{ps}).
\end{theorem}

\begin{remark}
	\begin{enumerate}[(i)]\setlength{\itemsep}{-0.03in}
		\item Evidently, Theorem \ref{thm2} also holds for (\ref{ps2}).
		\item Theorem \ref{thm2} shows that it suffices to select the parameter $r\in R_{+}^{m}\backslash\{0_{m}\}$ as constant and to vary the parameter $a$ only in the hyperplane $\mathcal{H}$.
	\end{enumerate}
\end{remark}

\begin{figure}[htbp]
	\centering
	\includegraphics[width=10cm,height=4.15cm]{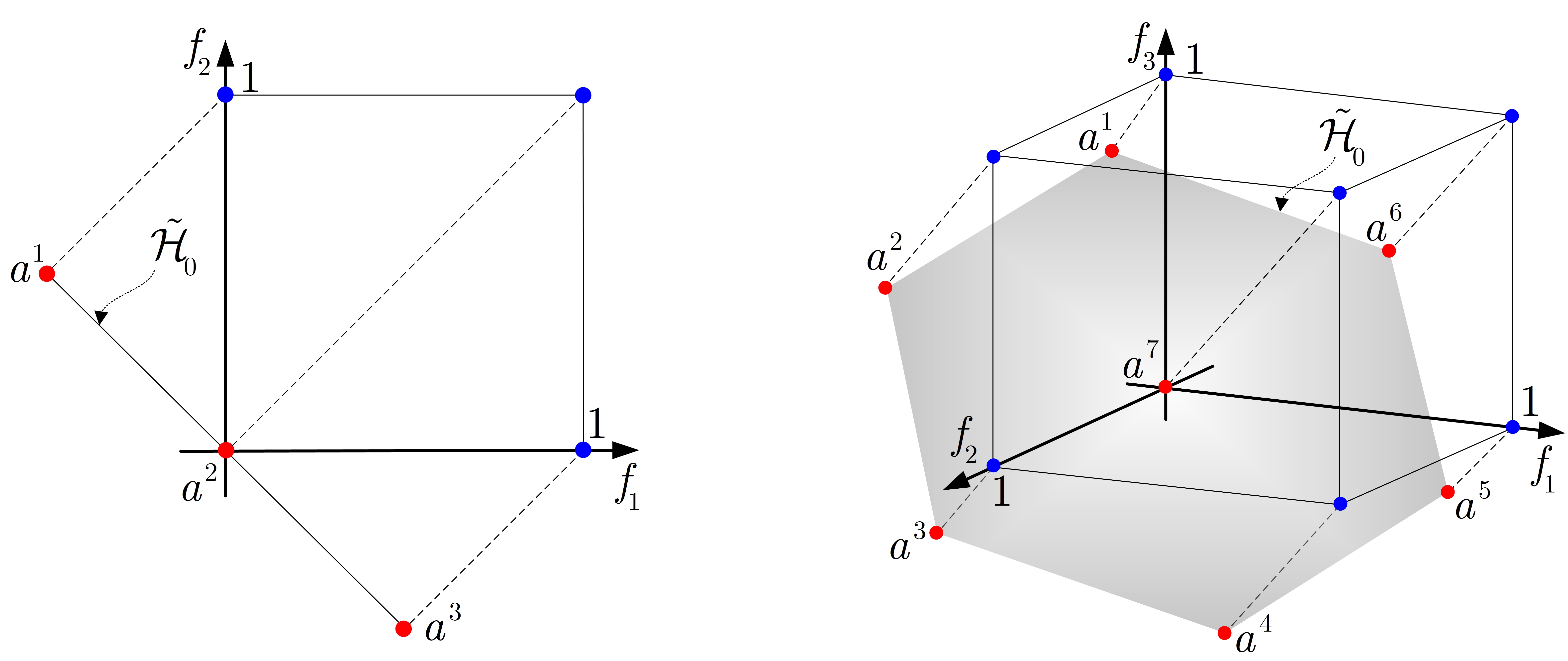}
	\caption{\footnotesize The illustration of the set $\mathcal{\tilde{H}}_{0}$ for $m=2$ and $m=3$.}
	\label{refer_points}
\end{figure}

In order to make the proposed algorithm having better performance in diversity, we can choose $r$ as the normal vector of the hyperplane $\mathcal{H}$ (i.e., $\varsigma=r$) and $\kappa=0$. More specifically, let $\varsigma=r=1_{m}$ and $\kappa=0$ (note that this is not the only option). In this case, $\mathcal{H}$ becomes the hyperplane

$$\mathcal{H}_{0}=\left\{y=(y_{1},\ldots,y_{m})\in R^{m}:\sum_{i=1}^{m}y_{i}=0\right\}$$
\noindent which passes through the origin. Then a set of reference points should be uniformly sampled on $\mathcal{H}_{0}$. In this way, different Pareto optimal solutions can be obtained by solving (\ref{ps2}). This way of sampling reference points on the entire $\mathcal{H}_{0}$ seems to be unwise, because it may lead to many redundant reference points. However, the range of true POFs of most test instances in the filed of evolutionary multiobjective optimization belong to $[0,1]^{m}$. Now, it is assumed that $P$ is the set of projection points of all vertices of the hypercube $[0,1]^{m}$ on $\mathcal{H}_{0}$ and let $\mathcal{\tilde{H}}_{0}$ be the convex hull of $P$. An example is shown in the left part of Fig. \ref{refer_points}, $P=\{a^{1},a^{2},a^{3}\}$ and

$$\mathcal{\tilde{H}}_{0}=\left\{y\in R^{2}:y=\lambda_{1}a^{1}+\lambda_{2}a^{3},\lambda_{1},\lambda_{2}\in[0,1]\right\}.$$
\noindent The right part of Fig. \ref{refer_points} is the case of $m=3$, i.e., $P=\{a^{1},a^{2},\ldots,a^{7}\}$ and

$$\mathcal{\tilde{H}}_{0}=\left\{y\in R^{3}:y=\sum_{i=1}^{6}\lambda_{i}a^{i},\lambda_{i}\in[0,1],i\in\langle 6\rangle\right\}.$$

\begin{figure}[H]
	\centering
	\includegraphics[width=8cm,height=4cm]{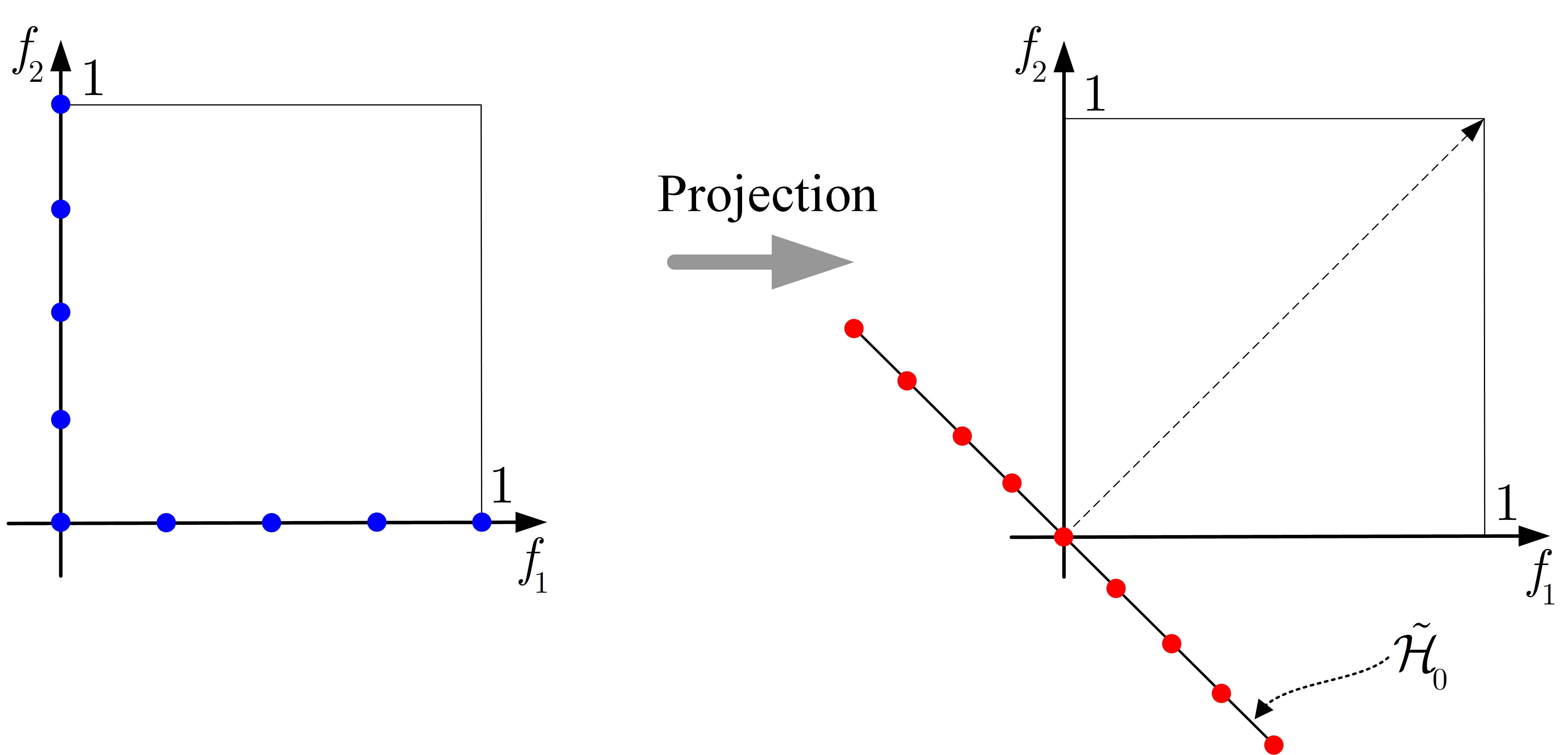}
	\caption{\footnotesize The generation process of reference points for $m=2$.}
	\label{two_refer_point}
\end{figure}

\begin{figure}[H]
	\centering
	\includegraphics[width=13.5cm,height=4cm]{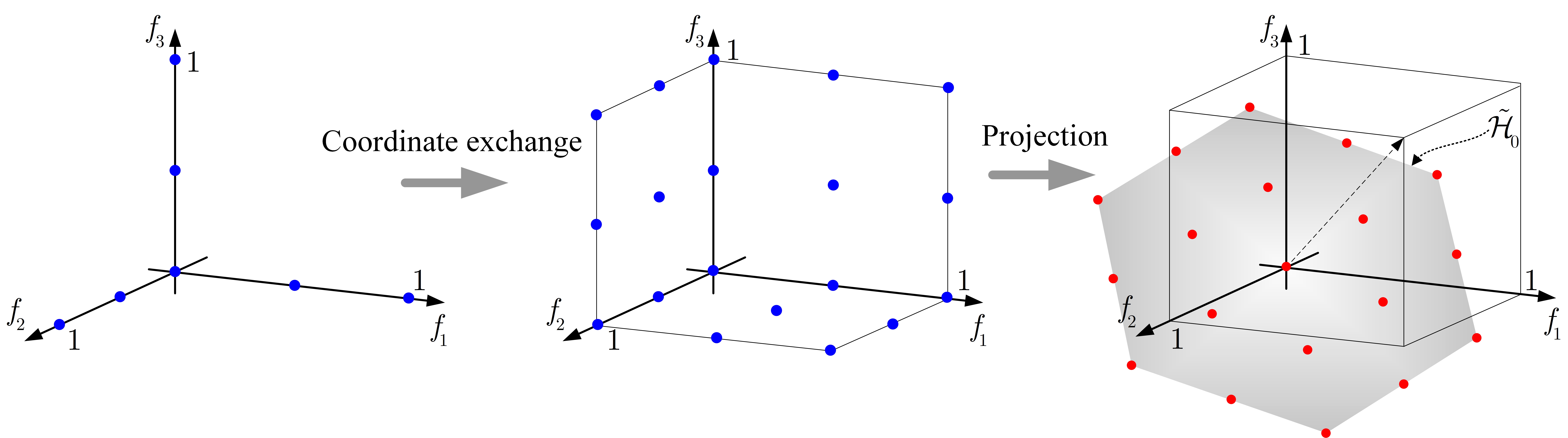}
	\caption{\footnotesize The generation process of reference points for $m=3$.}
	\label{three_refer_point}
\end{figure}

Therefore, we only need to sample a set of uniformly distributed reference points on $\mathcal{\tilde{H}}_{0}$. It should be noted that in this paper the uniformly distributed reference points on $\mathcal{\tilde{H}}_{0}$ are obtained by the following two steps:
\begin{enumerate}[(1)]
	\item \emph{Equidistant partition.} When $m=2$, we divide the interval $[0,1]$ on each axis into $l$ equal parts, and then $2l+1$ base points are obtained.
	The left part of Fig. \ref{two_refer_point} shows an example consisting of 9 base points marked by the blue solid points for $l=4$. When $m=3$, we also divide $[0,1]$ on each axis into $l$ equal parts. Then the coordinates of these base points are exchanged and $3l^{2}+3l+1$ new base points are obtained (see the blue solid points in the left and middle parts of Fig. \ref{three_refer_point}).
	\item \emph{Projection.}  The base points are projected into the hyperplane $\mathcal{H}_{0}$, and then these projection points are regarded as the reference points. Obviously, the convex hull formed by these reference points is $\mathcal{\tilde{H}}_{0}$. The right part of Fig. \ref{two_refer_point} shows the 9 uniformly distributed reference points for $l=4$ and the right part of  Fig. \ref{three_refer_point} shows an example consisting of 19 uniformly distributed reference points for $l=2$.
\end{enumerate}


We would like to emphasize that, if we consider the process of coordinate exchange for the case of $m=2$, then we can obtain the same reference points. Hence, we omit this process here. The aforementioned method can also be used for $m\geq 4$. In general, the range of true POFs of some test problems and real-world MOPs is not necessarily belong to $[0,1]^{m}$. It is apparent that the lines related to the reference points generated by the above-mentioned method cannot cover the whole POF. Herein, the normalization technique mentioned in \cite{ZL2007} is adopted to tackle this issue. After normalization, all solutions in current generation are normalized to points in $[0,1]^{m}$ in the normalized objective space. Therefore, the update of solutions can be performed by the following scalar optimization subproblem:

\begin{equation}\label{ps_norm}
	\min\limits_{x\in\Omega}\quad \tilde{g}^{{\rm ps}}(x|a,r,z^{*},z^{{\rm nad}})=\max\limits_{1\leq i\leq m}\left\{\frac{1}{r_{i}}\left(\frac{f_{i}(x)-z_{i}^{*}}{z_{i}^{{\rm nad}}-z_{i}^{*}}-a_{i}\right)\right\}.
\end{equation}

\begin{remark}
	It is noteworthy that the objective normalization, as  a transformation that maps the objective values from a scaled objective space onto the normalized objective space, changes the actual objective values, but does not affect the evaluation of $x$ by the scalarizing function value. Consequently, for $r=1_{m}$ and $a\in\mathcal{\tilde{H}}_{0}$, the optimal solutions between (\ref{ps2}) and (\ref{ps_norm}) are equivalent.
\end{remark}

\begin{remark}\label{rem3.4}
	If the true ideal point $z^{*}$ and the true nadir point $z^{{\rm nad}}$ in (\ref{ps_norm}) are not available, then we use the best value among all the examined solutions so far and worst value among the current population to assign values to $z^{*}$ and $z^{{\rm nad}}$, respectively.
\end{remark}

We conclude this subsection by giving a brief discussion between the way of generating weights in MOEA/D and the reference points generation technique in our method. In MOEA/D, a set of weights evenly distributed on $m$-dimensional unit simplex is used, which is usually generated by the lattice method introduced in \cite{DD1998}. However, in our proposed method, the reference points uniformly distributed on the set $\mathcal{\tilde{H}}_{0}$, which are generated by the techniques of equidistant partition and projection. The set $\mathcal{\tilde{H}}_{0}$ has a larger range than the simplex and can better cover the space $[0,1]^{m}$ from the diagonal perspective.

\subsection{The adaptation of multi-reference points}\label{sec3.4}
\begin{figure}[H]
	\centering
	\subfigure[Discontinuous]{
		\begin{minipage}[t]{0.3\linewidth}
			\centering
			\includegraphics[width=3.7cm,height=3.9cm]{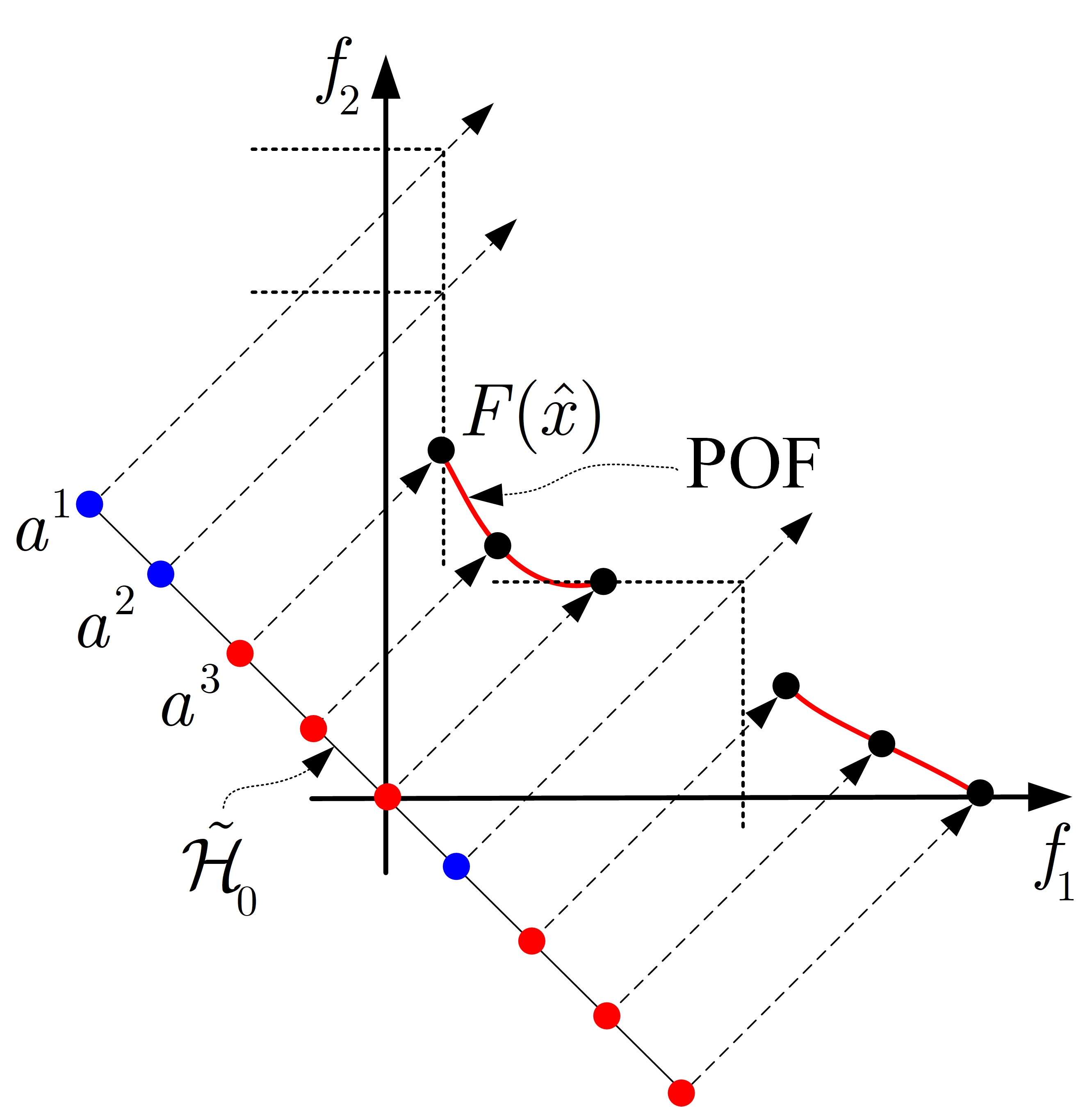}
		\end{minipage}%
	}%
	\subfigure[Simplex]{
		\begin{minipage}[t]{0.3\linewidth}
			\centering
			\includegraphics[width=4cm,height=4cm]{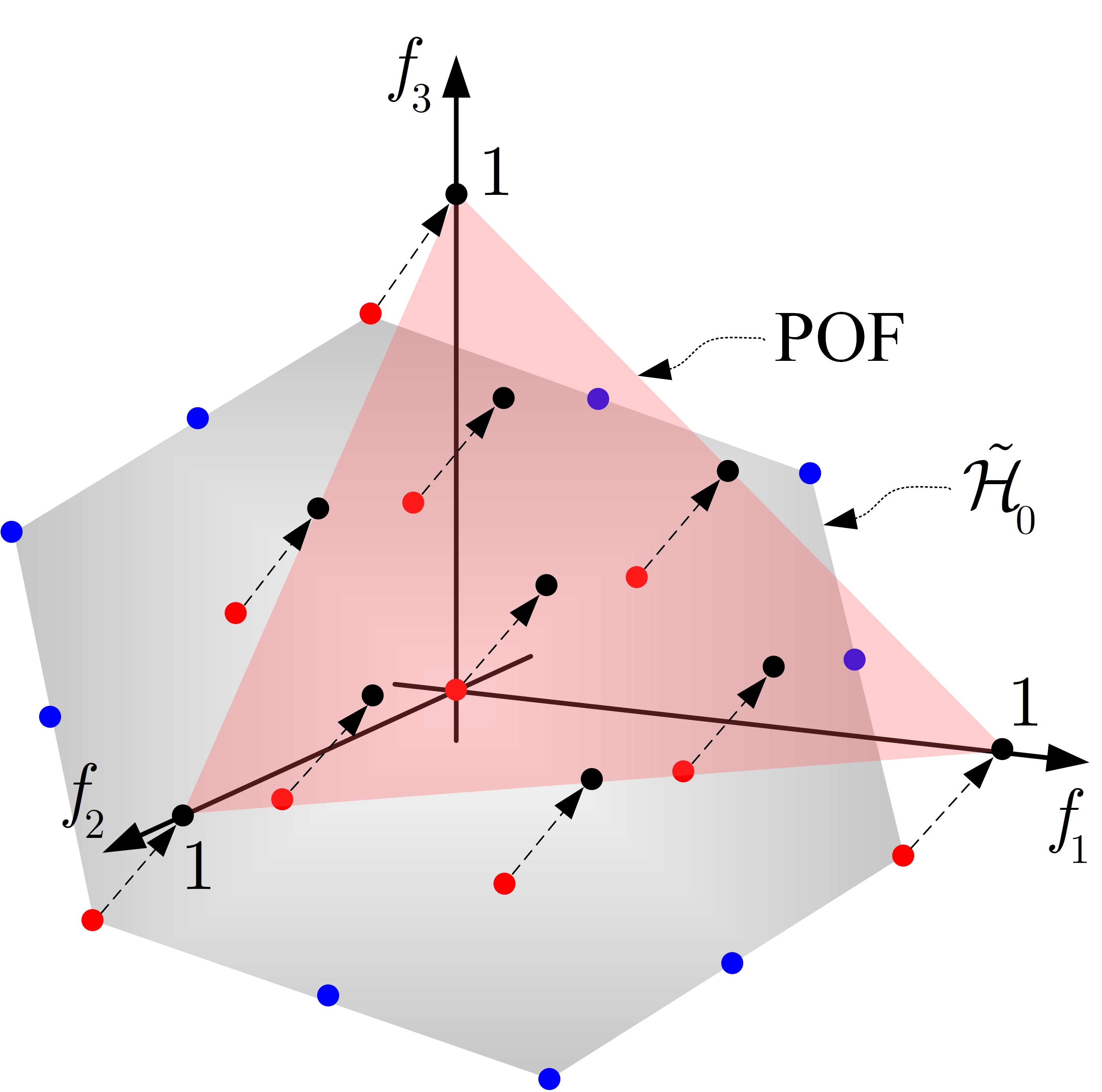}
		\end{minipage}%
	}%
	\subfigure[Degenerate]{
		\begin{minipage}[t]{0.3\linewidth}
			\centering
			\includegraphics[width=4cm,height=4cm]{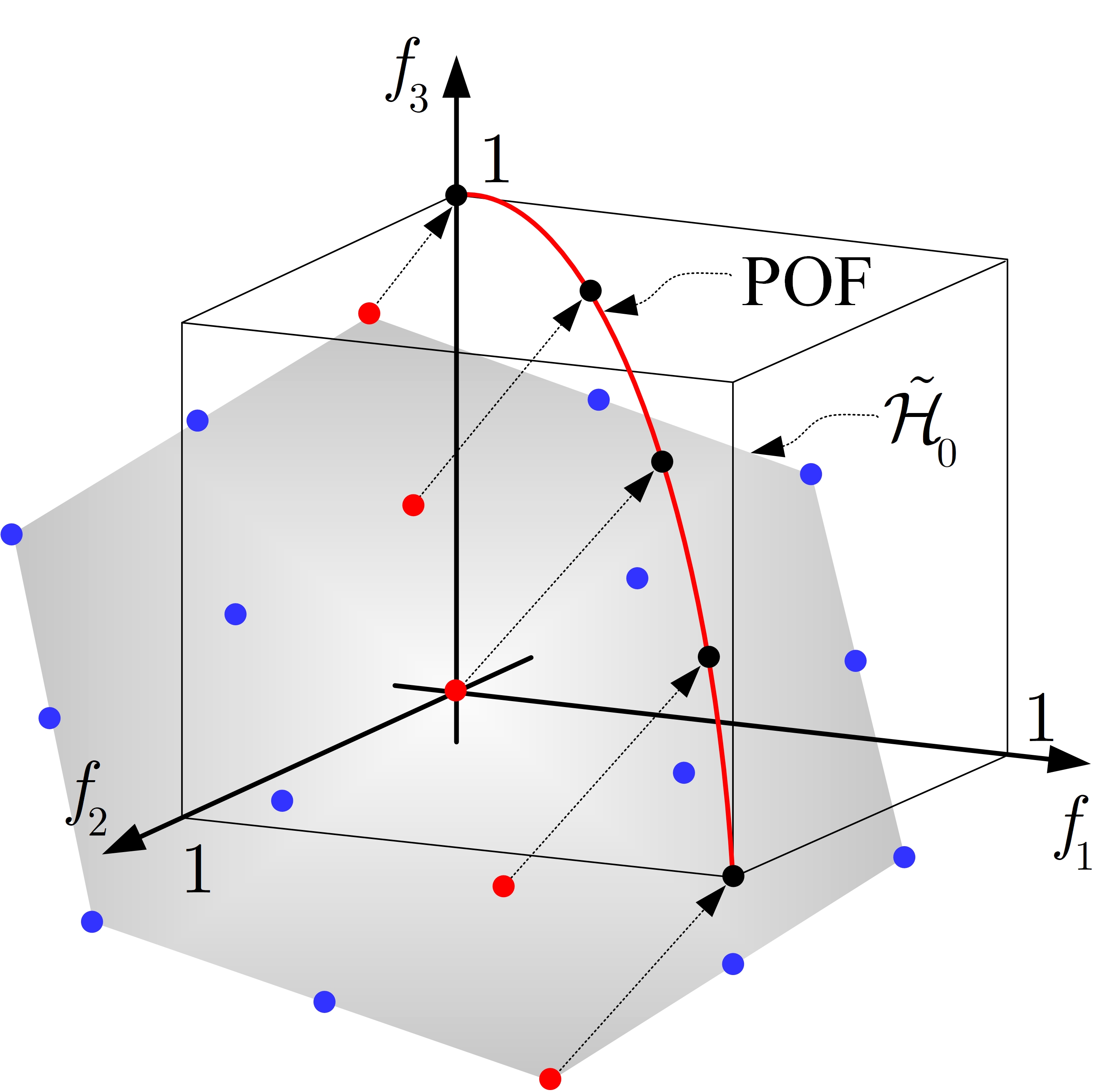}
		\end{minipage}%
	}%
	\caption{\footnotesize Illustrations the possible distribution of solutions for discontinuous, simplex and degenerate POFs. The black, red and blue solid points denote the solutions for subproblems, the promising reference points and the unpromising ones, respectively.}
	\label{valid _and_invalid}
\end{figure}

It is noteworthy that a set of equidistant parallel lines generated by the aforesaid method can cover the whole space $[0,1]^{m}$. However, if some straight lines, formed by the direction $r$ and some reference points, do not intersect the POF, then it follows from the geometric interpretation of the PS method in Subsection \ref{ps_scalarization_method} that the same solution may be obtained for some subproblems or many solutions concentrate on the boundary or discontinuous location of the true POF. Taking Fig. \ref{valid _and_invalid}(a) as an example, the subproblems related to the reference points $a^{1}$, $a^{2}$ and $a^{3}$ have the same Pareto optimal vector $F(\hat{x})$. This issue greatly affects the performance of the algorithm.

To improve algorithmic performance, it is imperative to consider when, where and how to adjust the reference points. Early adjustment of reference points could be unnecessary and ineffective because the population does not provide a good approximation for POF at early generation. A better approach would be to trigger the adjustment only when the population has roughly reached the POF in later generation. Herein, we use the evolutionary rate $\varepsilon$ to assist the reference points adaptation. When the ratio of iteration numbers to the maximal number of generations is $\varepsilon$, the adaptation of reference points will be started. In the context, a reference point is regarded as a promising one if it is associated with one or more solutions. On the other hand, a reference point is marked as unpromising if it is not associate with any solution, and then the marked unpromising reference point should be deleted. For example, for the discontinuous POF shown in Fig. \ref{valid _and_invalid}(a), the six red solid reference points are promising and three blue solid are unpromising. Similarly, for the simplex POF shown in Fig. \ref{valid _and_invalid}(b) and the degenerate POF shown in Fig. \ref{valid _and_invalid}(c), there also exist some promising and unpromising reference points. Additional reference points should be generated by utilizing the promising ones so as to keep a prespecified number of reference points. Overall, the key ingredients of adaptive multi-reference points adjustment are how to identify the promising reference points and to add the new ones. The specific implementation schemes will be discussed detailedly in Subsection \ref{ad_mul_ref}.

\subsection{The description of MOEA/D-AMR}
Based on the previous discussions, we are now in a position to propose a new algorithm called MOEA/D-AMR, which integrates the PS scalarization method and the adaptation of multi-reference points into the framework of MOEA/D-DE \citep{LZ2009}. Its pseudo-code is presented in Algorithm \ref{alg:Framwork}.
\vskip0.15in
\begin{algorithm}[H]\small
	\label{alg:Framwork}
	\caption{\texttt{MOEA/D-AMR Framework}}
	\LinesNumbered
	\KwIn{MOP (\ref{mop}), $l$: the number of division on each aixs, $T$: neighborhood size, $n^{{\rm rep}}$: replacement size, $G_{\max}$: the maximal number of generations.}
	\KwOut{Approximate POS: $X=\{x^{1},\ldots,x^{N}\}$, approximate POF: $F(X)=\{F(x^{1}),\ldots,F(x^{N})\}$.}
	$[L, B, X, F(X), z^{*}]\leftarrow$ \texttt{Initialization()}\; 
	gen $\leftarrow$ 1\;
	\While{{\rm gen} $\leq$ $G_{\max}$}{
		\For{$i\leftarrow1$ {\rm to} $N$}{
			\eIf{${\rm rand}[0,1]\leq\delta$}
			{
				$V\leftarrow B^{i}$\;
			}
			{
				$V\leftarrow \langle N\rangle$\;
			}
			Randomly select two indexes $v_{1}$ and $v_{2}$ from $V$\;
			$y\leftarrow$ \texttt{Reproduction-repair}($x^{i}, x^{v_{1}}, x^{v_{2}}$)\;
			Evaluate the function value of new solution $y$, update $z^{*}$ and $z^{{\rm nad}}$\;
			$c\leftarrow1$\;
			\While{$c\leq n^{{\rm rep}}${\rm and} $V\neq\emptyset$}{
				Randomly select an index $j$ from $V$, and $V\leftarrow V\backslash\{j\}$\;
				\If{$\tilde{g}^{{\rm ps}}(y|a^{j},r, z^{*}, z^{{\rm nad}})\leq \tilde{g}^{{\rm ps}}(x^{j}|a^{j},r,z^{*}, z^{{\rm nad}})$}{
					$x^{j}\leftarrow y$, $c\leftarrow c+1$\;
				}{
				}
			}
		}
		\If{${\rm gen}=\varepsilon G_{\max}$}
		{
			$[L^{{\rm pro}}, I^{{\rm pro}}, X^{{\rm pro}}] \leftarrow$ {\rm \texttt{Identifying-promising-reference-points}}$(L, X)$\;
			\eIf{$|L^{{\rm pro}}|=N$}
				{
					$L\leftarrow L^{{\rm pro}}$, $X\leftarrow X^{{\rm pro}}$\;
					Update $B$ using $L$\;
				}
			{$[L, X, B] \leftarrow$ {\rm \texttt{Adding-new-reference-points}}$(L^{{\rm pro}}, I^{{\rm pro}}, X^{{\rm pro}})$\;}
		}
		${\rm gen}\leftarrow {\rm gen}+1$\;
	}
\end{algorithm}
\vskip0.15in
Some important components of MOEA/D-AMR such as initialization (line 1 of Algorithm \ref{alg:Framwork}), reproduction and repair (lines 5--11 of Algorithm \ref{alg:Framwork}), update of solutions (lines 12--19 of Algorithm \ref{alg:Framwork}) and adjustment of multi-reference points (lines 21--29 of Algorithm \ref{alg:Framwork}) will be illustrated in detail in the following subsections.

\subsubsection{Initialization}

The first step is to create a set of $N$ reference points via the method introduced in Subsection \ref{sec3.3}. The set of all reference points is denoted by $L=\{a^{1},\ldots,a^{N}\}$. Secondly, we compute the Euclidean distance between any two reference points and then work out the $T$ closest reference points to each reference point. For each $i\in\langle N\rangle$, set the neighborhood index list $B^{i}=\{i_{1},\ldots,i_{T}\}$, where $a^{i_{1}},\ldots,a^{i_{T}}$ are the $T$ closest reference points to $a^{i}$. All the neighborhood index lists are defined as $B=\{B^{1},\ldots,B^{N}\}$. Thirdly, an initial population $X=\{x^{1},\ldots,x^{N}\}$, where $N$ is the population size, is generated by uniformly sampling from the decision space $\Omega$. The function values $F(x^{i})$ are calculated for every $x^{i}$, $i\in\langle N\rangle$ and let $F(X)=\{F(x^{1}),\ldots,F(x^{N})\}$. Finally, the ideal point $z^{*}=(z^{*}_{1},\ldots,z^{*}_{m})$ is initialized by setting $z_{j}^{*}=\min_{1\leq i\leq N}f_{j}(x^{i})$, $j\in\langle m\rangle$.

\subsubsection{Reproduction and repair}

A probability parameter $\delta$ is used to choose a mating pool $V$ from either the neighborhood of solutions or the whole population (lines 5--9 in Algorithm \ref{alg:Framwork}). The diffential evolution (DE) mutation operator and the polynomial mutation (PM) operator are used in this paper to produce an offspring solution from $x^{i}$, $i\in\langle N\rangle$, which are also considered in MOEA/D-DE \citep{LZ2009}. The DE operator generates a candidate solution $\bar{y}=(\bar{y}_{1},\ldots,\bar{y}_{n})$ by

\begin{equation*}
	\bar{y}_{k}=\left\{
	\begin{array}{ll}
		x_{k}^{i}+SF\times(x_{k}^{v_{1}}-x_{k}^{v_{2}}), & \hbox{if ${\rm rand}[0, 1]< CR$}, \\
		x_{k}^{i}, & \hbox{otherwise,}
	\end{array}
	\right.
\end{equation*}
\noindent where $v_{1},v_{2}$ are randomly selected from the mating pool $V$, $\bar{y}_{k}$ is the $k$-th component of $\bar{y}$, $k\in\langle n\rangle$, $SF$ is the scale factor, $CR$ is the cross rate and ${\rm rand}[0,1]$ is a uniform random number chosen from $[0,1]$. The PM operator is applied to generate a solution $y=(y_{1},\ldots,y_{m})$ from $\bar{y}$ in the following way:

\begin{equation*}
	y_{k}=\left\{
	\begin{array}{ll}
		\bar{y}_{k}+\sigma_{k}\times(u_{k}-l_{k}), & \hbox{if ${\rm rand}[0, 1]< p_{m}$}, \\
		\bar{y}_{k}, & \hbox{otherwise}
	\end{array}
	\right.
\end{equation*}
\noindent with

\begin{equation*}
	\sigma_{k}=\left\{
	\begin{array}{ll}
		(2\times {\rm rand}[0, 1])^{\frac{1}{1+\eta}}-1, & \hbox{if ${\rm rand}[0, 1]< 0.5$}, \\
		1- (2-2\times {\rm rand}[0, 1])^{\frac{1}{1+\eta}}, & \hbox{otherwise,}
	\end{array}
	\right.
\end{equation*}
\noindent where the distribution index $\eta>0$ and the mutation rate $p_{m}\in[0,1]$ are two control parameters. It is not always guaranteed that the new solution $y$ generated by reproduction belongs to the decision space $\Omega$. When a component of $y$ is out of the boundary of $\Omega$, a repair strategy is applied to $y$ such that $y\in\Omega$, i.e.,

\begin{equation*}
	y_{k}=\left\{
	\begin{array}{ll}
		l_{k}, & \hbox{if $y_{k}<l_{k}$}, \\
		u_{k}, & \hbox{if $y_{k}>u_{k}$,}\\
		y_{k}, & \hbox{otherwise.}\\
	\end{array}
	\right.
\end{equation*}

\subsubsection{Update of solutions}

After $y$ is generated, the procedure of updating solutions is performed, as shown in the lines 12--19 of Algorithm \ref{alg:Framwork}. First, the ideal point $z^{*}$ should be updated by $y$, i.e., for any $j\in\langle m\rangle$, if $z_{j}^{*}>f_{j}(y)$, then set $z_{j}^{*}=f_{j}(y)$ (line 12 of Algorithm \ref{alg:Framwork}). Then the nadir point $z^{{\rm nad}}$ is calculated by $z_{j}^{{\rm nad}}=\max_{1\leq i\leq N}f_{j}(x^{i})$ for each $j\in\langle m\rangle$. As mentioned in Remark \ref{rem3.4}, $z^{*}$ is the best value among all the examined solutions so far and $z^{{\rm nad}}$ is the worst value among the current population. Next, an index $j$ is randomly selected from $V$ and $j$ is subsequently deleted from $V$ (line 15 of Algorithm \ref{alg:Framwork}). Moreover, the individual $x^{j}$ is compared with the offspring $y$ based on the scalarization function $\tilde{g}^{{\rm ps}}$ defined in (\ref{ps_norm}). If $y$ is better than $x^{j}$ according to their scalarizing function values, then $x^{j}$ is replaced with $y$ (line 17 of Algorithm \ref{alg:Framwork}). $c$ is used to count the number of solutions replaced by $y$. If $c$ reaches the replacement size $n^{\rm rep}$, then the procedure of updating solutions terminates (line 14 of Algorithm \ref{alg:Framwork}).

\subsubsection{Adjustment of multi-reference points}\label{ad_mul_ref}
As analyzed in Subsection \ref{sec3.4}, we need to adjust the reference points adaptively in the later process of algorithm. Therefore, the evolutionary rate $\varepsilon$ is used to adaptively control evolutionary generations (line 21 of Algorithm \ref{alg:Framwork}). The adjustment strategy of multi-reference points consists of two parts:
\begin{enumerate}[(i)]\setlength{\itemsep}{-0.03in}
\item The identification of promising reference points (Algorithm \ref{proref});
\item The addition of new reference points (Algorithm \ref{newref}).
\end{enumerate}

For (i), we first need to give a distance criteria by virtue of the original reference points obtained by initialization process, i.e., the minimum distance $d^{L\to L}_{\min}$ between any two reference points in $L$ (lines 1--7 of Algorithm \ref{proref}). Secondly, the current population is normalized by the ideal point and the nadir point obtained by line 12 of Algorithm \ref{alg:Framwork}, and then project the normalized points onto the set $\mathcal{\tilde{H}}_{0}$ where the original reference points are located (line 9 of Algorithm \ref{proref}). Here, we denote the set of all obtained projection points as the set $Q=\{q^{1},\ldots,q^{N}\}$. Thirdly, we need obtain a distance matrix $D=(d^{L\to Q}_{ij})_{N\times N}$, where $d^{L\to Q}_{ij}$ stands for the distance between the $i$-th reference point in $L$ and the $j$-th projection point in $Q$ (lines 10--14 of Algorithm \ref{proref}). The next step is to find the minimum value of each row in matrix $D$ and to denote it as $d_{\min}^{L\to Q}(i)$, $i\in\langle N\rangle$. In other words, for each reference point $a^{i}$ in $L$, we need to find a projection point closest to $a^{i}$ from $Q$ and denote the distance between them as $d_{\min}^{L\to Q}(i)$. Finally, if $d_{\min}^{L\to Q}(i)$ is less than $d^{L\to L}_{\min}$, then the $i$-th reference point is recognized as a promising reference point and it is stored in a new set $L^{{\rm pro}}$, and its associated index and solution are preserved in $I^{{\rm pro}}$ and $X^{{\rm pro}}$, respectively (lines 16--20 of Algorithm \ref{proref}). To elaborate on the process, we explain it with an example.
\vskip0.15in
\begin{algorithm}[H]\small
	\label{proref}
	\caption{\texttt{Identifying-promising-reference-points}}
	\LinesNumbered
	\KwIn{$L=\{a^{1}, \ldots, a^{N}\}$: the original reference points, $X$: the current population.}
	\KwOut{The promising reference points set $L^{{\rm pro}}$, and its associated index set $I^{{\rm pro}}$ and the population $X^{{\rm pro}}$.}
	\For{$i\leftarrow 1$ {\rm to} $N$}
	{
		$d^{L\to L}_{ii}=+\infty$\;
		\For{$j\leftarrow 1$ {\rm to} $N$}
		{
			$d^{L\text{-}L}_{ij}=\|a^{i}-a^{j}\|$\;
		}
	}
	$d^{L\to L}_{\min}=\min_{1\leq i,j\leq N}d^{L\to L}_{ij}$\;
	Normalize the function values of the current population\;
	Project these normalized points into $\mathcal{H}_{0}$, and then obtain the set $Q=\{q^{1},\ldots,q^{N}\}$ of all the projection points\;
	\For{$i\leftarrow 1$ {\rm to} $N$}
	{
		\For{$j\leftarrow 1$ {\rm to} $N$}
		{
			$d^{L\to Q}_{ij}=\|a^{i}-q^{j}\|$\;
		}
	}
	$L^{{\rm pro}}=\emptyset$, $I^{{\rm pro}}=\emptyset$, $X^{{\rm pro}}=\emptyset$\;
	\For{$i\leftarrow 1$ {\rm to} $N$}
	{
		$d^{L\to Q}_{\min}(i)=\min_{1\leq j\leq N}d^{L\to Q}_{ij}$\;
		\If{$d^{L\to Q}_{\min}(i)\leq d^{L\to L}_{\min}$}
		{
			$I^{{\rm pro}}\leftarrow I^{{\rm pro}}\cup\{i\}$, $L^{{\rm pro}}\leftarrow L^{{\rm pro}}\cup\{a^{i}\}$, $X^{{\rm pro}}\leftarrow X^{{\rm pro}}\cup\{x^{i}\}$\;
		}
	}
\end{algorithm}
\vskip0.15in

\begin{figure}[H]
	\centering
	\includegraphics[width=11cm,height=4.9cm]{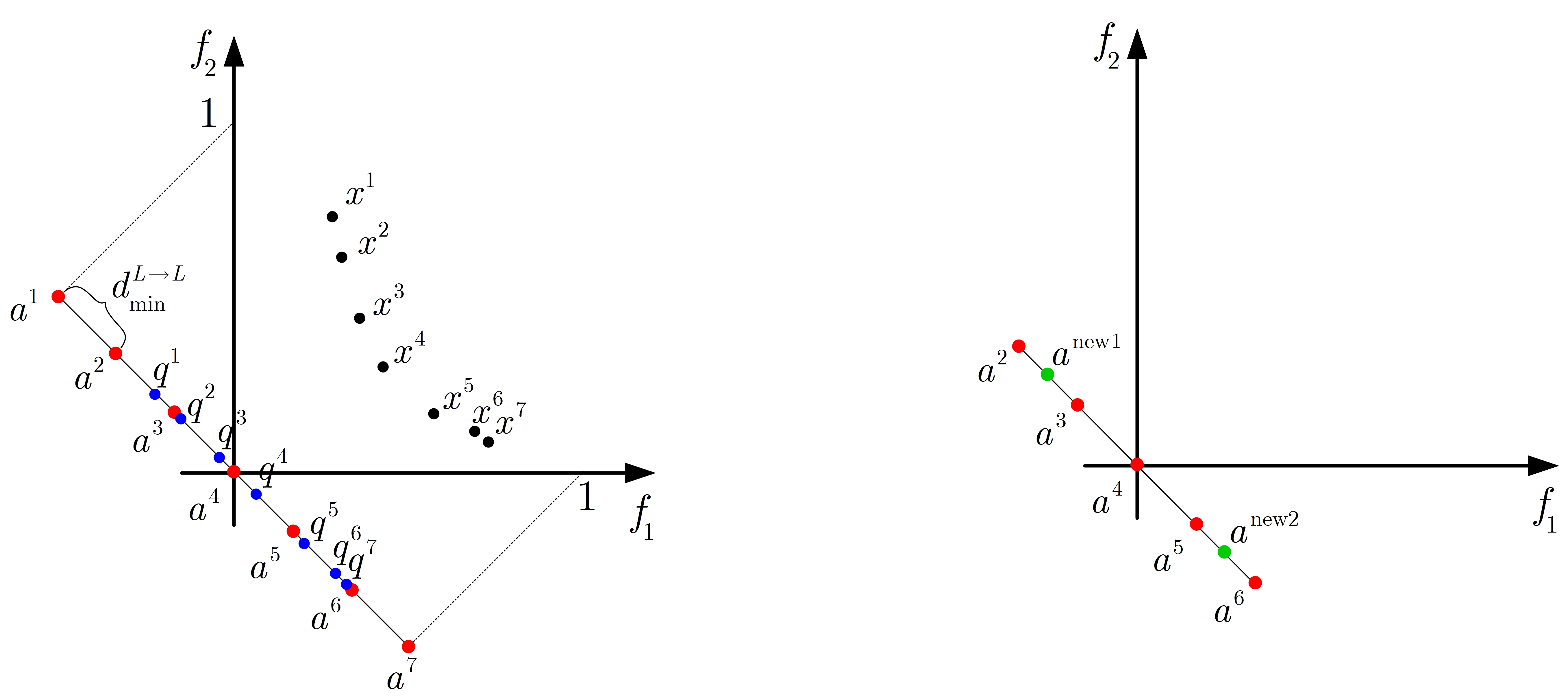}
	\caption{\footnotesize The identification of promising reference points for $m=2$.}
	\label{pro_refer_point}
\end{figure}

In the left part of Fig. \ref{pro_refer_point}, the red solid points $a^{1},\ldots,a^{7}$ stand for all original reference points, the black solid points $x^{1},\ldots,x^{7}$ represent all normalized points and the blue solid points $q^{1},\ldots,q^{7}$ denote all projection points. The closest points to $a^{1},a^{2},a^{3},a^{4},a^{5},a^{6},a^{7}$ are $q^{1},q^{1},q^{2},q^{3},q^{5},q^{7},q^{7}$, respectively. It is clear to see that $d_{\min}^{L\to Q}(1)=\|a^{1}-q^{1}\|> d_{\min}^{L\to L}$ and $d_{\min}^{L\to Q}(7)=\|a^{7}-q^{7}\|> d_{\min}^{L\to L}$. Therefore, the reference points $a^{1}$ and $a^{7}$ are regarded as unpromising ones and they are subsequently dropped. The other reference points are marked as promising ones.

If the cardinality of $L^{{\rm pro}}$ obtained by Algorithm \ref{proref} is equal to $N$, then all reference points are deemed as promising ones. Otherwise, some new reference points need to be generated by using the elements of $L^{{\rm pro}}$. Put another way, Algorithm \ref{newref} needs to be executed.
\vskip0.15in
\begin{algorithm}[H]\small
	\label{newref}
	\caption{\texttt{Adding-new-reference-points}}
	\LinesNumbered
	\KwIn{$L^{{\rm pro}}$: the promising reference points, $I^{{\rm pro}}$: the associated index set, $X^{{\rm pro}}$: the associated population.}
	\KwOut{The adjusted population $X$, the updated reference points $L$, the updated neighborhood index set $B$.}
	\For{$i\leftarrow 1$ {\rm to} $\langle N\rangle\setminus I^{{\rm pro}}$}
	{
		$X\leftarrow X^{{\rm pro}}\cup\{x^{i}\}$\;
	}
	Find any two adjacent reference points in $L^{{\rm pro}}$ to form a point pair, and then construct the point pairs set $\Lambda$ whose elements do not repeat each other\;
	\While{$|L^{{\rm pro}}|\leq N$}
	{
		\eIf{$N-|L^{{\rm pro}}|\leq |\Lambda|$}
		{
			Randomly select $N-|L^{{\rm pro}}|$ point pairs to form set $\bar{\Lambda}$ from $\Lambda$\;
			\For{$k\leftarrow1$ to $|\bar{\Lambda}|$}
			{
				$a^{{\rm new}}=\frac{a_{1}^{\Lambda(k)}+a_{2}^{\Lambda(k)}}{2}$\;
				$L^{{\rm pro}}\leftarrow L^{{\rm pro}}\cup\{a^{{\rm new}}\}$\;
			}
		}
		{
			\For{$k\leftarrow1$ to $|\Lambda|$}
			{
				$a^{{\rm new}}=\frac{a_{1}^{\Lambda(k)}+a_{2}^{\Lambda(k)}}{2}$\;
				$L^{{\rm pro}}\leftarrow L^{{\rm pro}}\cup\{a^{{\rm new}}\}$\;
			}
			Update the point pairs set $\Lambda$ using $L^{{\rm pro}}$ and Step 7\;
		}
	}
	$L\leftarrow L^{{\rm pro}}$\;
	Update $B$ using the updated reference points $L$\;
\end{algorithm}
\vskip0.15in
Therefore, for (ii), firstly, we need find any two adjacent reference points in $L^{{\rm pro}}$ to form a point pair, and the set of all point pairs is denoted by $\Lambda$ where the elements do not repeat (line 4 of Algorithm \ref{newref}). Next, we consider the following two cases:

\textbf{Case 1.} If $N-|L^{{\rm pro}}|$ is smaller than $|\Lambda|$, then we randomly choose $N-|L^{{\rm pro}}|$ elements from $\Lambda$ and use the midpoint of these selected elements to form new reference points (lines 7--11 of Algorithm \ref{newref}). For example, in the left part Fig. \ref{pro_refer_point}, we can obtain $L^{{\rm pro}}=\{a^{2},a^{3},a^{4},a^{5},a^{6}\}$ and the point pair set $\Lambda=\{(a^{2},a^{3}),(a^{3},a^{4}),(a^{4},a^{5}),(a^{5},a^{6})\}$. Obviously, $N-|L^{{\rm pro}}|<|\Lambda|$, then two elements are randomly selected from $\Lambda$ (we suppose that they are $(a^{2},a^{3})$ and $(a^{5},a^{6})$). Then $\bar{\Lambda}=\{(a^{2},a^{3}),(a^{5},a^{6})\}$. According to lines 9--10 of Algorithm \ref{newref}, new reference points $a^{{\rm new1}}=\frac{a^{2}+a^{3}}{2}$ and $a^{{\rm new2}}=\frac{a^{5}+a^{6}}{2}$ are added into $L^{{\rm pro}}$ (see the right part of Fig. \ref{pro_refer_point} and $a^{{\rm new1}},a^{{\rm new2}}$ are marked by the green solid points).

\begin{figure}[h]
	\centering
	\includegraphics[width=14cm,height=4.7cm]{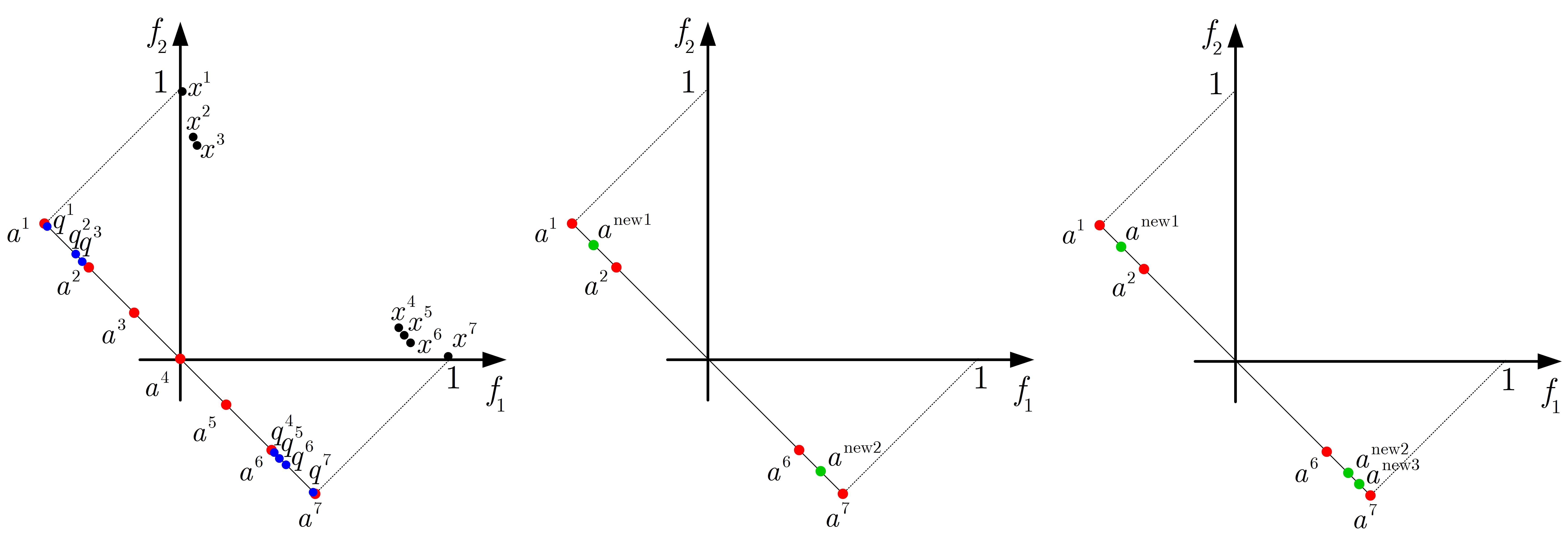}
	\caption{\footnotesize The identification of promising reference points and the addition of new reference point for $m=2$.}
	\label{pro_refer_point1}
\end{figure}

\textbf{Case 2.} If $N-|L^{{\rm pro}}|$ is strictly bigger than $|\Lambda|$, then all elements in $\Lambda$ need to be selected to generate new reference points, and then these newly generated points are added into $L^{{\rm pro}}$. Next, according to the obtained set $L^{{\rm pro}}$ and line 4 of Algorithm \ref{newref}, we reconstruct the point pair set $\Lambda$ and then repeat lines 13--17 until $|L^{{\rm pro}}|= N$. For example, similar to the above analysis, we can obtain $L^{{\rm pro}}=\{a^{1},a^{2},a^{6},a^{7}\}$ and the point pair set $\Lambda=\{(a^{1},a^{2}),(a^{6},a^{7})\}$ in the left part of Fig. \ref{pro_refer_point1}.

Clearly, $N-|L^{{\rm pro}}|>|\Lambda|$. At the first cycle, all elements in $\Lambda$ are selected to generate new reference points, i.e., $a^{{\rm new1}}=\frac{a^{1}+a^{2}}{2}$ and $a^{{\rm new2}}=\frac{a^{6}+a^{7}}{2}$. After the first cycle, the reference points set becomes $L^{{\rm pro}}=\{a^{1},a^{2},a^{6},a^{7},a^{{\rm new1}},a^{{\rm new2}}\}$ (see the middle part of Fig. \ref{pro_refer_point1}) and the point pair set becomes $\Lambda=\{(a^{1},a^{{\rm new1}}),(a^{{\rm new1}},a^{2}),(a^{6},a^{{\rm new2}}),(a^{{\rm new2}},a^{7})\}$. Obviously, $|L^{{\rm pro}}|<N$ and $N-|L^{{\rm pro}}|<|\Lambda|$. At the second cycle, only one element is randomly selected from $\Lambda$ and it is assumed that  $(a^{{\rm new2}},a^{7})$. Therefore, a new reference point $a^{{\rm new3}}=\frac{a^{{\rm new2}}+a^{7}}{2}$ is added into $L^{{\rm pro}}$ (see the right part of Fig. \ref{pro_refer_point1}).
\section{Experimental studies}\label{sec4}
This section is devoted to the experimental studies for the verification of the performance of the proposed algorithm. We compare it with four existing state-of-the-art algorithms: MOEA/D-DE \citep{LZ2009}, NSGA-III \citep{DJ2014}, RVEA* \citep{CJOS2016} and MOEA/D-PaS \citep{WZZ2016}. MOEA/D-DE possesses the TCH scalarization method and it is a representative steady-state decomposition-based MOEA. NSGA-III is an extension of NSGA-II, which maintains the diversity of population via decomposition. The main characteristic of RVEA* is the adaptive strategy of weight vectors. MOEA/D-PaS is equipped with $L_{p}$ scalarization method and it uses a Pareto adaptive scalarizing approximation to obtain the optimal $p$ value. Before presenting the results, we first give the experimental settings in the next subsection.

\subsection{Experimental settings}\label{sec4.1}

\begin{enumerate}[(1)]
	\item \emph{Benchmark problems.} In this paper, a set of benchmark test problems with a variety of representative POFs is used. Moreover, we also construct two modified test problems based on ZDT1 \citep{ZDT2000} and IDTLZ2 \citep{JD2014}. The first test instance denoted as F1 has the following form:
	
	\begin{equation*}
		\begin{aligned}
		&\text{min}\quad
		\left(\begin{array}{cc}
			g(x)\left(1-\frac{1}{1+e^{-10x_{1}}}\right)\\
			x_{1}
		\end{array}\right)\\
		&\text{s.t.}\quad\;\; x\in[-1,1]\times[0,1]^{n-1},
		\end{aligned}
	\end{equation*}
	\noindent where $g(x)=1+\frac{9\sum_{i=2}^{n}x_{i}}{n-1}$. The mathematical description of the second test problem is as follows:
	
	\begin{equation*}
		\begin{aligned}
			&\text{min}\quad
				\left(\begin{array}{cc}
				((1+g(x))-(1+g(x))\cos(0.5\pi x_1)\cos(0.5\pi x_2))^{1.8}\\
				((1+g(x))-(1+g(x))\cos(0.5\pi x_1)\sin(0.5\pi x_2))^{1.8}\\
				((1+g(x))-(1+g(x))\sin(0.5\pi x_2))^{1.8}\\
			\end{array}\right)\\
			&\text{s.t.}\quad\;\; x\in[0,1]^{n},
		\end{aligned}
	\end{equation*}
	where $g(x)=\sum_{i=3}^{n}(x_{i}-0.5)^{2}$. Note that the parameter in the variant of ZDT1 (termed as mZDT1) introduced in \cite{Q2014} is set as $M=0.5$ and the variable space of DTLZ1 is set as $[0.0001,0.9999]^{n}$. Other configures of all these problems are described in the corresponding literature. Table \ref{testproblem} provides a brief summary of these problems.
	
	\begin{table}[htbp]\scriptsize
		\centering
		\caption{\footnotesize Test Problems. $m$ and $n$ denote the number of objectives and decision variables, repectively.}
		\begin{tabular}{llllll}
			\hline
			$m$   & Problem                                                                                                                                    & $n$                                                                             & The POF shape                                                                                                                                                                                  &  &  \\ \hline
			$m=2$ & \begin{tabular}[c]{@{}l@{}}ZDT1 \citep{ZDT2000}\\ mZDT1 \citep{Q2014} \\ GLT3  \citep{GLT2012}\\ SCH1 \citep{VU1997}\\F1\\ZDT3  \citep{ZDT2000}\\ GLT1  \citep{GLT2012}\\\end{tabular}                                   & \begin{tabular}[c]{@{}l@{}}30\\30\\ 10\\1\\ 30\\ 30\\10\end{tabular}                   & \begin{tabular}[c]{@{}l@{}}Simplex-like, Convex\\ Highly nonlinear, Concave\\Highly nonlinear (piecewise linear), Convex\\ Highly nonlinear, Convex \\ Highly nonlinear, Nonconvex-nonconcave \\Disconnected \\ Disconnected  \end{tabular}                                                  &  &  \\ \hline
			$m=3$ & \begin{tabular}[c]{@{}l@{}}DTLZ1 \citep{DTLZ2005}\\ DTLZ2 \citep{DTLZ2005}\\ DTLZ5 \citep{DTLZ2005}\\VNT2 \citep{VFM1996}\\ DTLZ7 \citep{DTLZ2005}\\ IDTLZ1 \citep{JD2014} \\ IDTLZ2 \citep{JD2014}\\F2\\ \end{tabular} & \begin{tabular}[c]{@{}l@{}}7\\ 12\\12\\2\\ 15\\ 7\\ 12\\12\end{tabular} & \begin{tabular}[c]{@{}l@{}}Simplex-like, Linear\\ Simplex-like, Concave\\Degenerate, Concave\\ Degenerate, Convex\\ Disconnected\\ Inverted , Linear\\ Inverted, Concave\\Inverted, Highly concave \end{tabular} &  &  \\
			\hline
		\end{tabular}
		\label{testproblem}
	\end{table}
	\item  \emph{Parameter settings.} In MOEA/D-AMR, the number of division on each axis is $l=50$ for $m=2$ and $l=10$ for $m=3$. Therefore, we set the population size as $N=101$ and 331 for biobjective and triobjective MOPs, respectively. To make all algorithms comparable, the population size for the other four algorithms are the same as MOEA/D-AMR and the initial weights are kept the same for the four compared algorithms. The maximum number of generations of all algorithms is set as $G_{\max}=500$ on all the test problems. The other parameter settings of MOEA/D-AMR are listed as follows:
\begin{itemize}\setlength{\itemsep}{-0.03in}
\item the neighborhood size: $T=20$;
\item the probability of selecting parent from the neighborhood: $\delta=0.9$;
\item the replacement size: $n^{{\rm rep}}=2$;
\item the control parameters in DE operator: $SF=0.5$ and $CR=1$;
\item the parameters in PM operator: $p_{m}=1/n$ and $\eta=20$;
\item the evolutionary rate: $\varepsilon=0.8$.
\end{itemize}

The parameters $T$, $\delta$, $n^{{\rm rep}}$, $SF$, $CR$, $p_{m}$ and $\eta$ in MOEA/D-DE and MOEA/D-PaS share the same settings with the MOEA/D-AMR. The rate of change of penalty and the frequency of employing weight vector adaptation in RVEA* are set as $\alpha=2$ and $f_{r}=0.1$, respectively.

\item  \emph{Performance metrics.} Various performance metrics have been summarized in \cite{AB2020} for measure the quality of POF approximations. In this paper, two widely used performance metrics in MOEAs, the inverted generational distance (IGD) \citep{ZTL2003} and Hypervolume (HV) \citep{ZT1999}, are utilized to measure the obtained solution sets in terms of diversity. In the calculation of IGD, we select roughly 1000 scattered points evenly distributed in the true POF for all biobjective test problems and 5000 for the test problems with three objectives. All the objective function values are normalized by the ideal and nadir points of the POF before calculating HV metric. Then, the normalized HV value of the solution set is computed with a reference point $(1.1,1.1,\ldots,1.1)$. Every algorithm is run 30 times independently for each test problem. The mean and standard deviation values of the IGDs and HVs are calculated and listed in tables, where the results of best mean for each problem are highlighted with gray background. Furthermore, the Wilcoxon rank sum test with a significance level of 0.05 is adopted to perform statistics analysis on the experimental results, where the symbols '$+$', '$\approx$' and '$-$' denote that the result of other algorithms is significantly better, statistically similar and significantly worse to that of MOEA/D-AMR, respectively.
\end{enumerate}

\subsection{Experimental results and analysis}

The quantitative results on mean and standard deviation values of the performance indicators obtained by the five algorithms on these test instances are summarized in Tables \ref{igd} and \ref{hv}.

\begin{table}[H]\tiny
	\setlength\tabcolsep{4pt}
	\linespread{2.0}
	\centering
	\caption{\footnotesize Statistical results of IGD values (mean and standard deviation) found by different algorithms on test problems.}
	\begin{tabular}{lcccccc}
		\hline
		Property      & Problem                                                                & MOEA/D-DE                                                                                                                     & NSGA-III                                                                                                                      & RVEA*                                                                                                                         & MOEA/D-PaS                                                                                                                    & MOEA/D-AMR                                                                                                                     \\ \hline
		Simplex-like  & \begin{tabular}[c]{@{}l@{}}ZDT1\\ DTLZ1\\ DTLZ2\end{tabular}           & \begin{tabular}[c]{@{}l@{}}5.787e-3(3.62e-4) $-$\\ 1.469e-2(2.86e-5) $-$\\ 3.665e-2(1.29e-4) $-$\end{tabular}                       & \begin{tabular}[c]{@{}l@{}}4.845e-3(9.86e-6) $-$\\ \hl{1.025e-2(4.99e-6)} $+$\\ \hl{2.722e-2(6.02e-6)} $+$\end{tabular}                       & \begin{tabular}[c]{@{}l@{}}5.750e-3(4.23e-4) $-$\\ 1.076e-2(7.35e-5) $\approx$\\ 2.792e-2(1.03e-4) $\approx$\end{tabular}                       & \begin{tabular}[c]{@{}l@{}}5.257e-3(2.52e-4) $-$\\ 1.286e-2(1.33e-4) $\approx$\\ 3.227e-2(4.21e-4) $-$\end{tabular}                       & \begin{tabular}[c]{@{}l@{}} \hl{4.424e-3(3.40e-4)}\\ 1.163e-2(1.18e-4)\\ 3.070e-2(1.85e-4)\end{tabular}                       \\\hline
		Highly linear & \begin{tabular}[c]{@{}l@{}}mZDT1\\ GLT3\\ SCH1\\ F1\end{tabular}       & \begin{tabular}[c]{@{}l@{}}4.432e-3(5.18e-5) $\approx$\\ 2.265e-2(1.08e-2) $-$\\ 4.733e-2(1.18e-4) $-$\\ 3.736e-2(9.83e-5) $-$\end{tabular} & \begin{tabular}[c]{@{}l@{}}4.358e-3(9.57e-7)  $\approx$\\ 8.951e-2(2.84e-2) $-$\\ 4.724e-2(1.37e-4) $-$\\ 3.734e-2(1.02e-4) $-$\end{tabular} & \begin{tabular}[c]{@{}l@{}}4.310e-3(3.35e-5)  $\approx$\\ 1.062e-1(2.62e-2) $-$\\ 5.997e-2(8.37e-3) $-$\\ 3.135e-2(2.68e-3) $-$\end{tabular} & \begin{tabular}[c]{@{}l@{}}5.285e-3(7.26e-4)  $-$\\ 3.014e-2(5.15e-3) $-$\\ 4.468e-2(3.25e-3) $-$\\ 7.380e-2(2.57e-2) $-$\end{tabular} & \begin{tabular}[c]{@{}l@{}}\hl{4.050e-3(2.06e-6)}\\ \hl{4.776e-3(1.90e-5)}\\ \hl{1.698e-2(1.35e-3)}\\ \hl{1.446e-2(8.60e-3)}\end{tabular} \\\hline
		Degenerate    & \begin{tabular}[c]{@{}l@{}}DTLZ5\\ VNT2\end{tabular}                   & \begin{tabular}[c]{@{}l@{}}4.841e-3(2.06e-5) $-$\\ 2.262e-2(1.82e-4) $-$\end{tabular}                                             & \begin{tabular}[c]{@{}l@{}}3.547e-3(4.74e-4) $-$\\ 2.150e-2(8.83e-3) $-$\end{tabular}                                             & \begin{tabular}[c]{@{}l@{}}2.157e-3(7.27e-5) $\approx$\\ 1.168e-2(1.21e-3) $-$\end{tabular}                                             & \begin{tabular}[c]{@{}l@{}}6.955e-3(3.17e-4) $-$\\ 3.379e-2(4.52e-4) $-$\end{tabular}                                             & \begin{tabular}[c]{@{}l@{}}\hl{1.297e-3(1.82e-5)}\\ \hl{8.988e-3(1.44e-3)}\end{tabular}                                             \\\hline
		Inverted      & \begin{tabular}[c]{@{}l@{}}IDTLZ1\\ IDTLZ2\\F2\end{tabular} & \begin{tabular}[c]{@{}l@{}}1.875e-2(4.31e-5) $-$\\ 5.134e-2(1.58e-4) $-$\\1.131e-1(1.19e-3) $-$\end{tabular}                                             & \begin{tabular}[c]{@{}l@{}}1.460e-2(1.98e-4) $-$\\ 4.010e-2(9.73e-4) $-$\\6.901e-2(3.03e-3) $-$\end{tabular}                                             & \begin{tabular}[c]{@{}l@{}}1.174e-2(1.41e-4) $\approx$\\ 3.960e-2(1.00e-3) $-$\\7.025e-2(2.59e-3) $-$\end{tabular}                                             & \begin{tabular}[c]{@{}l@{}}2.413e-2(5.18e-4) $-$\\ 5.044e-2(8.56e-4) $-$\\1.176e-1(1.10e-3) $-$\end{tabular}                                             & \begin{tabular}[c]{@{}l@{}}\hl{1.169e-2(1.15e-4)}\\ \hl{2.902e-2(1.10e-4)}\\\hl{3.422e-2(1.39e-4)}\\\end{tabular}                                             \\\hline
		Disconnected  & \begin{tabular}[c]{@{}l@{}}ZDT3\\ GLT1\\ DTLZ7\end{tabular}            & \begin{tabular}[c]{@{}l@{}}7.498e-3(3.77e-4) $-$\\3.746e-3(4.74e-5) $-$ \\1.633e-1(1.41e-1) $-$\end{tabular}                                          & \begin{tabular}[c]{@{}l@{}}6.980e-3(5.14e-3) $-$\\1.287e-1(3.08e-2) $-$\\ 3.648e-2(7.70e-4) $\approx$\end{tabular}                                          & \begin{tabular}[c]{@{}l@{}}6.569e-3(4.17e-4) $-$\\ 1.358e-1(3.18e-2) $-$\\ 7.066e-2(1.05e-1) $-$\end{tabular}                                          & \begin{tabular}[c]{@{}l@{}}8.540e-3(4.53e-4) $-$\\ 3.910e-3(2.19e-4) $-$\\ 7.145e-1(1.16e-0) $-$\end{tabular}                                          & \begin{tabular}[c]{@{}l@{}}\hl{5.389e-3(2.79e-4)}\\\hl{2.134e-3(9.85e-5)}\\ \hl{3.505e-2(1.03e-3)}\end{tabular}                                          \\ \hline
		\multicolumn{2}{c}{$+/\approx/-$} &     0/1/14                                                    &     2/2/11    &   0/5/10    &    0/1/14       &           \\
		\hline
	\end{tabular}
	\label{igd}
\end{table}

\begin{table}[H]\tiny
	\setlength\tabcolsep{4pt}
	\centering
	\caption{\footnotesize Statistical results of HV values (mean and standard deviation) found by different algorithms on test problems.}
	\begin{tabular}{llccccc}
		\hline
		Property      & Problem                                                                & MOEA/D-DE                                                                                                                     & NSGA-III                                                                                                                      & RVEA*                                                                                                                         & MOEA/D-PaS                                                                                                                    & MOEA/D-AMR                                                                                                                     \\ \hline
		Simplex-like                    & \begin{tabular}[c]{@{}l@{}}ZDT1\\ DTLZ1\\ DTLZ2\end{tabular}     & \begin{tabular}[c]{@{}l@{}}9.270e-1(8.37e-4) $-$\\ 1.163e-0(9.26e-4) $-$\\ 8.966e-1(6.69e-4) $-$\end{tabular}                       & \begin{tabular}[c]{@{}l@{}}\hl{9.303e-1(4.71e-6)} $\approx$\\ \hl{1.190e-0(1.28e-4)} $+$\\ \hl{9.135e-1(1.67e-5)} $+$\end{tabular}                       & \begin{tabular}[c]{@{}l@{}}9.297e-1(2.79e-4) $\approx$\\ 1.188e-0(3.33e-4) $\approx$\\ 9.128e-1(3.87e-4) $+$\end{tabular}                       & \begin{tabular}[c]{@{}l@{}}9.294e-1(2.79e-4)  $\approx$\\ 1.181e-0(1.19e-3) $-$\\ 9.057e-1(5.13e-4) $-$\end{tabular}                       & \begin{tabular}[c]{@{}l@{}}9.282e-1(8.06e-4)\\ 1.185e-0(7.60e-4)\\ 9.097e-1(3.03e-4)\end{tabular}                       \\\hline
		Highly linear                   & \begin{tabular}[c]{@{}l@{}}mZDT1\\ GLT3\\ SCH1\\ F1\end{tabular} & \begin{tabular}[c]{@{}l@{}}5.173e-1(2.34e-4) $\approx$\\ 1.163e-0(1.57e-3) $-$\\ 1.068e-0(4.74e-6) $-$\\ 7.930e-1(7.14e-5) $-$\end{tabular} & \begin{tabular}[c]{@{}l@{}}5.179e-1(4.50e-6) $\approx$\\ 1.153e-0(4.29e-3) $-$\\ 1.068e-0(6.38e-6) $-$\\ 7.931e-1(6.31e-6) $-$\end{tabular} & \begin{tabular}[c]{@{}l@{}}5.180e-1(5.70e-5) $\approx$\\ 1.150e-0(3.38e-3) $-$\\ 1.067e-0(7.74e-4) $-$\\ 7.925e-1(3.13e-4) $-$\end{tabular} & \begin{tabular}[c]{@{}l@{}}5.174e-1(2.02e-4) $\approx$\\ 1.163e-0(9.43e-4) $-$\\ 1.068e-0(2.43e-4) $-$\\ 7.900e-1(2.05e-3) $-$\end{tabular} & \begin{tabular}[c]{@{}l@{}}\hl{5.185e-1(2.67e-5)}\\ \hl{1.167e-0(8.23e-5)}\\ \hl{1.069e-0(5.52e-5)}\\ \hl{7.944e-1(9.06e-5)}\end{tabular} \\\hline
		Degenerate                      & \begin{tabular}[c]{@{}l@{}}DTLZ5\\ VNT2\end{tabular}             & \begin{tabular}[c]{@{}l@{}}4.005e-1(1.51e-5) $-$\\ 5.301e-1(2.65e-5) $-$\end{tabular}                                             & \begin{tabular}[c]{@{}l@{}}4.007e-1(3.82e-4) $-$\\ 5.328e-1(2.22e-3) $-$\end{tabular}                                             & \begin{tabular}[c]{@{}l@{}}4.023e-1(4.90e-5) $\approx$\\ 5.313e-1(3.16e-4) $-$\end{tabular}                                             & \begin{tabular}[c]{@{}l@{}}3.981e-1(2.05e-4) $-$\\ 5.257e-1(1.57e-4) $-$\end{tabular}                                             & \begin{tabular}[c]{@{}l@{}}\hl{4.027e-1(3.59e-5)}\\ \hl{5.329e-1(6.09e-4)}\end{tabular}                                             \\\hline
		Inverted                        & \begin{tabular}[c]{@{}l@{}}IDTLZ1\\ IDTLZ2\\ F2\end{tabular}     & \begin{tabular}[c]{@{}l@{}}4.341e-1(1.24e-4) $-$\\ 8.250e-1(1.54e-4) $-$\\ 1.175e-0(3.32e-4) $-$\end{tabular}                       & \begin{tabular}[c]{@{}l@{}}4.343e-1(6.77e-4) $-$\\ 8.247e-1(8.82e-4) $-$\\ 1.182e-0(2.19e-3) $-$\end{tabular}                       & \begin{tabular}[c]{@{}l@{}}4.371e-1(7.63e-4) $\approx$\\ 8.337e-1(7.14e-4) $\approx$\\ 1.184e-0(1.44e-3) $-$\end{tabular}                       & \begin{tabular}[c]{@{}l@{}}4.248e-1(6.87e-4) $-$\\ 8.251e-1(7.76e-4) $-$\\ 1.175e-0(9.22e-4) $-$\end{tabular}                       & \begin{tabular}[c]{@{}l@{}}\hl{4.403e-1(8.18e-4)}\\ \hl{8.354e-1(2.89e-4)}\\ \hl{1.189e-0(1.64e-4)}\\                                                                                                                                                                                                                                                            \end{tabular}                       \\\hline
		Disconnected                    & \begin{tabular}[c]{@{}l@{}}ZDT3\\ GLT1\\ DTLZ7\end{tabular}      & \begin{tabular}[c]{@{}l@{}}8.101e-1(1.02e-3) $-$\\ 6.909e-1(1.04e-2) $+$\\ 4.692e-1(2.33e-2) $-$\end{tabular}                                         & \begin{tabular}[c]{@{}l@{}}\hl{8.119e-1(1.60e-2)} $+$\\4.492e-1(4.81e-2) $-$\\ 5.010e-1(4.26e-4) $\approx$\end{tabular}                                         & \begin{tabular}[c]{@{}l@{}}8.076e-1(1.44e-3) $-$\\ 4.392e-1(5.01e-2) $-$\\ 4.945e-1(2.23e-2) $\approx$\end{tabular}                                         & \begin{tabular}[c]{@{}l@{}}8.093e-1(5.24e-4) $-$\\  \hl{6.910e-1(1.52e-4)} $+$\\ 3.933e-1(1.59e-1) $-$\end{tabular}                       & \begin{tabular}[c]{@{}l@{}}8.095e-1(3.13e-4)\\ 6.892e-1(1.04e-3)\\ \hl{5.016e-1(3.17e-4)}\end{tabular}                       \\ \hline
		\multicolumn{2}{c}{$+/\approx/-$}  &         1/1/13                                             &   3/3/9       &   1/7/7    &    1/2/12        &           \\
		\hline
	\end{tabular}
	\label{hv}
\end{table}

\begin{figure}[H]
	\centering
	\includegraphics[width=\textwidth,height=3.8cm]{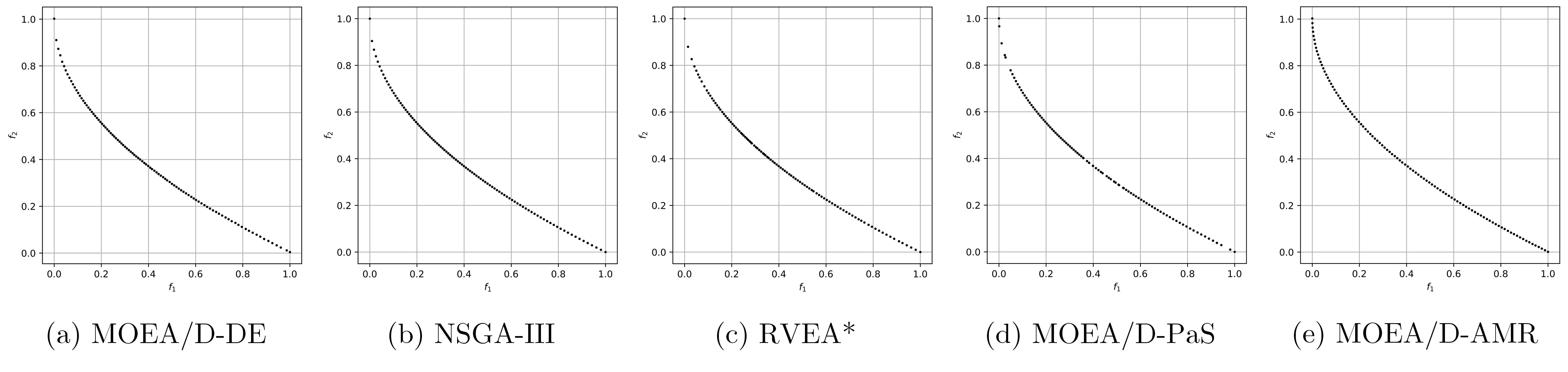}
	\caption{\footnotesize The final solution set with the median IGD value among 30 runs obtained by five algorithms on ZDT1.}
	\label{zdt1}
\end{figure}

In order to intuitively show the effectiveness of the proposed algorithm in diversity, the final solution sets with the median IGD values obtained by MOEA/D-DE, NSGA-III, RVEA*, MOEA/D-PaS and MOEA/D-AMR in 30 runs on these test problems are plotted. Since the POFs of these test problems have different properties, as stated in Table \ref{testproblem}, we discuss these results in the following five different groups:

\emph{1) Simplex-like POFs:} As can be observed from Fig. \ref{zdt1}, MOEA/AMR can obtain more evenly distributed solutions than the other four compared algorithms on ZDT1. It is apparently that the solutions obtained by other compared algorithms are sparse in the upper left part and are crowded in the middle part. However, the HV results indicated in Table \ref{hv} show that MOEA/D-AMR is slightly inferior than NSGA-II, RVEA* and MOEA/D-PaS. As mentioned by \cite{IIS2018}, the optimal distribution of solutions for hypervolume maximization may not be even. Note that Algorithm \ref{newref} on ZDT1 is not triggered in our experiments since all the reference points are promising, i.e., $|L^{{\rm pro}}|=N$ in line 23 of Algorithm \ref{alg:Framwork}.

\begin{figure}[H]
	\centering
	\includegraphics[width=\textwidth,height=3.5cm]{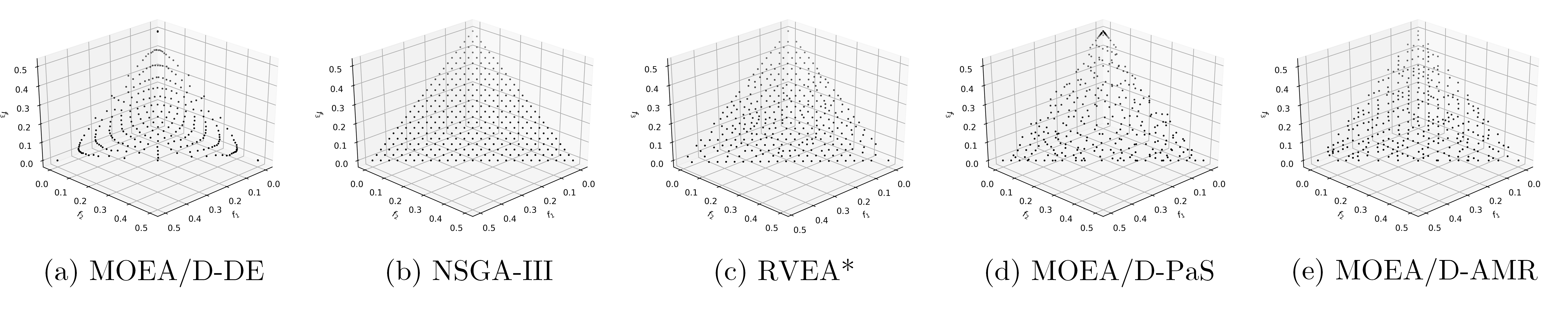}
	\caption{\footnotesize The final solution set with the median IGD value among 30 runs obtained by five algorithms on DTLZ1.}
	\label{dtlz1}
\end{figure}

\begin{figure}[H]
	\centering
	\includegraphics[width=\textwidth,height=3.5cm]{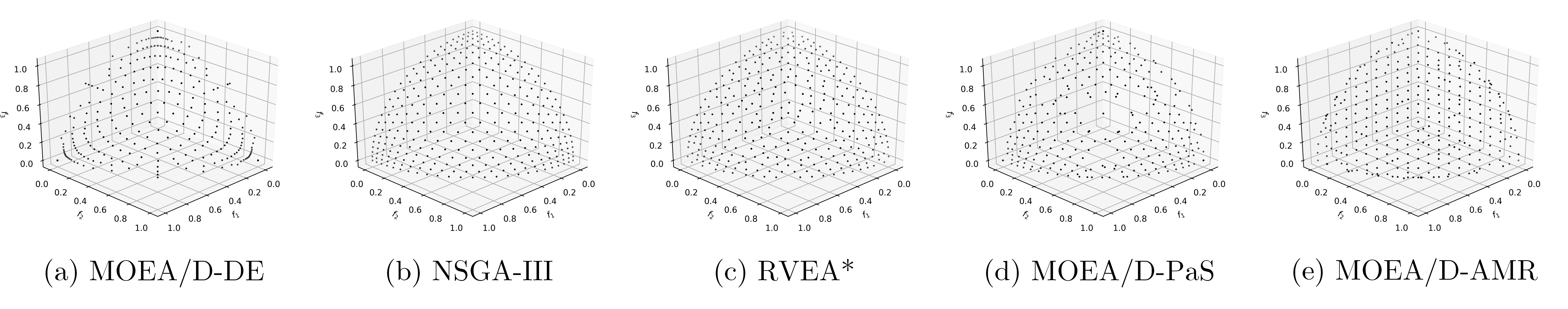}
	\caption{\footnotesize The final solution set with the median IGD value among 30 runs obtained by five algorithms on DTLZ2.}
	\label{dtlz2}
\end{figure}

For DTLZ1 and DTLZ2, Figs. \ref{dtlz1} and \ref{dtlz2} suggest that the performance of NSGA-III is better than the other algorithms. This is confirmed by looking at the statistical results displayed in Tables \ref{igd} and \ref{hv}. As described by \cite{DJ2014}, NSGA-III works very well on the two test problems. It is worth mentioning that MOEA/D-AMR is slightly worse than NSGA-III and RVEA* for DTLZ1 and DTLZ2. A reasonable explanation is that the creation of the additional reference points is based on the midpoint of two adjacent promising reference points. Therefore, many new reference points are embedded in the middle part of the original POF.

\begin{figure}[H]
	\centering
	\subfigure[\footnotesize DTLZ1]{
		\begin{minipage}[t]{0.3\linewidth}
			\centering
			\includegraphics[width=3.5cm,height=3cm]{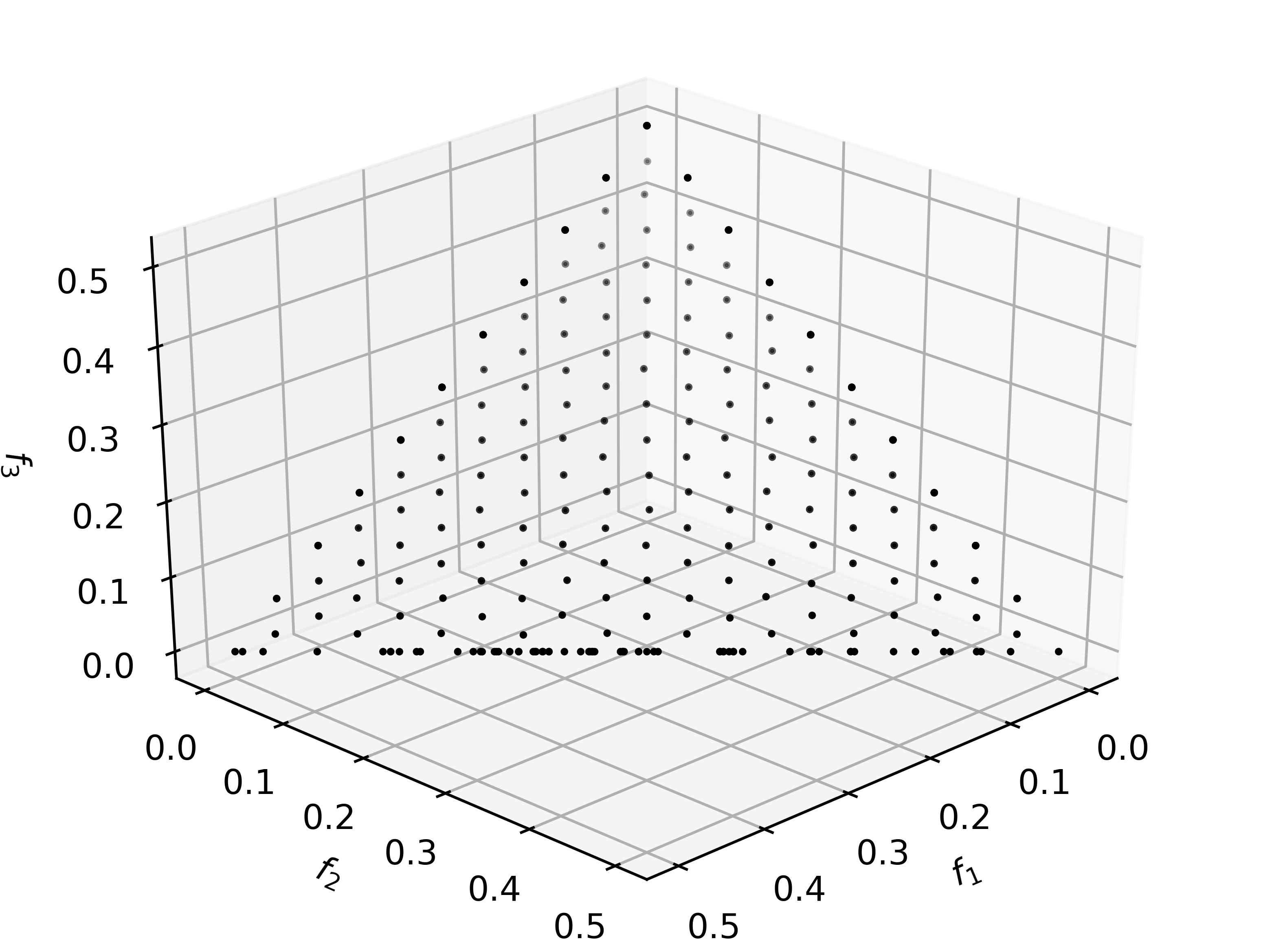}
		\end{minipage}
	}%
	\subfigure[\footnotesize DTLZ2]{
		\begin{minipage}[t]{0.3\linewidth}
			\centering
			\includegraphics[width=3.5cm,height=3cm]{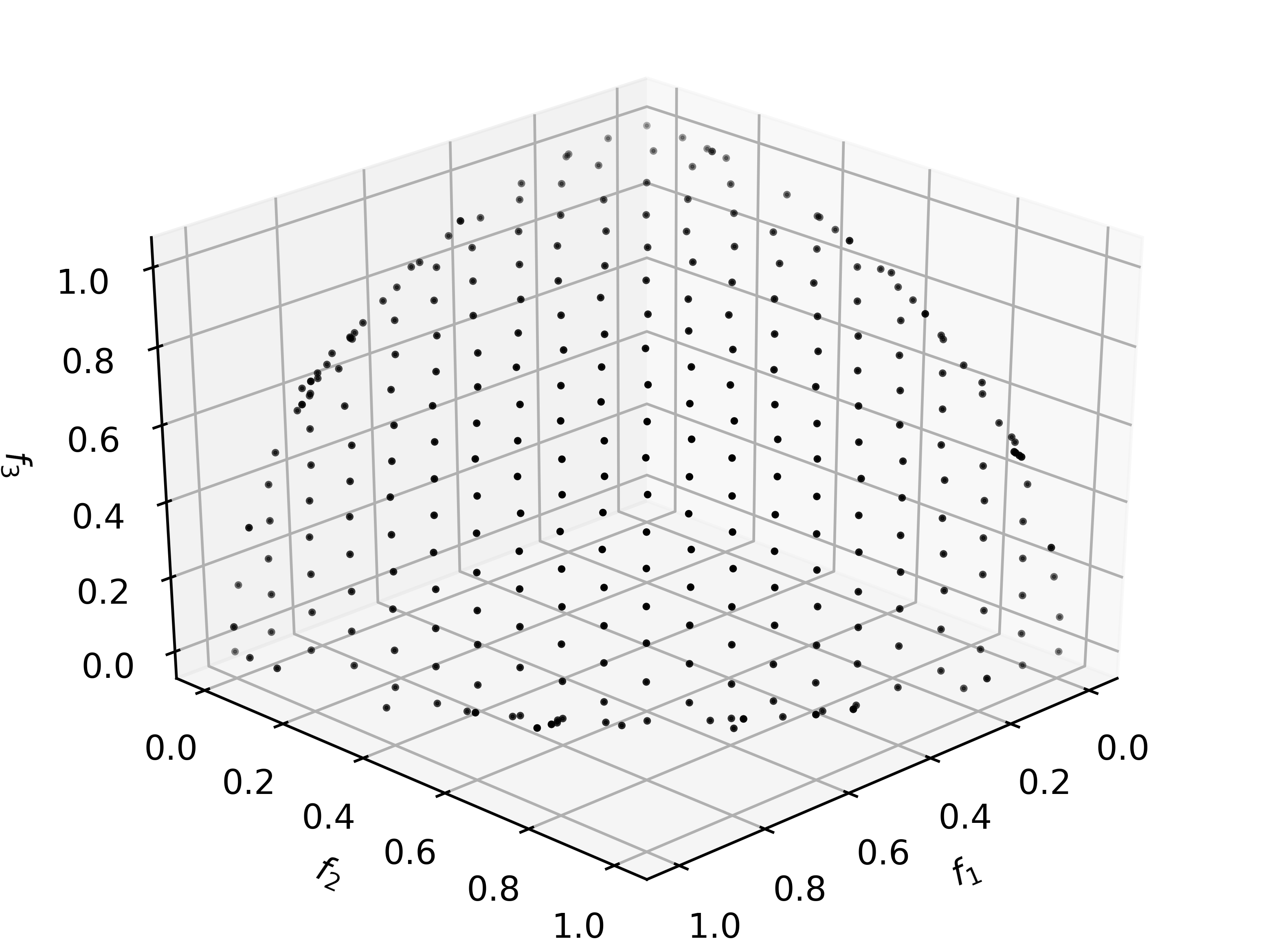}
		\end{minipage}
	}%
	\centering
	\caption{\footnotesize The final solution set obtained by MOEA/D-AMR on DTLZ1 and DTLZ2 without using the update strategy of reference points.}
	\label{DTLZ1_2}
\end{figure}

If we do not use the reference points update strategy in the proposed algorithm, as shown in Fig. \ref{DTLZ1_2}, the final solution set is very uniform in the inner part of the POFs of DTLZ1 and DTLZ2, but several solutions converge to other subproblems or concentrate on the boundary of the POFs.

\begin{figure}[H]
	\centering
	\includegraphics[width=\textwidth,height=3.8cm]{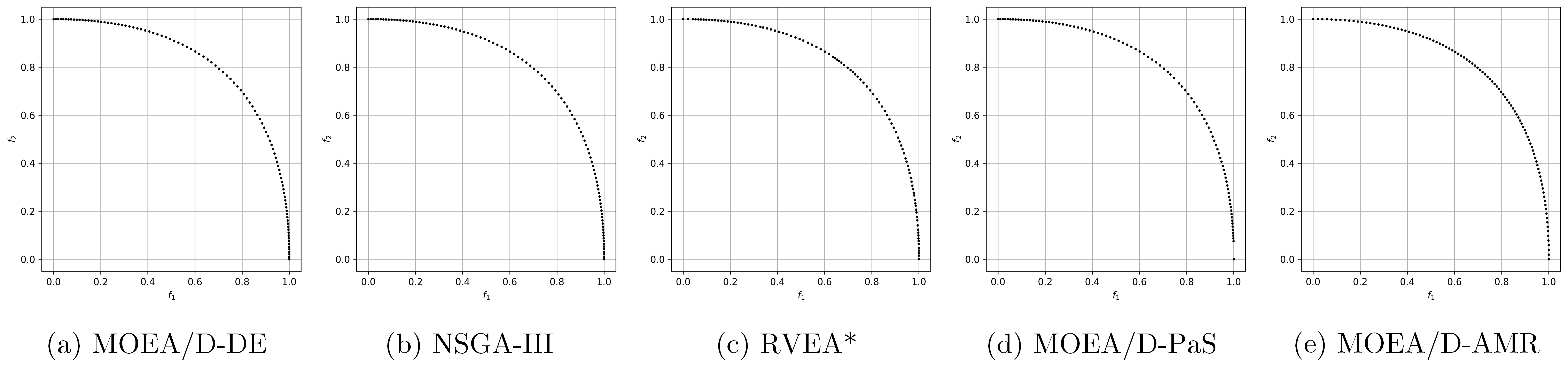}
	\caption{\footnotesize The final solution set with the median IGD value among 30 runs obtained by five algorithms on mZDT1.}
	\label{mzdt1}
\end{figure}

\begin{figure}[H]
	\centering
	\includegraphics[width=\textwidth,height=3.8cm]{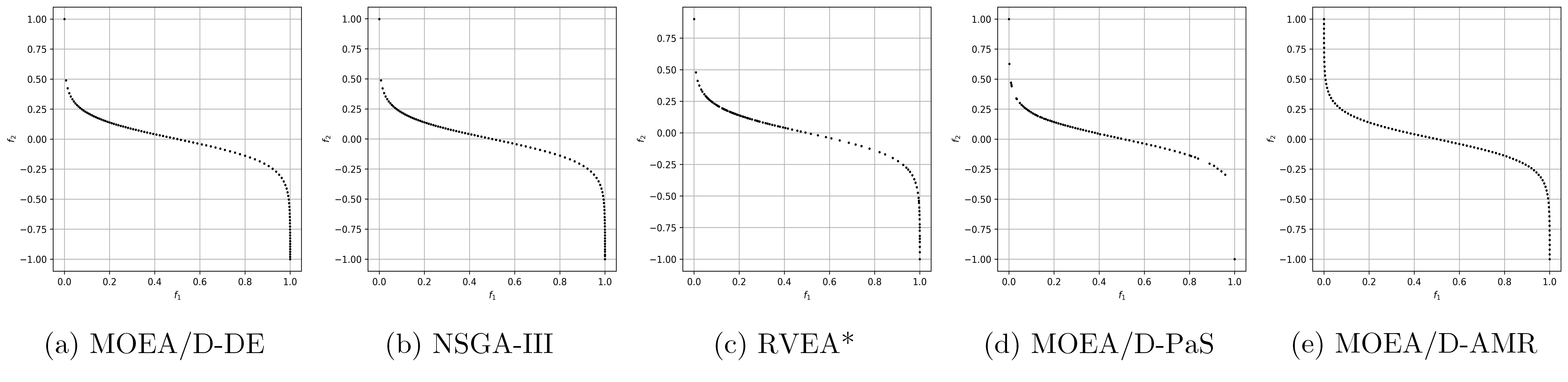}
	\caption{\footnotesize The final solution set with the median IGD value among 30 runs obtained by five algorithms on F1.}
	\label{sig}
\end{figure}

\emph{2) Highly nonlinear POFs:} The test problems with highly nonlinear POFs shown in Table \ref{testproblem} can be classified three groups: the problem with a concave POF (i.e., mZDT1), the problems with a convex POF (i.e., SCH1 and GLT3) and the problem with a nonconvex-nonconcave POF (i.e., F1). For the first group, Fig. \ref{mzdt1} shows that all the algorithms work well while MOEA/D-AMR performing slightly better than the other four compared algorithms. As for the second group, it is clear that only the proposed algorithm has an excellent performance in diversity and other four algorithms struggle in the location of sharp peak and long tail. For the third group, it follows from Fig. \ref{sig} that only MOEA/D-AMR can obtain a good performance in diversity. Overall, the highly nonlinear convex POF still poses a challenge to the corresponding method even if some adaptive strategies of weight vectors and scalarizing function are proposed.
\begin{figure}[H]
	\centering
	\includegraphics[width=\textwidth,height=3.8cm]{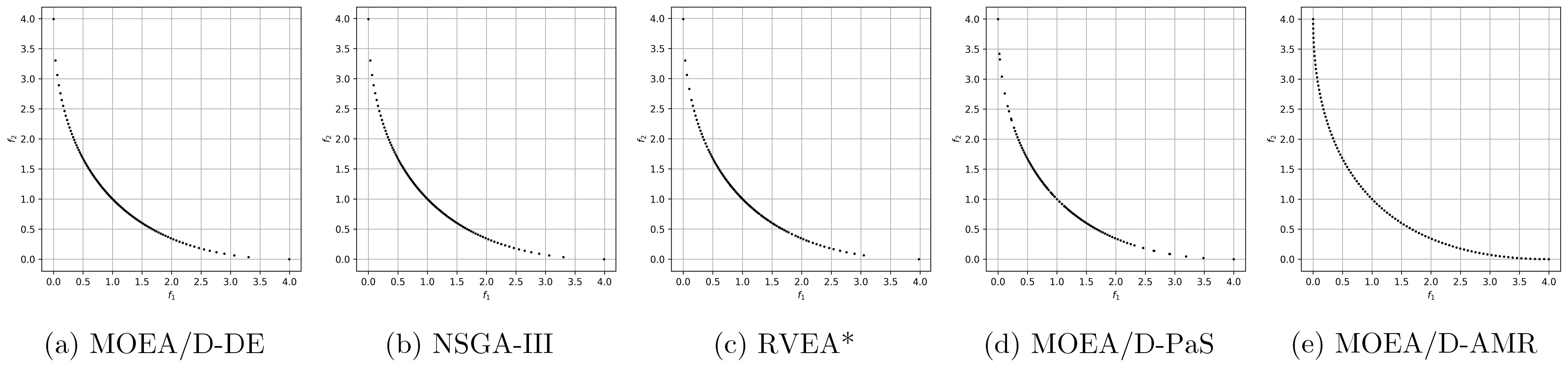}
	\caption{\footnotesize The final solution set  with the median IGD value among 30 runs obtained by five algorithms on SCH1.}
	\label{sch1}
\end{figure}

\begin{figure}[H]
	\centering
	\includegraphics[width=\textwidth,height=3.8cm]{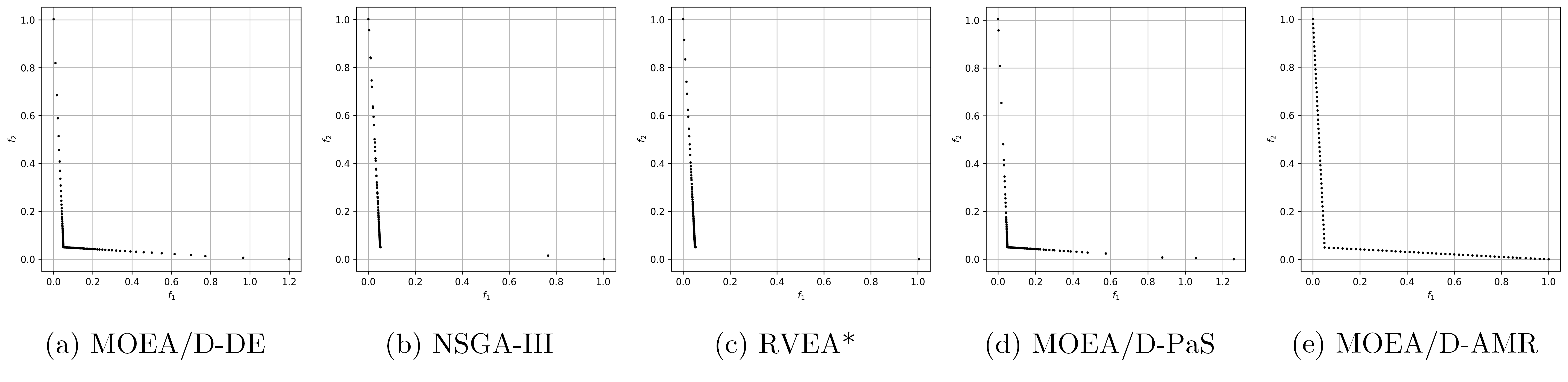}
	\caption{\footnotesize The final solution set with the median IGD value among 30 runs obtained by five algorithms on GLT3.}
	\label{glt3}
\end{figure}

\begin{figure}[H]
	\centering
	\includegraphics[width=\textwidth,height=3.5cm]{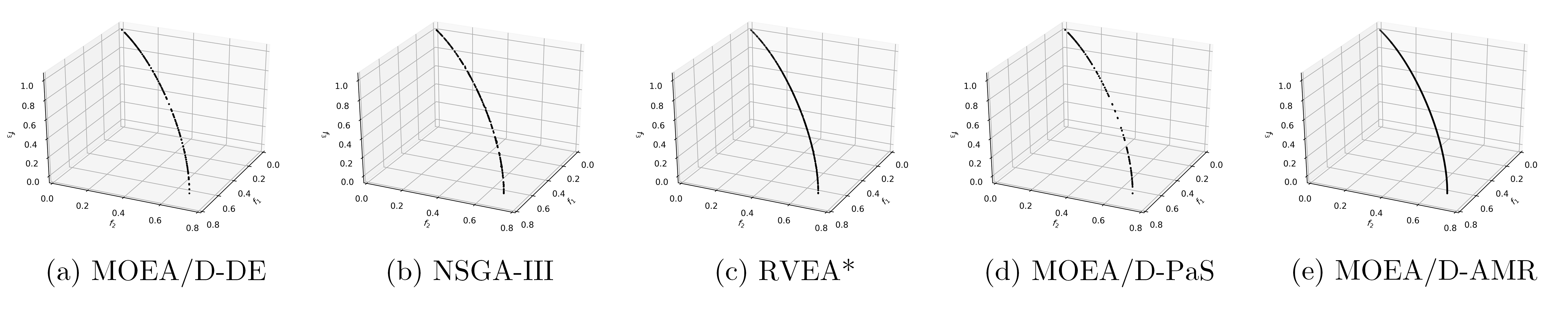}
	\caption{\footnotesize The final solution set with the median IGD value among 30 runs obtained by five algorithms on DTLZ5.}
	\label{dtlz5}
\end{figure}

\begin{figure}[H]
	\centering
	\includegraphics[width=\textwidth,height=3.5cm]{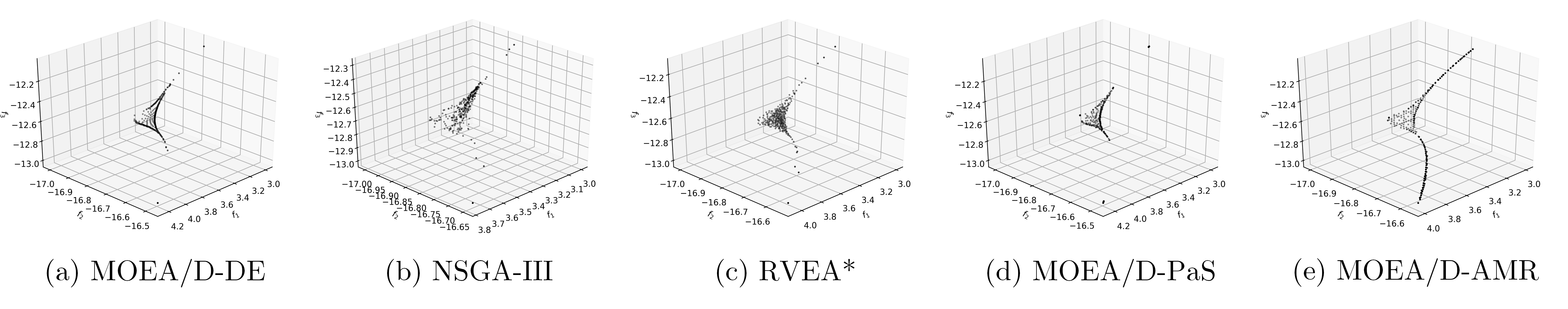}
	\caption{\footnotesize The final solution set with the median IGD value among 30 runs obtained by five algorithms on VNT2.}
	\label{vnt2}
\end{figure}

\emph{3) Degenerated POFs:} It follows from Tables \ref{igd} and \ref{hv} and Figs. \ref{dtlz5} and \ref{vnt2} that the proposed algorithm has shown a significant advantage over its competitors on this group of problems. RVEA* performs better than the rest three algorithms as it has a weight vectors regeneration strategy. For this kind of problems, MOEA/D-PaS equipped with the adaptation strategy of scalarization function seems not so effective.

\begin{figure}[H]
	\centering
	\includegraphics[width=\textwidth,height=3.5cm]{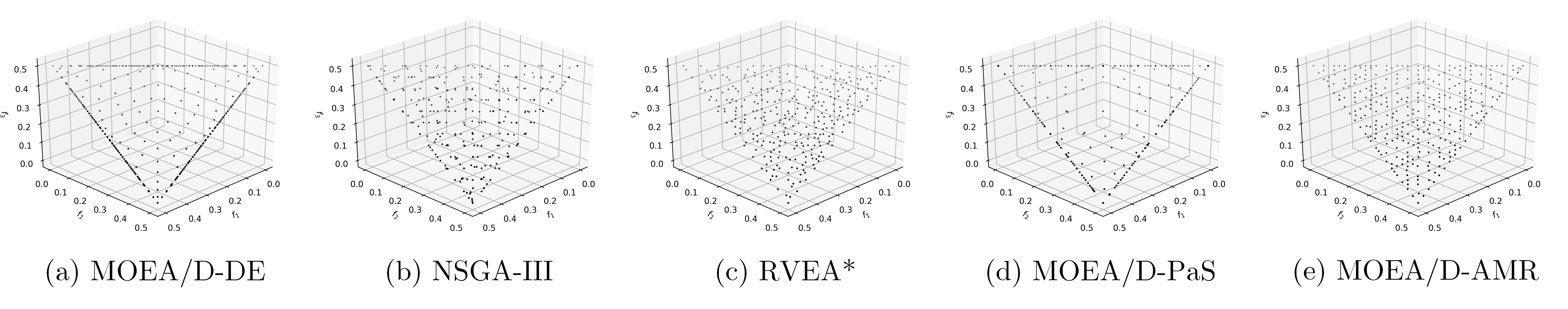}
	\caption{\footnotesize The final solution set with the median IGD value among 30 runs obtained by five algorithms on IDTLZ1.}
	\label{idtlz1}
\end{figure}

\begin{figure}[H]
	\centering
	\includegraphics[width=\textwidth,height=3.5cm]{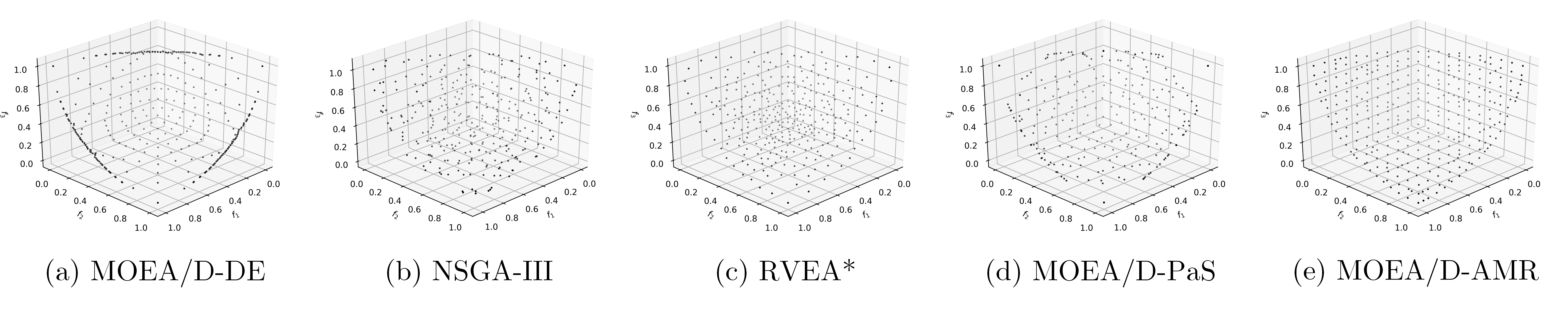}
	\caption{\footnotesize The final solution set with the median IGD value among 30 runs obtained by five algorithms on IDTLZ2.}
	\label{idtlz2}
\end{figure}

\begin{figure}[H]
	\centering
	\includegraphics[width=\textwidth,height=3.5cm]{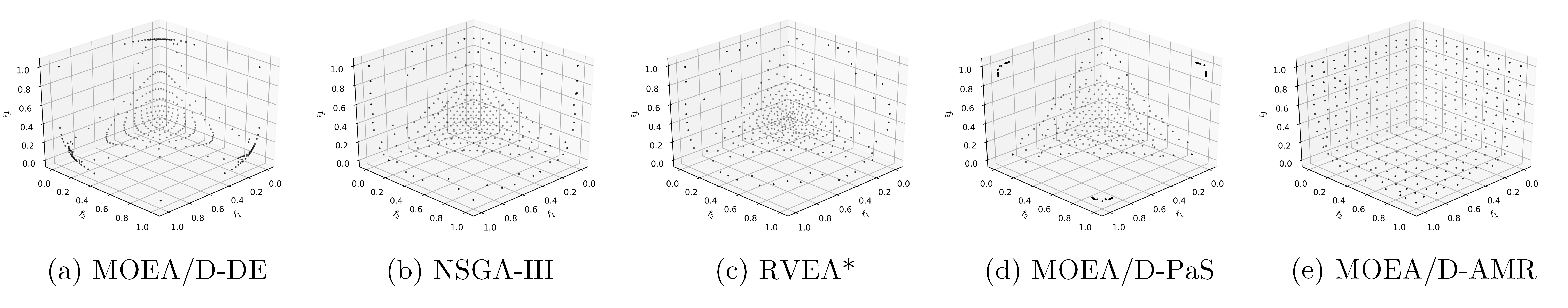}
	\caption{\footnotesize The final solution set with the median IGD value among 30 runs obtained by five algorithms on F2.}
	\label{f2}
\end{figure}

\emph{4) Inverted POFs:} For MOEA/D-AMR, Figs. \ref{idtlz1}--\ref{f2} illustrate that this group of problems has no effect on the algorithmic performance and the obtained solution set has a good coverage and diversity over the entire POF. For IDTLZ1 and IDTLZ2, a small amount of solutions in the inner part of POFs obtained by MOEA/D-DE and MOEA/PaS have a little uniformity, while most of the solutions locate in the boundary part of the POFs. By contrast, the solutions obtained by NSGA-III and RVEA* achieve a good coverage and poor diversity. On the constructed problem F2, the algorithms, i.e., MOEA/DE, NSGA-III, RVEA* and MOEA/D-PaS, are not very satisfactory and effective in coverage and diversity preservation.

\begin{figure}[H]
	\centering
	\includegraphics[width=\textwidth,height=3.8cm]{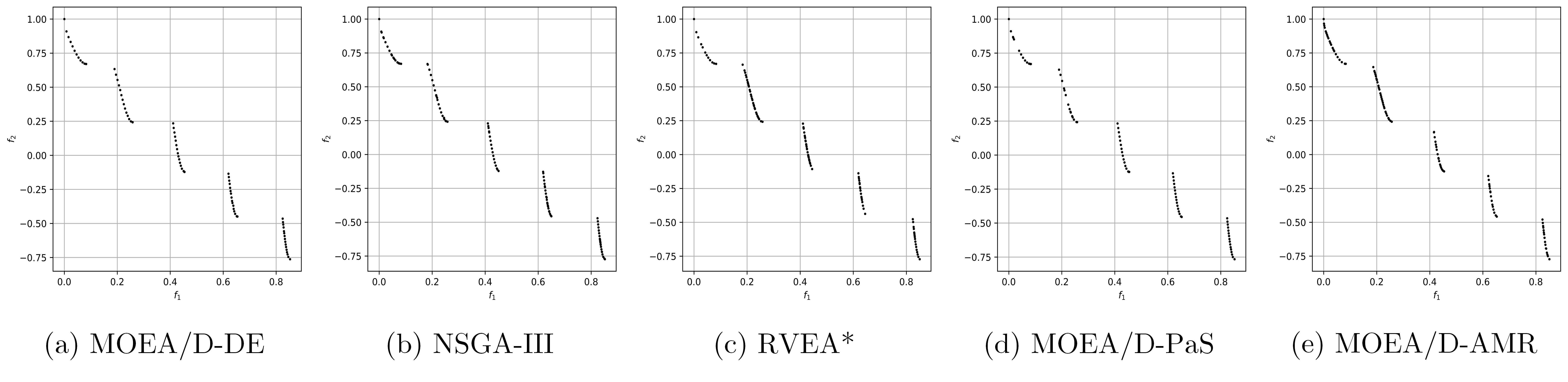}
	\caption{\footnotesize The final solution set with the median IGD value among 30 runs obtained by five algorithms on ZDT3.}
	\label{zdt3}
\end{figure}

\begin{figure}[H]
	\centering
	\includegraphics[width=\textwidth,height=3.8cm]{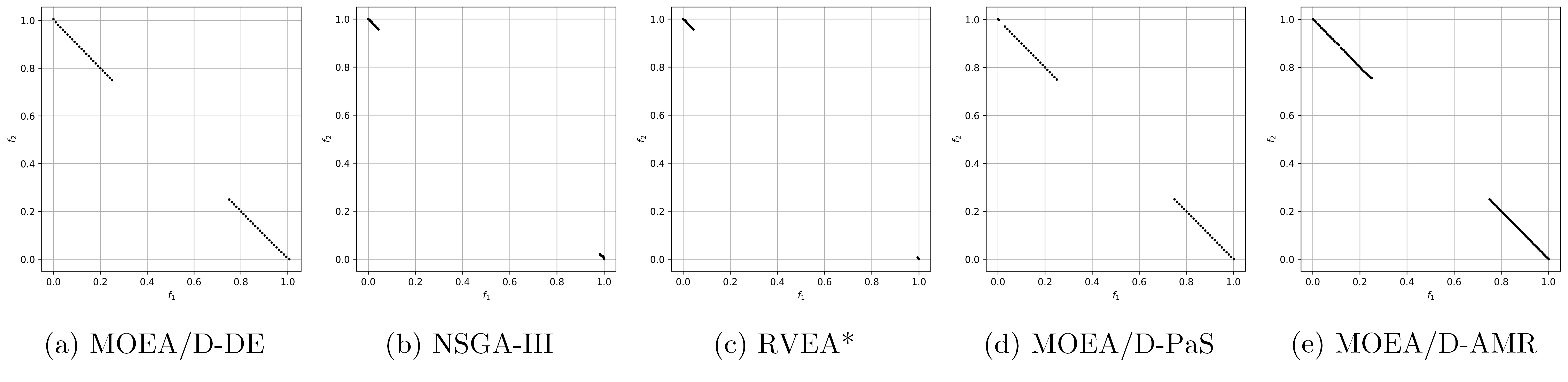}
	\caption{\footnotesize The final solution set with the median IGD value among 30 runs obtained by five algorithms on GLT1.}
	\label{glt1}
\end{figure}

\begin{figure}[H]
	\centering
	\includegraphics[width=\textwidth,height=3.5cm]{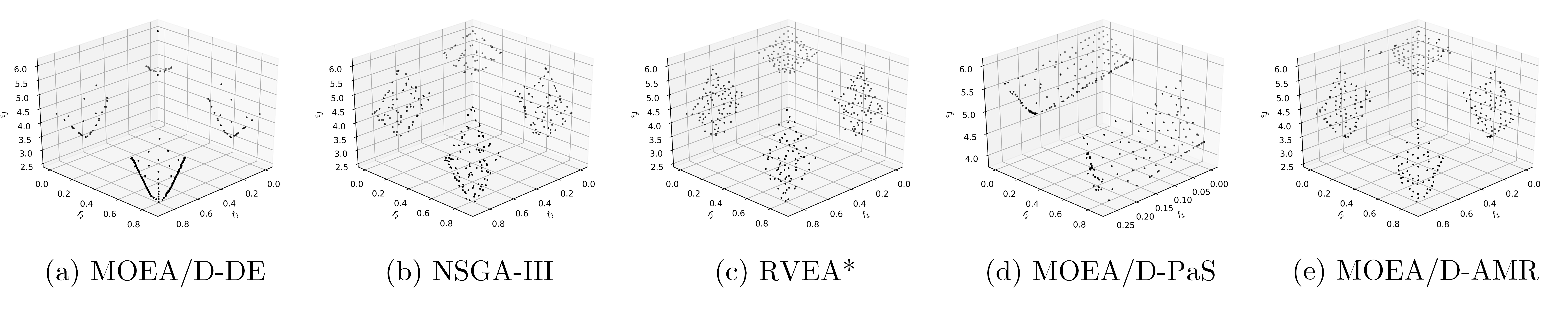}
	\caption{\footnotesize The final solution set with the median IGD value among 30 runs obtained by five algorithms on DTLZ7.}
	\label{dtlz7}
\end{figure}

\emph{5) Disconnected POFs:} Figs. \ref{zdt3}--\ref{dtlz7} show that the final solution set with the median IGD obtained by the five algorithms on ZDT3, GLT1 and DTLZ7, respectively. Togethering with Table \ref{igd}, we can see that only MOEA/D-AMR can maintain a good distribution of the final solution set. For ZDT3 and GLT1, there are some solutions obtained by MOEA/D-DE and MOEA/D-PaS concentrate on the discontinuous position of the POFs. An interesting observation is that when looking at the HV results shown in Table \ref{hv}, NSGA-III and MOEA/D-PaS perform the best on ZDT3 and GLT1, respectively. The most rational explanation is that the optimal distribution of solutions for hypervolume maximization may not be even \citep{IIS2018}.

\section{Applications on real-world MOPs}
As described in \cite{TI2020}, many real-world MOPs have irregular POFs. To further assess the performance of the proposed algorithm, we consider two multiobjective engineering optimal design optimization problems shown in \cite{TI2020}. The performance is compared with other competitive algorithms presented in Section \ref{sec4}. Herein, we use the same parameter settings and performance metrics given in Subsection \ref{sec4.1}. Since the true POFs of the two practical problems are not known, we make use of the reference POFs of the two design problems provided in the supplementary website (\url{https://github.com/ryojitanabe/reproblems}) mentioned in \cite{TI2020} to give a performance assessment of these algorithms.

\subsection{The hatch cover (HC) design problem}

The design of HC was studied by \cite{AH1989} who gave a detailed analysis. It is designed with the weight of the cover as objective functions subject to four constraints. \cite{TI2020} indicated that the constraint functions of HC design problem can be simultaneously minimized. Thus, the original HC design problem is remodeled as a biobjective optimization problem with two variables in \cite{TI2020}. Mathematical formulation of the biobjective optimization problem is presented in (\ref{hc_pro}), where the first objective is to minimize the weight of the hatch cover and the second objective is actually the sum of four constraint violations. The decision variables $x_{1}$ and $x_{2}$ denote the flange thickness (cm) and the beam height of the hatch cover (cm), respectively.

\begin{equation}\label{hc_pro}
	\begin{aligned}
		&\text{min}\quad \left(x_{1}+120x_{2}, \sum_{i=1}^{4}\max\{-g_{i}(x),0\}\right)^{T}\\
		&\text{s.t.}\quad\; x_{1}\in[0.5, 4],x_2\in[0.5, 50],
	\end{aligned}
\end{equation}
where $g_{1}(x)=1-\frac{\varrho_{b}}{\varrho_{b,\max}}$, $g_{2}(x)=1-\frac{\tau}{\tau_{\max}}$, $g_{3}(x)=1-\frac{\vartheta}{\vartheta_{\max}}$ and $g_{4}(x)=1-\frac{\varrho_{b}}{\varrho_{k}}$. The parameters and their descriptions are shown in Table \ref{hc_parameters}.

\begin{table}[htbp]\footnotesize
	\centering
	\caption{\footnotesize The parameter settings in (\ref{hc_pro}).}
	\begin{tabular}{llll}
		\hline
		Parameter                                          & Description                               & Value                      & Unit        \\ \hline
		$\varrho_{b}$                                       & Calculated bending stresses               & $4500/(x_{1}x_{2})$  & kg/cm$^{2}$ \\
		$\varrho_{b,{\rm max}}$ & Maximum allowable bending stresses        & 700                        & kg/cm$^{2}$ \\
		$\tau$                                             & Calculated  shearing  stresses            & $1800/x_{2}$               & kg/cm$^{2}$ \\
		$\tau_{{\rm max}}$                                 & Maximum  allowable  shearing  stresses    & 450                        & kg/cm$^{2}$       \\
		$\vartheta$                                           & Deflections  at  the mid  of  cover       & $562000/(Ex_{1}x_{2}^{2})$ & cm          \\
		$\vartheta_{{\rm max}}$                                           & Maximum  allowable  deflections        & 1.5 & cm          \\
		$\varrho_{k}$                                       & The  buckling  stresses  of  the  flanges & $Ex_{1}^{2}/100$           & kg/cm$^{2}$ \\
		$E$                                                & Young  Modulus                            & 700000                     & kg/cm$^{2}$\\
		\hline
	\end{tabular}
	\label{hc_parameters}
\end{table}

\begin{figure}[H]
	\centering
	\includegraphics[width=\textwidth,height=3.8cm]{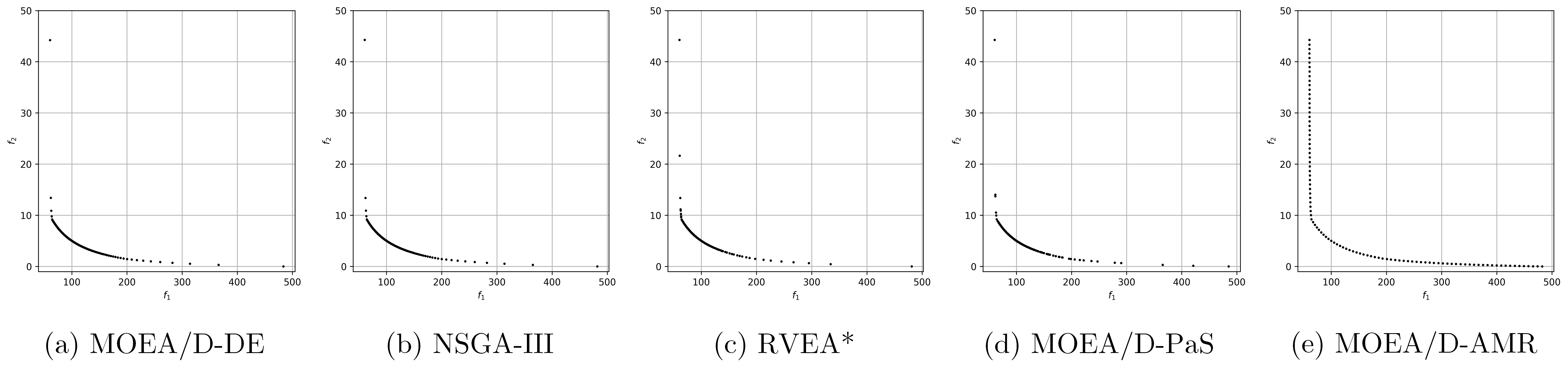}
	\caption{\footnotesize The final solution set with the median IGD value among 30 runs obtained by five algorithms on HC.}
	\label{hc}
\end{figure}

Fig. \ref{hc} shows that the graphical results of the final solution set obtained by MOEA/D-DE, NSGA-III, RVEA*, MOEA/D-PaS and MOEA/D-AMR on problem (\ref{hc_pro}).

Obviously, the solution set obtained by MOEA/D-AMR has much better distribution than those of others, which can provide a more reasonable choice for decision makers. A large portion of solutions obtained by MOEA/D-DE, NSGA-III, RVEA* and MOEA/D-PaS concentrate on the middle part of the POF. As a result, much smaller value of IGD and much larger value of HV are obtained by the proposed algorithm (see Table \ref{hchv}). Hence, our algorithm significantly outperforms than that of other compared algorithms in terms of diversity and coverage.

\begin{table}[H]\footnotesize
	\setlength\tabcolsep{4pt}
	\centering
	\caption{\footnotesize Statistical results of IGD and HV values (mean and standard deviation) found by different algorithms on HC design problem.}
	\begin{tabular}{ccccccc}
		\hline
		Problem             & Metric & MOEA/D-DE            & NSGA-III             & RVEA*                & MOEA/D-PaS           & MOEA/D-AMR            \\ \hline
		\multirow{2}{*}{HC} & IGD    & 8.8194e-0(2.98e-2) & 8.7709e-0(1.84e-2) & 9.9514e-0(1.83e-0) & 7.5090e-0(4.92e-1) & \hl{1.1498e-0(1.59e-2)} \\
		& HV     & 1.0536e-0(1.59e-5) & 1.0537e-0(5.55e-6) & 1.0536e-0(4.86e-4) & 1.0541e-0(2.06e-4) & \hl{1.0554e-0(5.49e-6)} \\
		\hline
	\end{tabular}
	\label{hchv}
\end{table}

\subsection{The rocket injector (RI) design problem}

The improvement of performance and life are the two primary objectives of the injector design.  The performance of the injector is expressed by the axial length of the thrust chamber, and the viability of the injector is related to the thermal field in the thrust chamber. A visual representation of the objectives is shown in \cite{VTPS2003, GVH2007}. High temperature produce high thermal stress on the injector and thrust chamber, which will reduce the service life of components, but improve the performance of the injector. Consequently, the dual goal of maximizing the performance and the life was cast as a four objective design problem \citep{VTPS2003}, i.e.,

$f_{1}$ is to minimize the maximum temperature of the injector face;

$f_{2}$ is to minimize the distance from the inlet;

$f_{3}$ is to minimize the maximum temperature on the post tip of the injector;

$f_{4}$ is to minimize the wall temperature at a distance three inches from the injector face.

Note that the objectives $f_{3}$ and $f_{4}$ were reported in \cite{GVH2007} strongly correlated in the design space. Hence, $f_{4}$ was dropped from the objectives list and the optimization problem was formulated with the remaining three objectives. This reformulated problem is as follows:

\begin{equation}\label{ri_pro}
	\begin{aligned}
		&\text{min}\quad (f_{1}(x),f_{2}(x),f_{3}(x))^{T}\\
		&\text{s.t.}\quad\; x_{i}\in[0,1],i\in\langle 4\rangle,
	\end{aligned}
\end{equation}
\noindent where

\begin{equation*}
	\begin{aligned}
		f_{1}(x)&= 0.692+0.477 x_{1}-0.687 x_{2}-0.08 x_{3}-0.065 x_{4}-0.167 x_{1}^{2}-0.0129 x_{1} x_{2} \\
		&\quad+0.0796 x_{2}^{2}-0.0634 x_{1} x_{3}-0.0257 x_{2} x_{3}+0.0877 x_{3}^{2}-0.0521 x_{1} x_{4}\\
		&\quad+0.00156 x_{2} x_{4}+0.00198 x_{3} x_{4}+0.0184 x_{4}^{2}, \\
		f_{2}(x)&= 0.153-0.322 x_{1}+0.396 x_{2}+0.424 x_{3}+0.0226 x_{4}+0.175 x_{1}^{2}+0.0185 x_{1} x_{2} \\
		&\quad-0.0701 x_{2}^{2} -0.251 x_{1} x_{3}+0.179 x_{2} x_{3}+0.015 x_{3}^{2}+0.0134 x_{1} x_{4} \\
		&\quad+0.0296 x_{2} x_{4}+0.0752 x_{3} x_{4}+0.0192 x_{4}^{2},\\
		f_{3}(x)&= 0.37-0.205 x_{1}+0.0307 x_{2}+0.108 x_{3}+1.019 x_{4}-0.135 x_{1}^{2}+0.0141 x_{1} x_{2} \\
		&\quad+0.0998 x_{2}^{2}+0.208 x_{1} x_{3}-0.0301 x_{2} x_{3}-0.226 x_{3}^{2}+0.353 x_{1} x_{4}  \\
		&\quad-0.0497 x_{3} x_{4}-0.423 x_{4}^{2} +0.202 x_{1}^{2} x_{2}-0.281 x_{1}^{2} x_{3}-0.342x_{1} x_{2}^{2}  \\
		&\quad-0.245 x_{2}^{2} x_{3}+0.281x_{2} x_{3}^{2} -0.184 x_{1} x_{4}^{2}-0.281 x_{1} x_{2} x_{3}.
	\end{aligned}
\end{equation*}
There are four design variables in this problem, which we need to make decisions. The four variables $x_{1}$, $x_{2}$, $x_{3}$ and $x_{4}$ describe the hydrogen flow angle, the hydrogen area, the oxygen area and the oxidizer post tip thickness, respectively.

\begin{figure}[H]
	\centering
	\includegraphics[width=\textwidth,height=3.5cm]{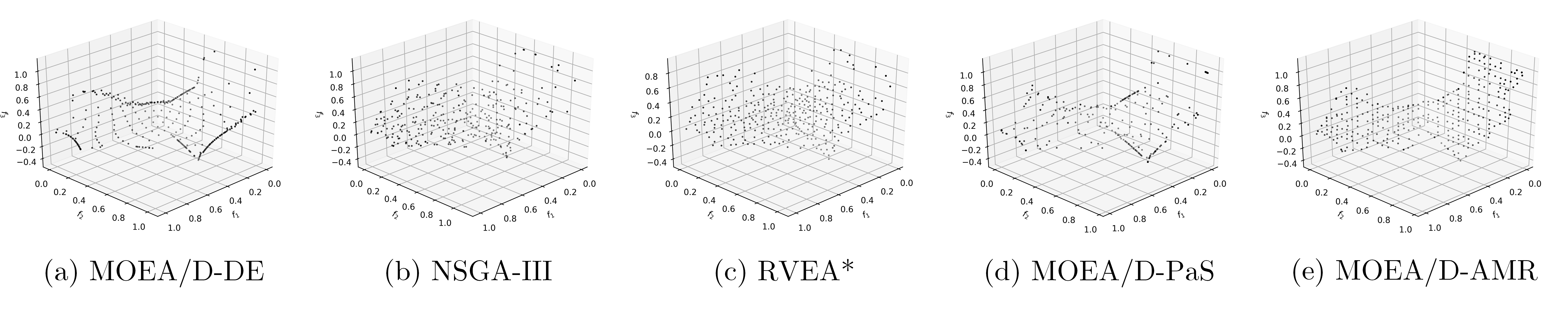}
	\caption{\footnotesize The final solution set with the median IGD value among 30 runs obtained by five algorithms on RI.}
	\label{ri}
\end{figure}

The statistic results of the mean and standard deviation of IGD and HV metrics obtained by the five algorithms are shown in Table \ref{rihv}. It can be concluded that the proposed algorithm is clearly better than the other compared algorithms. Furthermore, we plot the final solution set of the median IGD value among 30 runs for each algorithm (see Fig. \ref{ri}). As we have seen, the solutions obtained by MOEA/D-AMR are distributed more uniformly than those of other competitors.

\begin{table}[H]\footnotesize
	\setlength\tabcolsep{4pt}
	\centering
	\caption{\footnotesize Statistical results of IGD and HV values (mean and standard deviation) found by different algorithms on RI design problem.}
	\begin{tabular}{ccccccc}
		\hline
		Problem             & Metric & MOEA/D-DE            & NSGA-III             & RVEA*                & MOEA/D-PaS           & MOEA/D-AMR            \\ \hline
		\multirow{2}{*}{RI} & IGD    & 4.3249e-2(3.17e-4) & 3.6629e-2(1.08e-3) & 3.6760e-2(4.94e-3) & 4.7835e-2(9.66e-4) & \hl{3.2578e-2(1.04e-3)} \\
		& HV     & 9.5235e-1(6.66e-5) & 9.4930e-1(1.23e-3) & 9.5079e-1(1.61e-3) & 9.4939e-1(6.61e-4) & \hl{9.5588e-1(2.33e-3)} \\
		\hline
	\end{tabular}
	\label{rihv}
\end{table}

\section{Conclusion and Future Work}

In this paper, we have proposed a new algorithm named MOEA/D-AMR for enhancing the diversity of population in tacking some MOPs with regular and irregular POFs. Specifically, a scalarization approach, known as the Pascoletti-Serafini scalarization approach, has been introduced into the framework of MOEA/D-DE. More importantly, the Pascoletti-Serafini scalarization method has been equivalently transformed into a minimax formulation when restricting the direction to positive components. For the converted form, some scalarization methods such as TCH, $m$-TCH and $p$-TCH used in the framework of MOEA/D are its special cases intuitively based on the different selection of reference point and direction. Additionally, with the help of equidistant partitial and projection techniques, a new way of generating multi-reference points with relatively uniform distribution has been proposed. To avoid the occurrence of many unpromising reference points, that is, the solutions of the subproblems associated with these reference points converge to other subproblems, a strategy for adjusting the reference points has been suggested to tune the reference points in the later evolutionary stage. To assess the performance of the proposed algorithm, experimental comparisons have been conducted by comparing MOEAD/AMR with four state-of-the-art MOEAs, i.e., MOEA/D-DE, NSGA-III, RVEA* and MOEA/D-PaS, on several widely used test problems with different POFs and two real-world MOPs in engineering optimization. The experimental results have demonstrated that MOEAD/AMR is capable of obtaining superior or comparable performance to the other four state-of-the-art algorithms.

In the future, we would like to further investigate the following three issues. First, the proposed algorithm will be used to solve several many-objective optimization problems, which refers to the class of problems with four and more number of objectives. Note that the MOPs considered in this paper are all box-constrained problems. In fact, MOEA/D-AMR can easily incorporate some constraint-handling techniques such as the penalty function method reported in \cite{JZ2010} to deal with the constrained MOPs via slightly modifying the repair operation of the proposed algorithm. We leave addressing this issue as our second research topic. It is worth pointing out that, in the proposed algorithm, we use the objective space normalization by means of the estimated ideal and nadir points updated adaptively during the evolutionary process. However, \cite{HL2021} pointed out that if the two points are inaccurately estimated, then the objective space normalization may deteriorate the performance of a MOEA. Therefore, an effective and robust normalization method should be embedded into the proposed algorithm, which is also our third future work.


\small
\begin{spacing}{0.95}
	\bibliographystyle{model5-names}
	\biboptions{authoryear}
	\bibliography{mybibfile}

\begin{thebibliography}{65}
\expandafter\ifx\csname natexlab\endcsname\relax\def\natexlab#1{#1}\fi
\providecommand{\url}[1]{\texttt{#1}}
\providecommand{\href}[2]{#2}
\providecommand{\path}[1]{#1}
\providecommand{\DOIprefix}{doi:}
\providecommand{\ArXivprefix}{arXiv:}
\providecommand{\URLprefix}{URL: }
\providecommand{\Pubmedprefix}{pmid:}
\providecommand{\doi}[1]{\href{http://dx.doi.org/#1}{\path{#1}}}
\providecommand{\Pubmed}[1]{\href{pmid:#1}{\path{#1}}}
\providecommand{\bibinfo}[2]{#2}
\ifx\xfnm\relax \def\xfnm[#1]{\unskip,\space#1}\fi
\bibitem[{Akbari et~al.(2018)Akbari, Ghaznavi \& Khorram}]{AGK2018}
\bibinfo{author}{Akbari, F.}, \bibinfo{author}{Ghaznavi, M.}, \&
  \bibinfo{author}{Khorram, E.} (\bibinfo{year}{2018}).
\newblock \bibinfo{title}{A revised {P}ascoletti-{S}erafini scalarization
  method for multiobjective optimization problems}.
\newblock {\it \bibinfo{journal}{Journal of Optimization Theory and
  Applications}\/},  {\it \bibinfo{volume}{178}\/}, \bibinfo{pages}{560--590}.
\newblock \bibinfo{note}{\url{https://doi.org/10.1007/s10957-018-1289-2}}.
\bibitem[{Amir \& Hasegawa(1989)}]{AH1989}
\bibinfo{author}{Amir, H.~M.}, \& \bibinfo{author}{Hasegawa, T.}
  (\bibinfo{year}{1989}).
\newblock \bibinfo{title}{Nonlinear mixed-discrete structural optimization}.
\newblock {\it \bibinfo{journal}{Journal of Structural Engineering}\/},  {\it
  \bibinfo{volume}{115}\/}, \bibinfo{pages}{626--646}.
\newblock \bibinfo{note}{\url{https://doi.org/10.1061/(ASCE)0733-9445}}.
\bibitem[{Ashby(2000)}]{A2000}
\bibinfo{author}{Ashby, M.} (\bibinfo{year}{2000}).
\newblock \bibinfo{title}{Multi-objective optimization in material design and
  selection}.
\newblock {\it \bibinfo{journal}{Acta Materialia}\/},  {\it
  \bibinfo{volume}{48}\/}, \bibinfo{pages}{359--369}.
\newblock \bibinfo{note}{\url{https://doi.org/10.1016/S1359-6454(99)00304-3}}.
\bibitem[{Audet et~al.(2020)Audet, Bigeon, Cartier, Le~Digabel \&
  Salomon}]{AB2020}
\bibinfo{author}{Audet, C.}, \bibinfo{author}{Bigeon, J.},
  \bibinfo{author}{Cartier, D.}, \bibinfo{author}{Le~Digabel, S.}, \&
  \bibinfo{author}{Salomon, L.} (\bibinfo{year}{2020}).
\newblock \bibinfo{title}{Performance indicators in multiobjective
  optimization}.
\newblock {\it \bibinfo{journal}{European Journal of Operational Research}\/},
  {\it \bibinfo{volume}{292}\/}, \bibinfo{pages}{397--422}.
\newblock \bibinfo{note}{\url{https://doi.org/10.1016/j.ejor.2020.11.016}}.
\bibitem[{Bader \& Zitzler(2011)}]{BZ2011}
\bibinfo{author}{Bader, J.}, \& \bibinfo{author}{Zitzler, E.}
  (\bibinfo{year}{2011}).
\newblock \bibinfo{title}{Hyp{E}: An algorithm for fast hypervolume-based
  many-objective optimization}.
\newblock {\it \bibinfo{journal}{Evolutionary Computation}\/},  {\it
  \bibinfo{volume}{19}\/}, \bibinfo{pages}{45--76}.
\newblock \bibinfo{note}{\url{https://doi.org/10.1162/EVCO_a_00009}}.
\bibitem[{Bortz et~al.(2014)Bortz, Burger, Asprion, Blagov, B{\"o}ttcher,
  Nowak, Scheithauer, Welke, K{\"u}fer \& Hasse}]{BBA2014}
\bibinfo{author}{Bortz, M.}, \bibinfo{author}{Burger, J.},
  \bibinfo{author}{Asprion, N.}, \bibinfo{author}{Blagov, S.},
  \bibinfo{author}{B{\"o}ttcher, R.}, \bibinfo{author}{Nowak, U.},
  \bibinfo{author}{Scheithauer, A.}, \bibinfo{author}{Welke, R.},
  \bibinfo{author}{K{\"u}fer, K.-H.}, \& \bibinfo{author}{Hasse, H.}
  (\bibinfo{year}{2014}).
\newblock \bibinfo{title}{Multi-criteria optimization in chemical process
  design and decision support by navigation on pareto sets}.
\newblock {\it \bibinfo{journal}{Computers Chemical Engineering}\/},  {\it
  \bibinfo{volume}{60}\/}, \bibinfo{pages}{354--363}.
\newblock
  \bibinfo{note}{\url{https://doi.org/10.1016/j.compchemeng.2013.09.015}}.
\bibitem[{Cheaitou \& Cariou(2019)}]{CC2019}
\bibinfo{author}{Cheaitou, A.}, \& \bibinfo{author}{Cariou, P.}
  (\bibinfo{year}{2019}).
\newblock \bibinfo{title}{Greening of maritime transportation: a
  multi-objective optimization approach}.
\newblock {\it \bibinfo{journal}{Annals of Operations Research}\/},  {\it
  \bibinfo{volume}{273}\/}, \bibinfo{pages}{501--525}.
\newblock \bibinfo{note}{\url{https://doi.org/10.1007/s10479-018-2786-2}}.
\bibitem[{Cheng et~al.(2016)Cheng, Jin, Olhofer \& Sendhoff}]{CJOS2016}
\bibinfo{author}{Cheng, R.}, \bibinfo{author}{Jin, Y.},
  \bibinfo{author}{Olhofer, M.}, \& \bibinfo{author}{Sendhoff, B.}
  (\bibinfo{year}{2016}).
\newblock \bibinfo{title}{A reference vector guided evolutionary algorithm for
  many-objective optimization}.
\newblock {\it \bibinfo{journal}{IEEE Transactions on Evolutionary
  Computation}\/},  {\it \bibinfo{volume}{20}\/}, \bibinfo{pages}{773--791}.
\newblock \bibinfo{note}{\url{https://doi.org/10.1109/TEVC.2016.2519378}}.
\bibitem[{Coello et~al.(2007)Coello, Lamont \& Van~Veldhuizen}]{C2007}
\bibinfo{author}{Coello, C. A.~C.}, \bibinfo{author}{Lamont, G.~B.}, \&
  \bibinfo{author}{Van~Veldhuizen, D.~A.} (\bibinfo{year}{2007}).
\newblock {\it \bibinfo{title}{Evolutionary algorithms for solving
  multi-objective problems}\/}.
\newblock \bibinfo{publisher}{Springer}.
\bibitem[{Das \& Dennis(1998)}]{DD1998}
\bibinfo{author}{Das, I.}, \& \bibinfo{author}{Dennis, J.~E.}
  (\bibinfo{year}{1998}).
\newblock \bibinfo{title}{Normal-boundary intersection: {A} new method for
  generating the {P}areto surface in nonlinear multicriteria optimization
  problems}.
\newblock {\it \bibinfo{journal}{SIAM Journal on Optimization}\/},  {\it
  \bibinfo{volume}{8}\/}, \bibinfo{pages}{631--657}.
\newblock \bibinfo{note}{\url{https://doi.org/10.1137/S1052623496307510}}.
\bibitem[{Deb \& Jain(2013)}]{DJ2014}
\bibinfo{author}{Deb, K.}, \& \bibinfo{author}{Jain, H.}
  (\bibinfo{year}{2013}).
\newblock \bibinfo{title}{An evolutionary many-objective optimization algorithm
  using reference-point-based nondominated sorting approach, part {I}: solving
  problems with box constraints}.
\newblock {\it \bibinfo{journal}{IEEE Transactions on Evolutionary
  Computation}\/},  {\it \bibinfo{volume}{18}\/}, \bibinfo{pages}{577--601}.
\newblock \bibinfo{note}{\url{https://doi.org/10.1109/TEVC.2013.2281535}}.
\bibitem[{Deb et~al.(2002)Deb, Pratap, Agarwal \& Meyarivan}]{DAPM2002}
\bibinfo{author}{Deb, K.}, \bibinfo{author}{Pratap, A.},
  \bibinfo{author}{Agarwal, S.}, \& \bibinfo{author}{Meyarivan, T.}
  (\bibinfo{year}{2002}).
\newblock \bibinfo{title}{A fast and elitist multiobjective genetic algorithm:
  {NSGA-II}}.
\newblock {\it \bibinfo{journal}{IEEE Transactions on Evolutionary
  Computation}\/},  {\it \bibinfo{volume}{6}\/}, \bibinfo{pages}{182--197}.
\newblock \bibinfo{note}{\url{https://doi.org/10.1109/4235.996017}}.
\bibitem[{Deb et~al.(2005)Deb, Thiele, Laumanns \& Zitzler}]{DTLZ2005}
\bibinfo{author}{Deb, K.}, \bibinfo{author}{Thiele, L.},
  \bibinfo{author}{Laumanns, M.}, \& \bibinfo{author}{Zitzler, E.}
  (\bibinfo{year}{2005}).
\newblock \bibinfo{title}{Scalable test problems for evolutionary
  multiobjective optimization}.
\newblock In {\it \bibinfo{booktitle}{Abraham A., Jain L., Goldberg R. (eds)
  Evolutionary Multiobjective Optimization. Advanced Information and Knowledge
  Processing}\/} (pp. \bibinfo{pages}{105--145}).
\newblock \bibinfo{publisher}{Springer, London}.
\newblock \bibinfo{note}{\url{https://doi.org/10.1007/1-84628-137-7_6}}.
\bibitem[{Dolatnezhadsomarin \& Khorram(2019)}]{DE2019t}
\bibinfo{author}{Dolatnezhadsomarin, A.}, \& \bibinfo{author}{Khorram, E.}
  (\bibinfo{year}{2019}).
\newblock \bibinfo{title}{Two efficient algorithms for constructing almost even
  approximations of the {P}areto front in multi-objective optimization
  problems}.
\newblock {\it \bibinfo{journal}{Engineering Optimization}\/},  {\it
  \bibinfo{volume}{51}\/}, \bibinfo{pages}{567--589}.
\newblock \bibinfo{note}{\url{https://doi.org/10.1080/0305215X.2018.1479405}}.
\bibitem[{Dong et~al.(2020)Dong, Wang \& Tang}]{DWT2020}
\bibinfo{author}{Dong, Z.}, \bibinfo{author}{Wang, X.}, \&
  \bibinfo{author}{Tang, L.} (\bibinfo{year}{2020}).
\newblock \bibinfo{title}{{MOEA/D} with a self-adaptive weight vector
  adjustment strategy based on chain segmentation}.
\newblock {\it \bibinfo{journal}{Information Sciences}\/},  {\it
  \bibinfo{volume}{521}\/}, \bibinfo{pages}{209--230}.
\newblock \bibinfo{note}{\url{https://doi.org/10.1016/j.ins.2020.02.056}}.
\bibitem[{Eichfelder(2008)}]{E2008}
\bibinfo{author}{Eichfelder, G.} (\bibinfo{year}{2008}).
\newblock {\it \bibinfo{title}{Adaptive scalarization methods in multiobjective
  optimization}\/}.
\newblock \bibinfo{publisher}{Springer, Berlin}.
\bibitem[{Eichfelder(2009)}]{E2009a}
\bibinfo{author}{Eichfelder, G.} (\bibinfo{year}{2009}).
\newblock \bibinfo{title}{An adaptive scalarization method in multiobjective
  optimization}.
\newblock {\it \bibinfo{journal}{SIAM Journal on Optimization}\/},  {\it
  \bibinfo{volume}{19}\/}, \bibinfo{pages}{1694--1718}.
\newblock \bibinfo{note}{\url{https://doi.org/10.1137/060672029}}.
\bibitem[{Eichfelder et~al.(2021)Eichfelder, Kirst, Meng \& Stein}]{E2021}
\bibinfo{author}{Eichfelder, G.}, \bibinfo{author}{Kirst, P.},
  \bibinfo{author}{Meng, L.}, \& \bibinfo{author}{Stein, O.}
  (\bibinfo{year}{2021}).
\newblock \bibinfo{title}{A general branch-and-bound framework for continuous
  global multiobjective optimization}.
\newblock {\it \bibinfo{journal}{Journal of Global Optimization}\/},  {\it
  \bibinfo{volume}{80}\/}, \bibinfo{pages}{195--227}.
\newblock \bibinfo{note}{\url{https://doi.org/10.1007/s10898-020-00984-y}}.
\bibitem[{Eichfelder \& Warnow(2020)}]{EW2020}
\bibinfo{author}{Eichfelder, G.}, \& \bibinfo{author}{Warnow, L.}
  (\bibinfo{year}{2020}).
\newblock \bibinfo{title}{An approximation algorithm for multi-objective
  optimization problems using a box-coverage}, .
\newblock
  \bibinfo{note}{\url{https://www.optimization-online.org/DB_HTML/2020/10/8079.html}}.
\bibitem[{Fliege et~al.(2009)Fliege, Gra{\~n}a~Drummond \& Svaiter}]{FDS2009}
\bibinfo{author}{Fliege, J.}, \bibinfo{author}{Gra{\~n}a~Drummond, L.~M.}, \&
  \bibinfo{author}{Svaiter, B.~F.} (\bibinfo{year}{2009}).
\newblock \bibinfo{title}{Newton's method for multiobjective optimization}.
\newblock {\it \bibinfo{journal}{SIAM Journal on Optimization}\/},  {\it
  \bibinfo{volume}{20}\/}, \bibinfo{pages}{602--626}.
\newblock \bibinfo{note}{\url{https://doi.org/10.1137/08071692X}}.
\bibitem[{Fukuda \& Gra{\~n}a~Drummond(2014)}]{FD2014}
\bibinfo{author}{Fukuda, E.~H.}, \& \bibinfo{author}{Gra{\~n}a~Drummond, L.~M.}
  (\bibinfo{year}{2014}).
\newblock \bibinfo{title}{A survey on multiobjective descent methods}.
\newblock {\it \bibinfo{journal}{Pesquisa Operacional}\/},  {\it
  \bibinfo{volume}{34}\/}, \bibinfo{pages}{585--620}.
\newblock
  \bibinfo{note}{\url{https://doi.org/10.1590/0101-7438.2014.034.03.0585}}.
\bibitem[{Goel et~al.(2007)Goel, Vaidyanathan, Haftka, Shyy, Queipo \&
  Tucker}]{GVH2007}
\bibinfo{author}{Goel, T.}, \bibinfo{author}{Vaidyanathan, R.},
  \bibinfo{author}{Haftka, R.~T.}, \bibinfo{author}{Shyy, W.},
  \bibinfo{author}{Queipo, N.~V.}, \& \bibinfo{author}{Tucker, K.}
  (\bibinfo{year}{2007}).
\newblock \bibinfo{title}{Response surface approximation of {P}areto optimal
  front in multi-objective optimization}.
\newblock {\it \bibinfo{journal}{Computer Methods in Applied Mechanics and
  Engineering}\/},  {\it \bibinfo{volume}{196}\/}, \bibinfo{pages}{879--893}.
\newblock \bibinfo{note}{\url{https://doi.org/10.1016/j.cma.2006.07.010}}.
\bibitem[{Gu et~al.(2012)Gu, Liu \& Tan}]{GLT2012}
\bibinfo{author}{Gu, F.}, \bibinfo{author}{Liu, H.-L.}, \&
  \bibinfo{author}{Tan, K.~C.} (\bibinfo{year}{2012}).
\newblock \bibinfo{title}{A multiobjective evolutionary algorithm using dynamic
  weight design method}.
\newblock {\it \bibinfo{journal}{International Journal of Innovative Computing,
  Information and Control}\/},  {\it \bibinfo{volume}{8}\/},
  \bibinfo{pages}{3677--3688}.
\bibitem[{He et~al.(2021)He, Ishibuchi, Trivedi, Wang, Nan \&
  Srinivasan}]{HL2021}
\bibinfo{author}{He, L.}, \bibinfo{author}{Ishibuchi, H.},
  \bibinfo{author}{Trivedi, A.}, \bibinfo{author}{Wang, H.},
  \bibinfo{author}{Nan, Y.}, \& \bibinfo{author}{Srinivasan, D.}
  (\bibinfo{year}{2021}).
\newblock \bibinfo{title}{A survey of normalization methods in multiobjective
  evolutionary algorithms}.
\newblock {\it \bibinfo{journal}{IEEE Transactions on Evolutionary
  Computation}\/}, . \DOIprefix\doi{https://doi.org/10.1109/TEVC.2021.3076514}.
\bibitem[{Ishibuchi et~al.(2018)Ishibuchi, Imada, Setoguchi \&
  Nojima}]{IIS2018}
\bibinfo{author}{Ishibuchi, H.}, \bibinfo{author}{Imada, R.},
  \bibinfo{author}{Setoguchi, Y.}, \& \bibinfo{author}{Nojima, Y.}
  (\bibinfo{year}{2018}).
\newblock \bibinfo{title}{How to specify a reference point in hypervolume
  calculation for fair performance comparison}.
\newblock {\it \bibinfo{journal}{Evolutionary Computation}\/},  {\it
  \bibinfo{volume}{26}\/}, \bibinfo{pages}{411--440}.
\newblock \bibinfo{note}{\url{https://doi.org/10.1162/evco_a_00226}}.
\bibitem[{Ishibuchi et~al.(2010)Ishibuchi, Sakane, Tsukamoto \&
  Nojima}]{ISTN2010}
\bibinfo{author}{Ishibuchi, H.}, \bibinfo{author}{Sakane, Y.},
  \bibinfo{author}{Tsukamoto, N.}, \& \bibinfo{author}{Nojima, Y.}
  (\bibinfo{year}{2010}).
\newblock \bibinfo{title}{Simultaneous use of different scalarizing functions
  in {MOEA/D}}.
\newblock In {\it \bibinfo{booktitle}{Proceedings of the 12th Annual Conference
  on Genetic and Evolutionary Computation}\/} (pp. \bibinfo{pages}{519--526}).
\newblock \bibinfo{note}{\url{https://doi.org/10.1145/1830483.1830577}}.
\bibitem[{Jain \& Deb(2013)}]{JD2014}
\bibinfo{author}{Jain, H.}, \& \bibinfo{author}{Deb, K.}
  (\bibinfo{year}{2013}).
\newblock \bibinfo{title}{An evolutionary many-objective optimization algorithm
  using reference-point based nondominated sorting approach, part {II}:
  Handling constraints and extending to an adaptive approach}.
\newblock {\it \bibinfo{journal}{IEEE Transactions on Evolutionary
  Computation}\/},  {\it \bibinfo{volume}{18}\/}, \bibinfo{pages}{602--622}.
\newblock \bibinfo{note}{\url{https://doi.org/10.1109/TEVC.2013.2281534}}.
\bibitem[{Jan \& Zhang(2010)}]{JZ2010}
\bibinfo{author}{Jan, M.~A.}, \& \bibinfo{author}{Zhang, Q.}
  (\bibinfo{year}{2010}).
\newblock \bibinfo{title}{{MOEA/D} for constrained multiobjective optimization:
  some preliminary experimental results}.
\newblock In {\it \bibinfo{booktitle}{2010 UK Workshop on Computational
  Intelligence (UKCI)}\/} (pp. \bibinfo{pages}{1--6}).
\newblock \bibinfo{note}{\url{https://doi.org/10.1109/UKCI.2010.5625585}}.
\bibitem[{Jiang et~al.(2011)Jiang, Cai, Zhang \& Ong}]{JCZ2011}
\bibinfo{author}{Jiang, S.}, \bibinfo{author}{Cai, Z.}, \bibinfo{author}{Zhang,
  J.}, \& \bibinfo{author}{Ong, Y.-S.} (\bibinfo{year}{2011}).
\newblock \bibinfo{title}{Multiobjective optimization by decomposition with
  {P}areto-adaptive weight vectors}.
\newblock In {\it \bibinfo{booktitle}{2011 Seventh International Conference on
  Natural Computation}\/} (pp. \bibinfo{pages}{1260--1264}).
\newblock volume~\bibinfo{volume}{3}.
\newblock \bibinfo{note}{\url{https://doi.org/10.1109/ICNC.2011.6022367}}.
\bibitem[{Jiang \& Yang(2015)}]{JY2015}
\bibinfo{author}{Jiang, S.}, \& \bibinfo{author}{Yang, S.}
  (\bibinfo{year}{2015}).
\newblock \bibinfo{title}{An improved multiobjective optimization evolutionary
  algorithm based on decomposition for complex {P}areto fronts}.
\newblock {\it \bibinfo{journal}{IEEE Transactions on Cybernetics}\/},  {\it
  \bibinfo{volume}{46}\/}, \bibinfo{pages}{421--437}.
\newblock \bibinfo{note}{\url{https://doi.org/10.1109/TCYB.2015.2403131}}.
\bibitem[{Jiang et~al.(2017)Jiang, Yang, Wang \& Liu}]{JYWL2017}
\bibinfo{author}{Jiang, S.}, \bibinfo{author}{Yang, S.}, \bibinfo{author}{Wang,
  Y.}, \& \bibinfo{author}{Liu, X.} (\bibinfo{year}{2017}).
\newblock \bibinfo{title}{Scalarizing functions in decomposition-based
  multiobjective evolutionary algorithms}.
\newblock {\it \bibinfo{journal}{IEEE Transactions on Evolutionary
  Computation}\/},  {\it \bibinfo{volume}{22}\/}, \bibinfo{pages}{296--313}.
\newblock \bibinfo{note}{\url{https://doi.org/10.1109/TEVC.2017.2707980}}.
\bibitem[{Jones et~al.(2002)Jones, Mirrazavi \& Tamiz}]{JM2002m}
\bibinfo{author}{Jones, D.~F.}, \bibinfo{author}{Mirrazavi, S.~K.}, \&
  \bibinfo{author}{Tamiz, M.} (\bibinfo{year}{2002}).
\newblock \bibinfo{title}{Multi-objective meta-heuristics: An overview of the
  current state-of-the-art}.
\newblock {\it \bibinfo{journal}{European Journal of Operational Research}\/},
  {\it \bibinfo{volume}{137}\/}, \bibinfo{pages}{1--9}.
\newblock \bibinfo{note}{\url{https://doi.org/10.1016/S0377-2217(01)00123-0}}.
\bibitem[{Khorram et~al.(2014)Khorram, Khaledian \& Khaledyan}]{KKK2014}
\bibinfo{author}{Khorram, E.}, \bibinfo{author}{Khaledian, K.}, \&
  \bibinfo{author}{Khaledyan, M.} (\bibinfo{year}{2014}).
\newblock \bibinfo{title}{A numerical method for constructing the pareto front
  of multi-objective optimization problems}.
\newblock {\it \bibinfo{journal}{Journal of Computational and Applied
  Mathematics}\/},  {\it \bibinfo{volume}{261}\/}, \bibinfo{pages}{158--171}.
\newblock \bibinfo{note}{\url{https://doi.org/10.1016/j.cam.2013.11.007}}.
\bibitem[{Li et~al.(2019{\natexlab{a}})Li, Deng, Zhang \& Sun}]{LDZS2019}
\bibinfo{author}{Li, H.}, \bibinfo{author}{Deng, J.}, \bibinfo{author}{Zhang,
  Q.}, \& \bibinfo{author}{Sun, J.} (\bibinfo{year}{2019}{\natexlab{a}}).
\newblock \bibinfo{title}{Adaptive epsilon dominance in decomposition-based
  multiobjective evolutionary algorithm}.
\newblock {\it \bibinfo{journal}{Swarm and Evolutionary Computation}\/},  {\it
  \bibinfo{volume}{45}\/}, \bibinfo{pages}{52--67}.
\newblock \bibinfo{note}{\url{https://doi.org/10.1016/j.swevo.2018.12.007}}.
\bibitem[{Li et~al.(2019{\natexlab{b}})Li, Sun, Zhang \& Shui}]{LSZ2019}
\bibinfo{author}{Li, H.}, \bibinfo{author}{Sun, J.}, \bibinfo{author}{Zhang,
  Q.}, \& \bibinfo{author}{Shui, Y.} (\bibinfo{year}{2019}{\natexlab{b}}).
\newblock \bibinfo{title}{Adjustment of weight vectors of penalty-based
  boundary intersection method in {MOEA/D}}.
\newblock In {\it \bibinfo{booktitle}{Deb K. et al. (eds) Evolutionary
  Multi-Criterion Optimization. EMO 2019. Lecture Notes in Computer Science}\/}
  (pp. \bibinfo{pages}{91--100}).
\newblock \bibinfo{note}{\url{https://doi.org/10.1007/978-3-030-12598-1_8}}.
\bibitem[{Li \& Zhang(2009)}]{LZ2009}
\bibinfo{author}{Li, H.}, \& \bibinfo{author}{Zhang, Q.}
  (\bibinfo{year}{2009}).
\newblock \bibinfo{title}{Multiobjective optimization problems with complicated
  pareto sets, {MOEA/D} and {NSGA-II}}.
\newblock {\it \bibinfo{journal}{IEEE Transactions on Evolutionary
  Computation}\/},  {\it \bibinfo{volume}{13}\/}, \bibinfo{pages}{284--302}.
\newblock \bibinfo{note}{\url{https://doi.org/10.1109/TEVC.2008.925798}}.
\bibitem[{Li et~al.(2013)Li, Zhang, Kwong, Li \& Wang}]{LZK2014}
\bibinfo{author}{Li, K.}, \bibinfo{author}{Zhang, Q.}, \bibinfo{author}{Kwong,
  S.}, \bibinfo{author}{Li, M.}, \& \bibinfo{author}{Wang, R.}
  (\bibinfo{year}{2013}).
\newblock \bibinfo{title}{Stable matching-based selection in evolutionary
  multiobjective optimization}.
\newblock {\it \bibinfo{journal}{IEEE Transactions on Evolutionary
  Computation}\/},  {\it \bibinfo{volume}{18}\/}, \bibinfo{pages}{909--923}.
\newblock \bibinfo{note}{\url{https://doi.org/10.1109/TEVC.2013.2293776}}.
\bibitem[{Ma et~al.(2020)Ma, Yu, Li, Qi \& Zhu}]{MYL2020}
\bibinfo{author}{Ma, X.}, \bibinfo{author}{Yu, Y.}, \bibinfo{author}{Li, X.},
  \bibinfo{author}{Qi, Y.}, \& \bibinfo{author}{Zhu, Z.}
  (\bibinfo{year}{2020}).
\newblock \bibinfo{title}{A survey of weight vector adjustment methods for
  decomposition-based multiobjective evolutionary algorithms}.
\newblock {\it \bibinfo{journal}{IEEE Transactions on Evolutionary
  Computation}\/},  {\it \bibinfo{volume}{24}\/}, \bibinfo{pages}{634--649}.
\newblock \bibinfo{note}{\url{https://doi.org/10.1109/TEVC.2020.2978158}}.
\bibitem[{Ma et~al.(2017)Ma, Zhang, Tian, Yang \& Zhu}]{MZT2018}
\bibinfo{author}{Ma, X.}, \bibinfo{author}{Zhang, Q.}, \bibinfo{author}{Tian,
  G.}, \bibinfo{author}{Yang, J.}, \& \bibinfo{author}{Zhu, Z.}
  (\bibinfo{year}{2017}).
\newblock \bibinfo{title}{On {T}chebycheff decomposition approaches for
  multiobjective evolutionary optimization}.
\newblock {\it \bibinfo{journal}{IEEE Transactions on Evolutionary
  Computation}\/},  {\it \bibinfo{volume}{22}\/}, \bibinfo{pages}{226--244}.
\newblock \bibinfo{note}{\url{https://doi.org/10.1109/TEVC.2017.2704118}}.
\bibitem[{Miettinen(2000)}]{M1999}
\bibinfo{author}{Miettinen, K.} (\bibinfo{year}{2000}).
\newblock {\it \bibinfo{title}{Nonlinear multiobjective optimization}\/}.
\newblock \bibinfo{publisher}{Boston, MAA: Kluwer}.
\bibitem[{Miglierina et~al.(2008)Miglierina, Molho \& Recchioni}]{MMR2008}
\bibinfo{author}{Miglierina, E.}, \bibinfo{author}{Molho, E.}, \&
  \bibinfo{author}{Recchioni, M.~C.} (\bibinfo{year}{2008}).
\newblock \bibinfo{title}{Box-constrained multi-objective optimization: a
  gradient-like method without ``a priori'' scalarization}.
\newblock {\it \bibinfo{journal}{European Journal of Operational Research}\/},
  {\it \bibinfo{volume}{188}\/}, \bibinfo{pages}{662--682}.
\newblock \bibinfo{note}{\url{https://doi.org/10.1016/j.ejor.2007.05.015}}.
\bibitem[{Murata et~al.(2001)Murata, Ishibuchi \& Gen}]{MIG2001}
\bibinfo{author}{Murata, T.}, \bibinfo{author}{Ishibuchi, H.}, \&
  \bibinfo{author}{Gen, M.} (\bibinfo{year}{2001}).
\newblock \bibinfo{title}{Specification of genetic search directions in
  cellular multi-objective genetic algorithms}.
\newblock In {\it \bibinfo{booktitle}{Zitzler E., Thiele L., Deb K., Coello
  Coello C.A., Corne D. (eds) Evolutionary Multi-Criterion Optimization. EMO
  2001. Lecture Notes in Computer Science}\/} (pp. \bibinfo{pages}{82--95}).
\newblock volume \bibinfo{volume}{1993}.
\newblock \bibinfo{note}{\url{https://doi.org/10.1007/3-540-44719-9_6}}.
\bibitem[{Pardalos et~al.(2017)Pardalos, Zilinskas \& Zilinskas}]{PZZ2017}
\bibinfo{author}{Pardalos, P.~M.}, \bibinfo{author}{Zilinskas, A.}, \&
  \bibinfo{author}{Zilinskas, J.} (\bibinfo{year}{2017}).
\newblock {\it \bibinfo{title}{Non-convex multi-objective optimization}\/}.
\newblock \bibinfo{publisher}{Springer}.
\bibitem[{Pascoletti \& Serafini(1984)}]{PS1984}
\bibinfo{author}{Pascoletti, A.}, \& \bibinfo{author}{Serafini, P.}
  (\bibinfo{year}{1984}).
\newblock \bibinfo{title}{Scalarizing vector optimization problems}.
\newblock {\it \bibinfo{journal}{Journal of Optimization Theory and
  Applications}\/},  {\it \bibinfo{volume}{42}\/}, \bibinfo{pages}{499--524}.
\newblock \bibinfo{note}{\url{https://doi.org/10.1007/BF00934564}}.
\bibitem[{Qi et~al.(2014)Qi, Ma, Liu, Jiao, Sun \& Wu}]{Q2014}
\bibinfo{author}{Qi, Y.}, \bibinfo{author}{Ma, X.}, \bibinfo{author}{Liu, F.},
  \bibinfo{author}{Jiao, L.}, \bibinfo{author}{Sun, J.}, \&
  \bibinfo{author}{Wu, J.} (\bibinfo{year}{2014}).
\newblock \bibinfo{title}{{MOEA/D} with adaptive weight adjustment}.
\newblock {\it \bibinfo{journal}{Evolutionary Computation}\/},  {\it
  \bibinfo{volume}{22}\/}, \bibinfo{pages}{231--264}.
\newblock \bibinfo{note}{\url{https://doi.org/10.1162/EVCO_a_00109}}.
\bibitem[{Rangaiah \& Bonilla-Petriciolet(2013)}]{RB2013}
\bibinfo{author}{Rangaiah, G.~P.}, \& \bibinfo{author}{Bonilla-Petriciolet, A.}
  (\bibinfo{year}{2013}).
\newblock {\it \bibinfo{title}{Multi-objective optimization in chemical
  engineering: developments and applications}\/}.
\newblock \bibinfo{publisher}{John Wiley \& Sons}.
\bibitem[{Rojas-Gonzalez \& {Van Nieuwenhuyse}(2020)}]{RI2020a}
\bibinfo{author}{Rojas-Gonzalez, S.}, \& \bibinfo{author}{{Van Nieuwenhuyse},
  I.} (\bibinfo{year}{2020}).
\newblock \bibinfo{title}{A survey on kriging-based infill algorithms for
  multiobjective simulation optimization}.
\newblock {\it \bibinfo{journal}{Computers Operations Research}\/},  {\it
  \bibinfo{volume}{116}\/}, \bibinfo{pages}{104869}.
\newblock \bibinfo{note}{\url{https://doi.org/10.1016/j.cor.2019.104869}}.
\bibitem[{Ruzika \& Wiecek(2005)}]{RW2005}
\bibinfo{author}{Ruzika, S.}, \& \bibinfo{author}{Wiecek, M.~M.}
  (\bibinfo{year}{2005}).
\newblock \bibinfo{title}{Approximation methods in multiobjective programming}.
\newblock {\it \bibinfo{journal}{Journal of Optimization Theory and
  Applications}\/},  {\it \bibinfo{volume}{126}\/}, \bibinfo{pages}{473--501}.
\newblock \bibinfo{note}{\url{https://doi.org/10.1007/s10957-005-5494-4}}.
\bibitem[{Tanabe \& Ishibuchi(2020)}]{TI2020}
\bibinfo{author}{Tanabe, R.}, \& \bibinfo{author}{Ishibuchi, H.}
  (\bibinfo{year}{2020}).
\newblock \bibinfo{title}{An easy-to-use real-world multi-objective
  optimization problem suite}.
\newblock {\it \bibinfo{journal}{Applied Soft Computing}\/},  {\it
  \bibinfo{volume}{89}\/}, \bibinfo{pages}{106078}.
\newblock \bibinfo{note}{\url{https://doi.org/10.1016/j.asoc.2020.106078}}.
\bibitem[{Tang \& Yang(2021)}]{TY2021}
\bibinfo{author}{Tang, L.}, \& \bibinfo{author}{Yang, X.}
  (\bibinfo{year}{2021}).
\newblock \bibinfo{title}{A modified direction approach for proper efficiency
  of multiobjective optimization}.
\newblock {\it \bibinfo{journal}{Optimization Methods and Software}\/},  {\it
  \bibinfo{volume}{36}\/}, \bibinfo{pages}{653--668}.
\newblock \bibinfo{note}{\url{https://doi.org/10.1080/10556788.2021.1891538}}.
\bibitem[{Trivedi et~al.(2016)Trivedi, Srinivasan, Sanyal \& Ghosh}]{TSS2016}
\bibinfo{author}{Trivedi, A.}, \bibinfo{author}{Srinivasan, D.},
  \bibinfo{author}{Sanyal, K.}, \& \bibinfo{author}{Ghosh, A.}
  (\bibinfo{year}{2016}).
\newblock \bibinfo{title}{A survey of multiobjective evolutionary algorithms
  based on decomposition}.
\newblock {\it \bibinfo{journal}{IEEE Transactions on Evolutionary
  Computation}\/},  {\it \bibinfo{volume}{21}\/}, \bibinfo{pages}{440--462}.
\newblock \bibinfo{note}{\url{https://doi.org/10.1109/TEVC.2016.2608507}}.
\bibitem[{Vaidyanathan et~al.(2003)Vaidyanathan, Tucker, Papila \&
  Shyy}]{VTPS2003}
\bibinfo{author}{Vaidyanathan, R.}, \bibinfo{author}{Tucker, K.},
  \bibinfo{author}{Papila, N.}, \& \bibinfo{author}{Shyy, W.}
  (\bibinfo{year}{2003}).
\newblock \bibinfo{title}{{CFD}-based design optimization for single element
  rocket injector}.
\newblock In {\it \bibinfo{booktitle}{41st Aerospace Sciences Meeting and
  Exhibit}\/} (pp. \bibinfo{pages}{1--21}).
\newblock \bibinfo{note}{\url{https://doi.org/10.2514/6.2003-296}}.
\bibitem[{Valenzuela-Rend{\'o}n et~al.(1997)Valenzuela-Rend{\'o}n,
  Uresti-Charre \& Monterrey}]{VU1997}
\bibinfo{author}{Valenzuela-Rend{\'o}n, M.}, \bibinfo{author}{Uresti-Charre,
  E.}, \& \bibinfo{author}{Monterrey, I.} (\bibinfo{year}{1997}).
\newblock \bibinfo{title}{A non-generational genetic algorithm for
  multiobjective optimization}.
\newblock In {\it \bibinfo{booktitle}{Proceedings of the Seventh International
  Conference on Genetic Algorithms}\/} (pp. \bibinfo{pages}{658--665}).
\newblock \bibinfo{publisher}{Morgan Kaufmann}.
\bibitem[{Vlennet et~al.(1996)Vlennet, Fonteix \& Marc}]{VFM1996}
\bibinfo{author}{Vlennet, R.}, \bibinfo{author}{Fonteix, C.}, \&
  \bibinfo{author}{Marc, I.} (\bibinfo{year}{1996}).
\newblock \bibinfo{title}{Multicriteria optimization using a genetic algorithm
  for determining a {P}areto set}.
\newblock {\it \bibinfo{journal}{International Journal of Systems Science}\/},
  {\it \bibinfo{volume}{27}\/}, \bibinfo{pages}{255--260}.
\newblock \bibinfo{note}{\url{https://doi.org/10.1080/00207729608929211}}.
\bibitem[{Wang et~al.(2020)Wang, Su, Lin, Ma, Gong, Li \& Ming}]{WSLM2020}
\bibinfo{author}{Wang, J.}, \bibinfo{author}{Su, Y.}, \bibinfo{author}{Lin,
  Q.}, \bibinfo{author}{Ma, L.}, \bibinfo{author}{Gong, D.},
  \bibinfo{author}{Li, J.}, \& \bibinfo{author}{Ming, Z.}
  (\bibinfo{year}{2020}).
\newblock \bibinfo{title}{A survey of decomposition approaches in
  multiobjective evolutionary algorithms}.
\newblock {\it \bibinfo{journal}{Neurocomputing}\/},  {\it
  \bibinfo{volume}{408}\/}, \bibinfo{pages}{308--330}.
\newblock \bibinfo{note}{\url{https://doi.org/10.1016/j.neucom.2020.01.114}}.
\bibitem[{Wang et~al.(2015)Wang, Purshouse, Giagkiozis \& Fleming}]{WPG2015}
\bibinfo{author}{Wang, R.}, \bibinfo{author}{Purshouse, R.~C.},
  \bibinfo{author}{Giagkiozis, I.}, \& \bibinfo{author}{Fleming, P.~J.}
  (\bibinfo{year}{2015}).
\newblock \bibinfo{title}{The i{PICEA}-g: a new hybrid evolutionary
  multi-criteria decision making approach using the brushing technique}.
\newblock {\it \bibinfo{journal}{European Journal of Operational Research}\/},
  {\it \bibinfo{volume}{243}\/}, \bibinfo{pages}{442--453}.
\newblock \bibinfo{note}{\url{https://doi.org/10.1016/j.ejor.2014.10.056}}.
\bibitem[{Wang et~al.(2016)Wang, Zhang \& Zhang}]{WZZ2016}
\bibinfo{author}{Wang, R.}, \bibinfo{author}{Zhang, Q.}, \&
  \bibinfo{author}{Zhang, T.} (\bibinfo{year}{2016}).
\newblock \bibinfo{title}{Decomposition-based algorithms using {P}areto
  adaptive scalarizing methods}.
\newblock {\it \bibinfo{journal}{IEEE Transactions on Evolutionary
  Computation}\/},  {\it \bibinfo{volume}{20}\/}, \bibinfo{pages}{821--837}.
\newblock \bibinfo{note}{\url{https://doi.org/10.1109/TEVC.2016.2521175}}.
\bibitem[{Xu et~al.(2018)Xu, Ding, Qu \& Li}]{QR2018}
\bibinfo{author}{Xu, Y.}, \bibinfo{author}{Ding, O.}, \bibinfo{author}{Qu, R.},
  \& \bibinfo{author}{Li, K.} (\bibinfo{year}{2018}).
\newblock \bibinfo{title}{Hybrid multi-objective evolutionary algorithms based
  on decomposition for wireless sensor network coverage optimization}.
\newblock {\it \bibinfo{journal}{Applied Soft Computing}\/},  {\it
  \bibinfo{volume}{68}\/}, \bibinfo{pages}{268--282}.
\newblock \bibinfo{note}{\url{https://doi.org/10.1016/j.asoc.2018.03.053}}.
\bibitem[{Xu et~al.(2013)Xu, Qu \& Li}]{QR2013}
\bibinfo{author}{Xu, Y.}, \bibinfo{author}{Qu, R.}, \& \bibinfo{author}{Li, R.}
  (\bibinfo{year}{2013}).
\newblock \bibinfo{title}{A simulated annealing based genetic local search
  algorithm for multi-objective multicast routing problems}.
\newblock {\it \bibinfo{journal}{Annals of Operations Research}\/},  {\it
  \bibinfo{volume}{206}\/}, \bibinfo{pages}{527--555}.
\newblock \bibinfo{note}{\url{https://doi.org/10.1007/s10479-013-1322-7}}.
\bibitem[{Zhang \& Li(2007)}]{ZL2007}
\bibinfo{author}{Zhang, Q.}, \& \bibinfo{author}{Li, H.}
  (\bibinfo{year}{2007}).
\newblock \bibinfo{title}{{MOEA/D}: A multiobjective evolutionary algorithm
  based on decomposition}.
\newblock {\it \bibinfo{journal}{IEEE Transactions on Evolutionary
  Computation}\/},  {\it \bibinfo{volume}{11}\/}, \bibinfo{pages}{712--731}.
\newblock \bibinfo{note}{\url{https://doi.org/10.1109/TEVC.2007.892759}}.
\bibitem[{Zhang et~al.(2010)Zhang, Li, Maringer \& Tsang}]{ZLD2010}
\bibinfo{author}{Zhang, Q.}, \bibinfo{author}{Li, H.},
  \bibinfo{author}{Maringer, D.}, \& \bibinfo{author}{Tsang, E.}
  (\bibinfo{year}{2010}).
\newblock \bibinfo{title}{{MOEA/D} with {NBI}-style tchebycheff approach for
  portfolio management}.
\newblock In {\it \bibinfo{booktitle}{IEEE Congress on Evolutionary
  Computation}\/} (pp. \bibinfo{pages}{1--8}).
\newblock \bibinfo{note}{\url{https://doi.org/10.1109/CEC.2010.5586185}}.
\bibitem[{Zitzler et~al.(2000)Zitzler, Deb \& Thiele}]{ZDT2000}
\bibinfo{author}{Zitzler, E.}, \bibinfo{author}{Deb, K.}, \&
  \bibinfo{author}{Thiele, L.} (\bibinfo{year}{2000}).
\newblock \bibinfo{title}{Comparison of multiobjective evolutionary algorithms:
  Empirical results}.
\newblock {\it \bibinfo{journal}{Evolutionary Computation}\/},  {\it
  \bibinfo{volume}{8}\/}, \bibinfo{pages}{173--195}.
\newblock \bibinfo{note}{\url{https://doi.org/10.1162/106365600568202}}.
\bibitem[{Zitzler \& Thiele(1999)}]{ZT1999}
\bibinfo{author}{Zitzler, E.}, \& \bibinfo{author}{Thiele, L.}
  (\bibinfo{year}{1999}).
\newblock \bibinfo{title}{Multiobjective evolutionary algorithms: a comparative
  case study and the strength {P}areto approach}.
\newblock {\it \bibinfo{journal}{IEEE Transactions on Evolutionary
  Computation}\/},  {\it \bibinfo{volume}{3}\/}, \bibinfo{pages}{257--271}.
\newblock \bibinfo{note}{\url{https://doi.org/10.1109/4235.797969}}.
\bibitem[{Zitzler et~al.(2003)Zitzler, Thiele, Laumanns, Fonseca \&
  Da~Fonseca}]{ZTL2003}
\bibinfo{author}{Zitzler, E.}, \bibinfo{author}{Thiele, L.},
  \bibinfo{author}{Laumanns, M.}, \bibinfo{author}{Fonseca, C.~M.}, \&
  \bibinfo{author}{Da~Fonseca, V.~G.} (\bibinfo{year}{2003}).
\newblock \bibinfo{title}{Performance assessment of multiobjective optimizers:
  An analysis and review}.
\newblock {\it \bibinfo{journal}{IEEE Transactions on Evolutionary
  Computation}\/},  {\it \bibinfo{volume}{7}\/}, \bibinfo{pages}{117--132}.
\newblock \bibinfo{note}{\url{https://doi.org/10.1109/TEVC.2003.810758}}.
\bibitem[{Zopounidis et~al.(2015)Zopounidis, Galariotis, Doumpos, Sarri \&
  Andriosopoulos}]{Z2015m}
\bibinfo{author}{Zopounidis, C.}, \bibinfo{author}{Galariotis, E.},
  \bibinfo{author}{Doumpos, M.}, \bibinfo{author}{Sarri, S.}, \&
  \bibinfo{author}{Andriosopoulos, K.} (\bibinfo{year}{2015}).
\newblock \bibinfo{title}{Multiple criteria decision aiding for finance: An
  updated bibliographic survey}.
\newblock {\it \bibinfo{journal}{European Journal of Operational Research}\/},
  {\it \bibinfo{volume}{247}\/}, \bibinfo{pages}{339--348}.
\newblock \bibinfo{note}{\url{https://doi.org/10.1016/j.ejor.2015.05.032}}.

\end{thebibliography}
\end{spacing}

\end{document}